\documentclass[10pt]{article} % For LaTeX2e
\usepackage[accepted]{tmlr}
\usepackage{hyperref}
\usepackage{url}
\usepackage[utf8]{inputenc} % allow utf-8 input
\usepackage[T1]{fontenc}    % use 8-bit T1 fonts

\usepackage{url}            % simple URL typesetting
\usepackage{booktabs}       % professional-quality tables
\usepackage{amsfonts}       % blackboard math symbols
\usepackage{nicefrac}       % compact symbols for 1/2, etc.
\usepackage{microtype}      % microtypography
\usepackage{graphicx}
\usepackage{amsmath}
\usepackage{caption}
\usepackage{subcaption}
\usepackage{multirow}
\usepackage{amssymb}
\usepackage{enumitem}
\usepackage{tikz}
\usepackage{xcolor}
\usepackage{pgfplots}\pgfplotsset{compat=1.9}
\usetikzlibrary{pgfplots.groupplots}
\usepackage{wrapfig}
\usepackage{adjustbox}
\usepackage{framed}
\usepackage{pgfplots}
\pgfplotsset{width=10cm,compat=1.9}
\usepackage{pifont}
\newcommand{\cmark}{\ding{51}}%
\newcommand{\deemph}[1]{{\color{black!40}#1}}
\usepackage{makecell}
\usepackage{color, colortbl}

\usepackage[ruled,vlined]{algorithm2e}
\title{Test-time Contrastive Concepts for Open-world Semantic Segmentation with Vision-Language Models}

\author{%
    Monika Wysoczańska$^1$\thanks{Corresponding author: 
    \texttt{monika.wysoczanska.dokt@pw.edu.pl}}, Antonin Vobecky$^{2,3,4}$, 
   Amaia Cardiel$^{2,7}$, 
   Tomasz Trzciński$^{1,5}$, \\
   Renaud Marlet$^{2,6}$,
   Andrei Bursuc$^2$,
   Oriane Siméoni$^2$
   \AND
  \normalfont $^1$Warsaw University of Technology 
  $\>$
  $^2$valeo.ai 
  $\>$
  $^3$CIIRC CTU Prague\thanks{Czech Technical University in Prague, Czech Institute of Informatics, Robotics and Cybernetics} 
  $\>$
  $^4$FEE CTU Prague 
  $\>$
  $^5$Tooploox 
  \\
  $^6$LIGM, École des Ponts et Chaussées, IP Paris, CNRS, France 
  $\>$
  $^7$Université Grenoble Alpes 
}
\def\month{05}  % Insert correct month for camera-ready version
\def\year{2025} % Insert correct year for camera-ready version
\def\openreview{\url{https://openreview.net/forum?id=4256}} % Insert correct link to OpenReview for camera-ready version

\definecolor{seat}{RGB}{64, 128, 0}
\definecolor{sofa}{RGB}{118, 116, 21}
\definecolor{furniture}{RGB}{23,116,19}
\definecolor{bed}{RGB}{192,25,251}
\definecolor{pillow}{RGB}{0,15,115}
\definecolor{nightstand}{RGB}{252,53,25}
\definecolor{lamp}{RGB}{251,0,136}
\definecolor{rug}{RGB}{165,29,71}
\definecolor{bed}{RGB}{192,25,251}
\definecolor{ceiling}{RGB}{120,120,85}
\definecolor{wall}{RGB}{0,0,123}
\definecolor{way}{RGB}{235,85,40}
\definecolor{style}{RGB}{68,8,125}
\definecolor{floor}{RGB}{237,114,178}
\definecolor{tree}{RGB}{128,128,128}
\definecolor{sky}{RGB}{255,255,84}
\definecolor{bird}{RGB}{187,40,246}
\definecolor{airport}{RGB}{55,126,127}
\definecolor{building}{RGB}{130,73,32}
\definecolor{cityscape}{RGB}{232,167,108}
\definecolor{aeroplane}{RGB}{171,123,121}
\definecolor{car}{RGB}{0,0,245}
\definecolor{city}{RGB}{55,126,34}
\definecolor{airport}{RGB}{55,126,127}
\definecolor{poster}{RGB}{235,85,40}
\definecolor{air}{RGB}{68,8,125}
\definecolor{branch}{RGB}{0,0,123}
\definecolor{background}{RGB}{0,0,0}
\definecolor{muffin}{RGB}{143,255,140}
\definecolor{bicycle}{RGB}{171,176,182}
\definecolor{bicycle}{RGB}{171,176,182}
\definecolor{boat}{RGB}{0, 191, 255}
\definecolor{skyscraper}{RGB}{144,238,143}
\definecolor{bed_sumpat}{RGB}{80,51,50}
\definecolor{arm}{RGB}{144,38,144}
\definecolor{person_sup}{RGB}{45,238,144}
\newcommand{\promptstyle}[2]{\footnotesize{\textbf{\texttt{\color{#1}#2}}}}

\newcolumntype{a}{>{\columncolor{white!80!black}}c}
\newcolumntype{d}{>{\columncolor{white!93!black}}c}

\newcommand{\metric}{\relax{IoU-single}\xspace}
\newcommand{\CC}{\mathcal{CC}}
\newcommand{\cc}{$\CC$\xspace}
\newcommand{\CCD}{\CC^D}
\newcommand{\ccD}{$\CCD$\xspace}
\newcommand{\CCLLM}{\CC^{L}}
\newcommand{\ccLLM}{$\CCLLM$\xspace}
\newcommand{\CCP}{\CC^{PI}}
\newcommand{\ccP}{$\CCP$}
\newcommand{\CCB}{\CC^{BG}}
\newcommand{\ccB}{$\CC^{BG}$}
\newcommand{\Naturals}{\mathbb{N}}
\newcommand{\Reals}{\mathbb{R}}
\newcommand{\rest}{\bot}
\newcommand{\tenc}{\phi_{\text T}}
\newcommand{\venc}{\phi_{\text V}}
\newcommand{\query}{q} 
\newcommand{\Queries}{Q}
\newcommand{\Lexicon}{\mathcal{T}}
\newcommand{\image}{\mathbf{I}}
\newcommand{\openwseg}{\mathbf{S}_{\text{openw}}}
\newcommand{\closewseg}{\mathbf{S}_{\text{closew}}}

\newcommand{\ccword}{contrastive concept}

\usepackage[capitalize]{cleveref}
\crefname{section}{Sec.}{Secs.}
\Crefname{section}{Section}{Sections}
\Crefname{table}{Table}{Tables}
\crefname{table}{Tab.}{Tabs.}

\begin{document}

\maketitle

\begin{abstract}
Recent CLIP-like Vision-Language Models (VLMs), pre-trained on large amounts of image-text pairs to align both modalities with a simple contrastive objective, have paved 
the way to open-vocabulary semantic segmentation. 
Given an arbitrary set of textual queries, image pixels are assigned the closest query in feature space. However, this works well when a user exhaustively lists all possible visual concepts in an image that contrast
against each other for the assignment.
This corresponds to the current evaluation setup in the literature, which relies on having access to a list of in-domain relevant concepts, typically classes of a benchmark dataset. Here, we consider the more challenging (and realistic) scenario of segmenting a single concept, given a textual prompt and nothing else.
To achieve good results, besides contrasting with the generic ``background'' text, we propose two different approaches to automatically generate, at test time, query-specific textual contrastive concepts. 
We do so by leveraging the distribution of text in the VLM's training set or crafted LLM prompts. We also propose a metric designed to evaluate this scenario and show the relevance of our approach on commonly used datasets.

\end{abstract}

\section{Introduction}

Vision-language models (VLMs) such as CLIP \cite{radford2021learning} are trained to align text and global image representations. % 
Recently, VLMs have been proposed for denser tasks \cite{zhou2022maskclip, ghiasi2022openseg, li2022languagedriven}. This includes the challenging pixel-level task of open-vocabulary semantic segmentation (OVSS), which consists of segmenting arbitrary \emph{visual concepts} in images, i.e., visual entities such as objects, stuff (e.g., grass), or visual phenomena (e.g., sky). To that end,  several methods exploit a frozen CLIP model with 
additional operations \cite{zhou2022maskclip, bousselham2023gem, wysoczanska2023clipdino, wysoczanska2023clipdiy}, or fine-tune the model with specific losses \cite{xu2022groupvit, clippy2022, cha2022tcl, lou2022segclip, mukhoti2023pacl}.  

\begin{figure}[t]
    \centering
    % \resizebox{\linewidth}{!}{
    % \fbox{
    \begin{tabular}{clc@{}c@{}c@{}c}
        & & Image & GT & query % pred. 
        & query % w/ bkg 
        \\ 
        & & &  & \texttt{+}\,``background'' % w/ bkg 
        & \texttt{+}``background''\texttt{+}\,\cc
        %\& our $\mathcal{CC}$ 
        \\ 
       & \raisebox{1.\normalbaselineskip}[0pt][0pt]{\rotatebox{90}{\texttt{corgi}}} & \includegraphics[width=3.5cm]{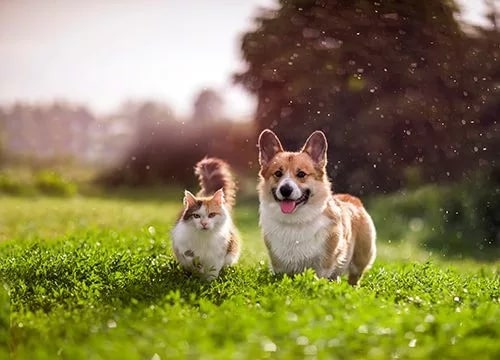} & \includegraphics[width=3.5cm]{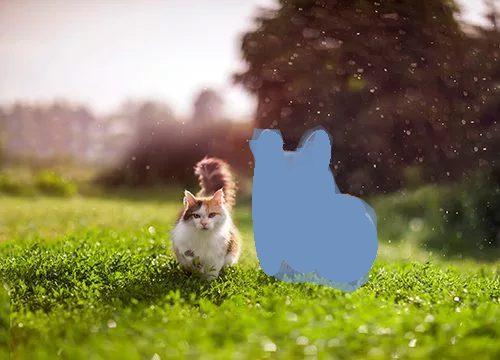} & 
        \includegraphics[width=3.5cm]{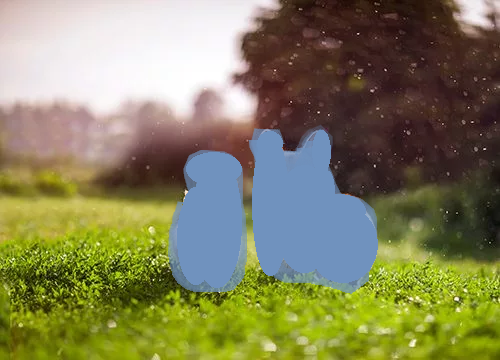} &
        \includegraphics[width=3.5cm, trim={0 5.5cm 19.8cm 0},clip]{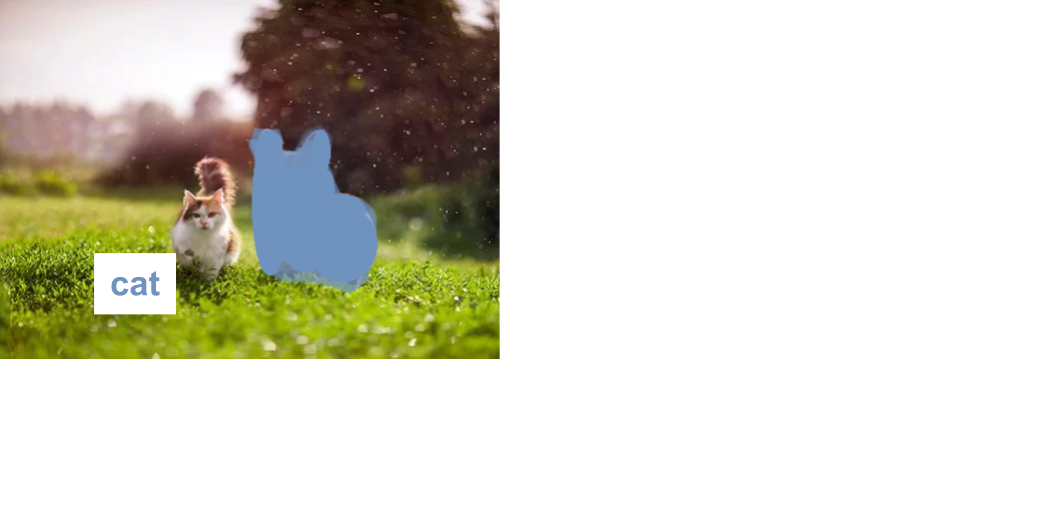} \\
      \raisebox{0.8\normalbaselineskip}[0pt][0pt]{\rotatebox{90}{\hspace{-5mm}Query prompts}} & \raisebox{1.5\normalbaselineskip}[0pt][0pt]{\rotatebox{90}{\texttt{car}}}   & \includegraphics[width=3.5cm]{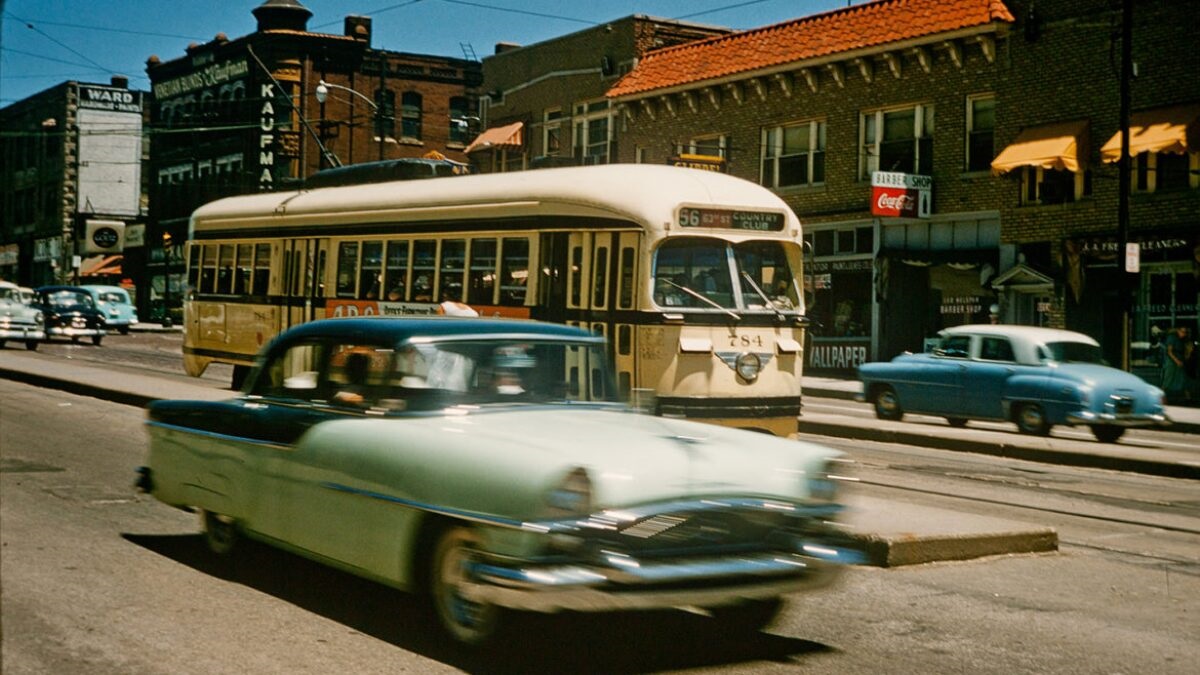} & \includegraphics[width=3.5cm]{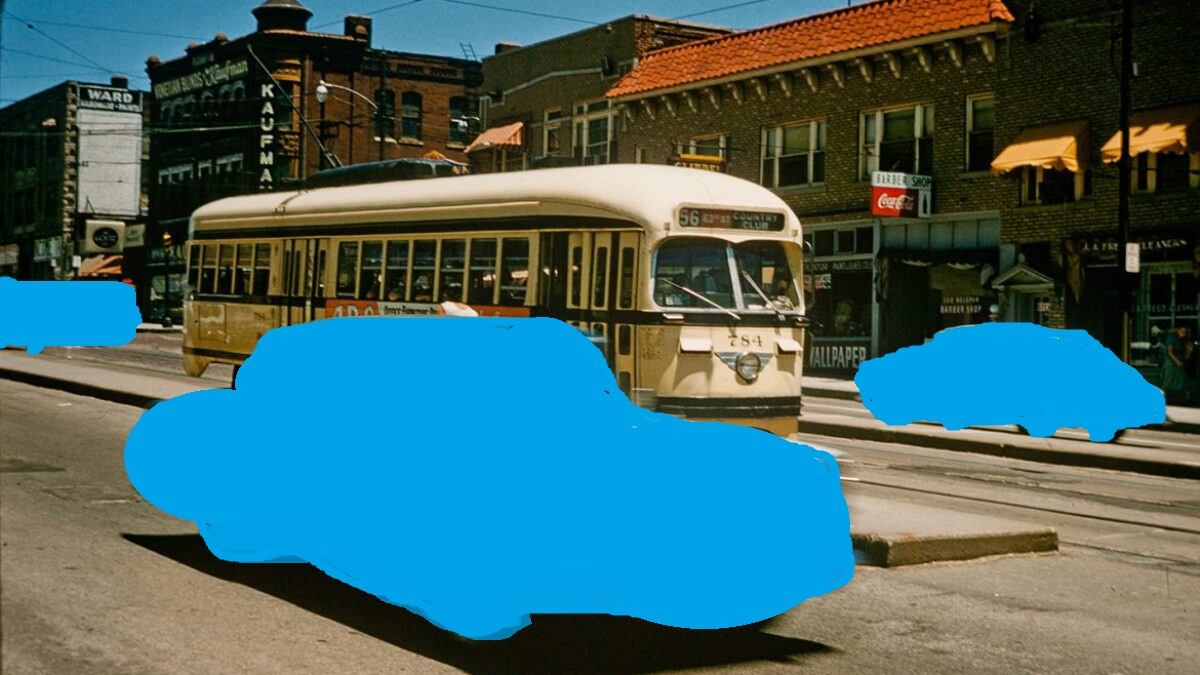} & 
        \includegraphics[width=3.5cm]{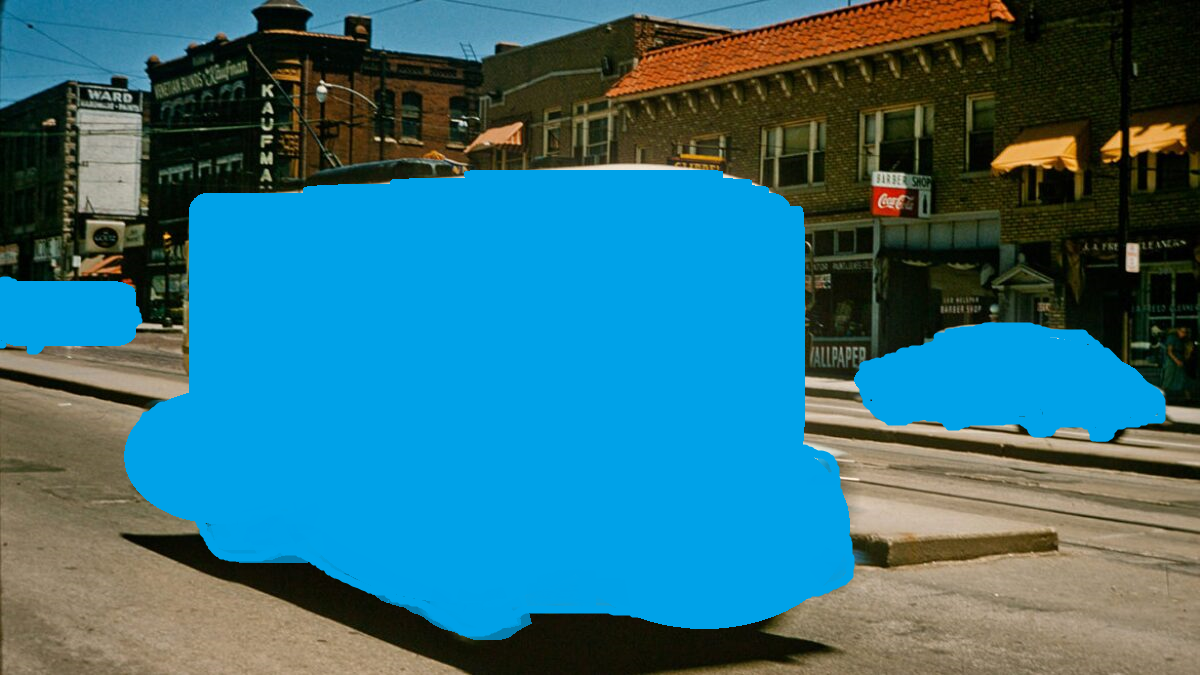} &
        \includegraphics[width=3.5cm, trim={0 0 5cm 0},clip]{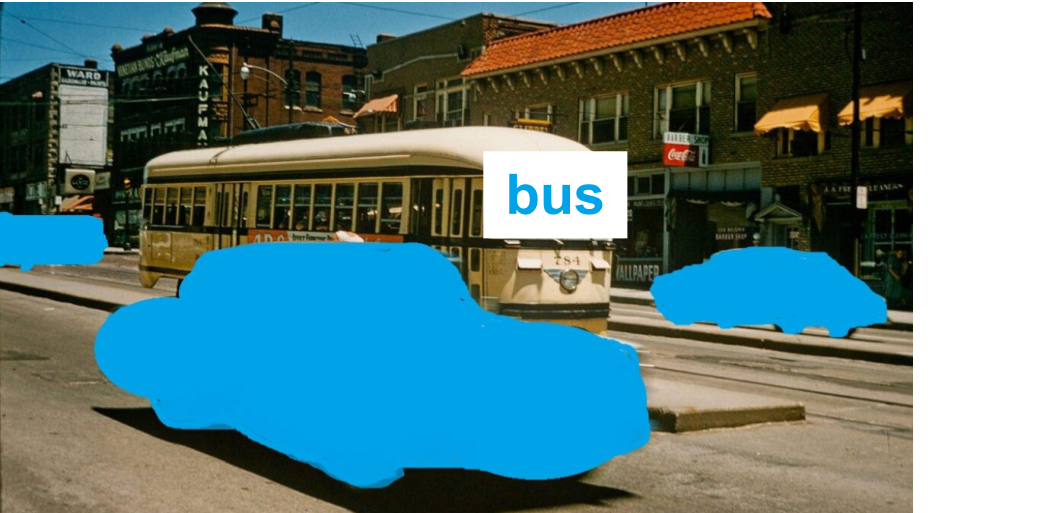} \\
         & \raisebox{1.75\normalbaselineskip}[0pt][0pt]{\rotatebox{90}{\texttt{boat}}} & \includegraphics[width=3.5cm]{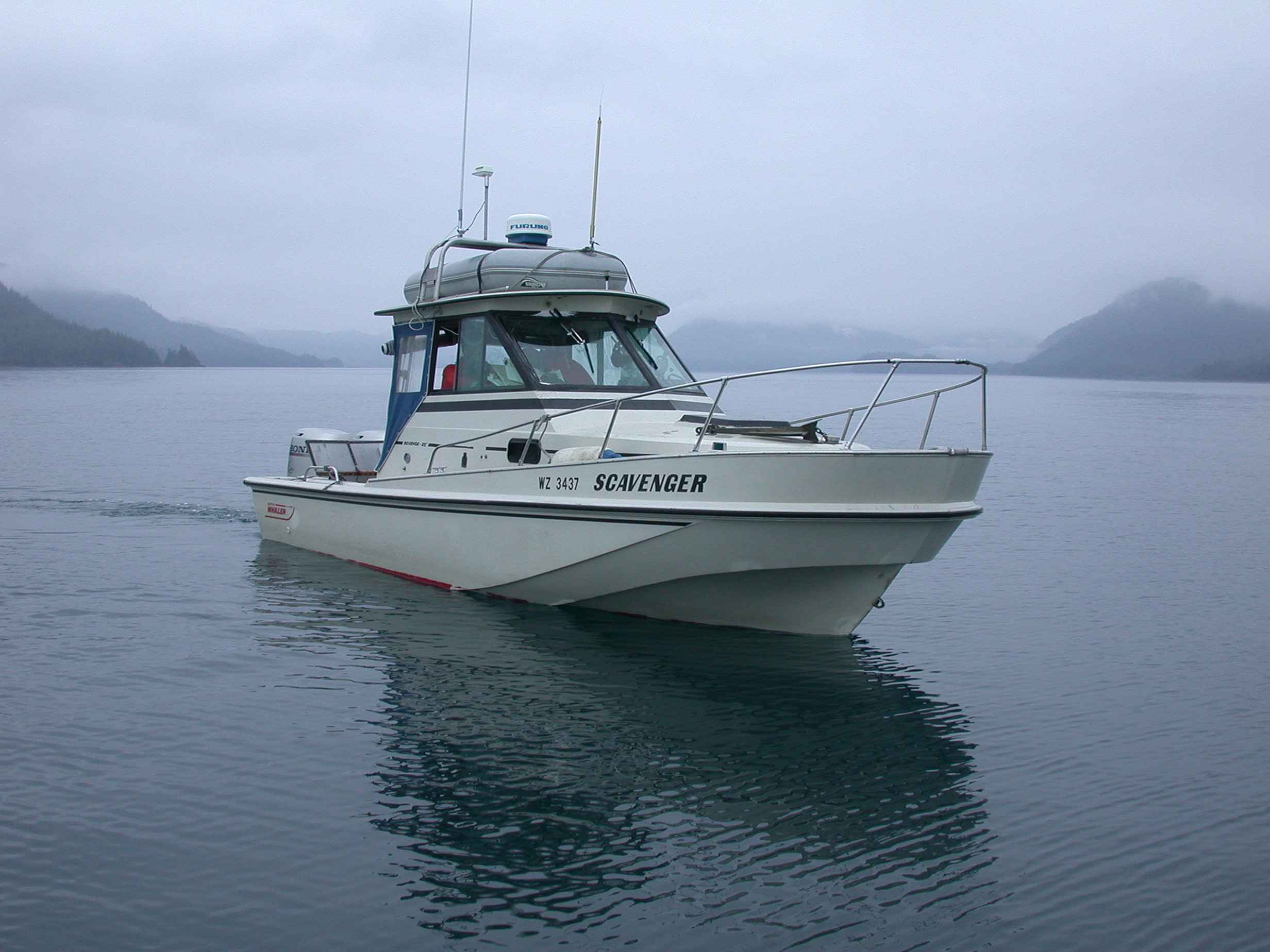} & \includegraphics[width=3.5cm]{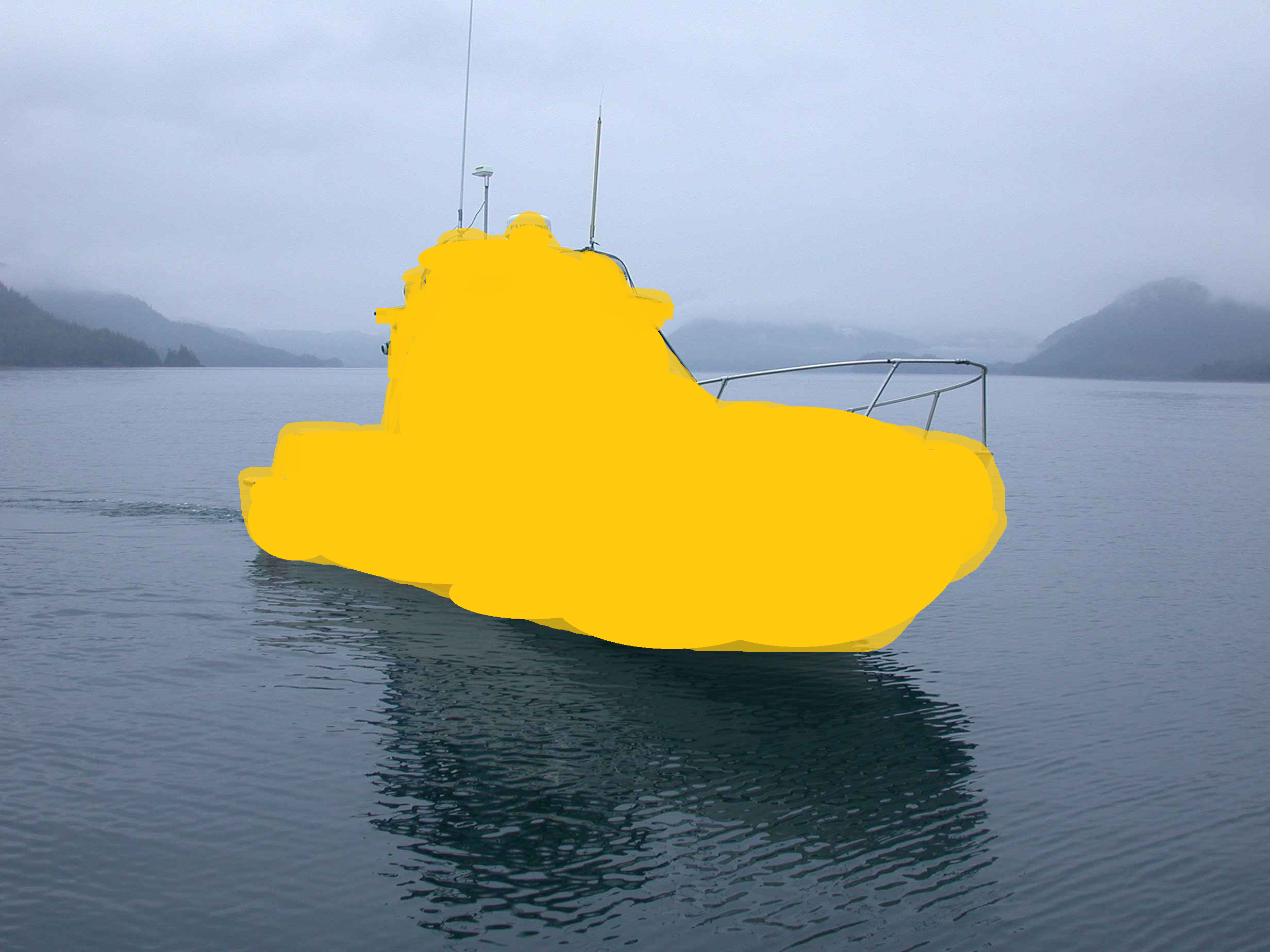} & 
        \includegraphics[width=3.5cm]{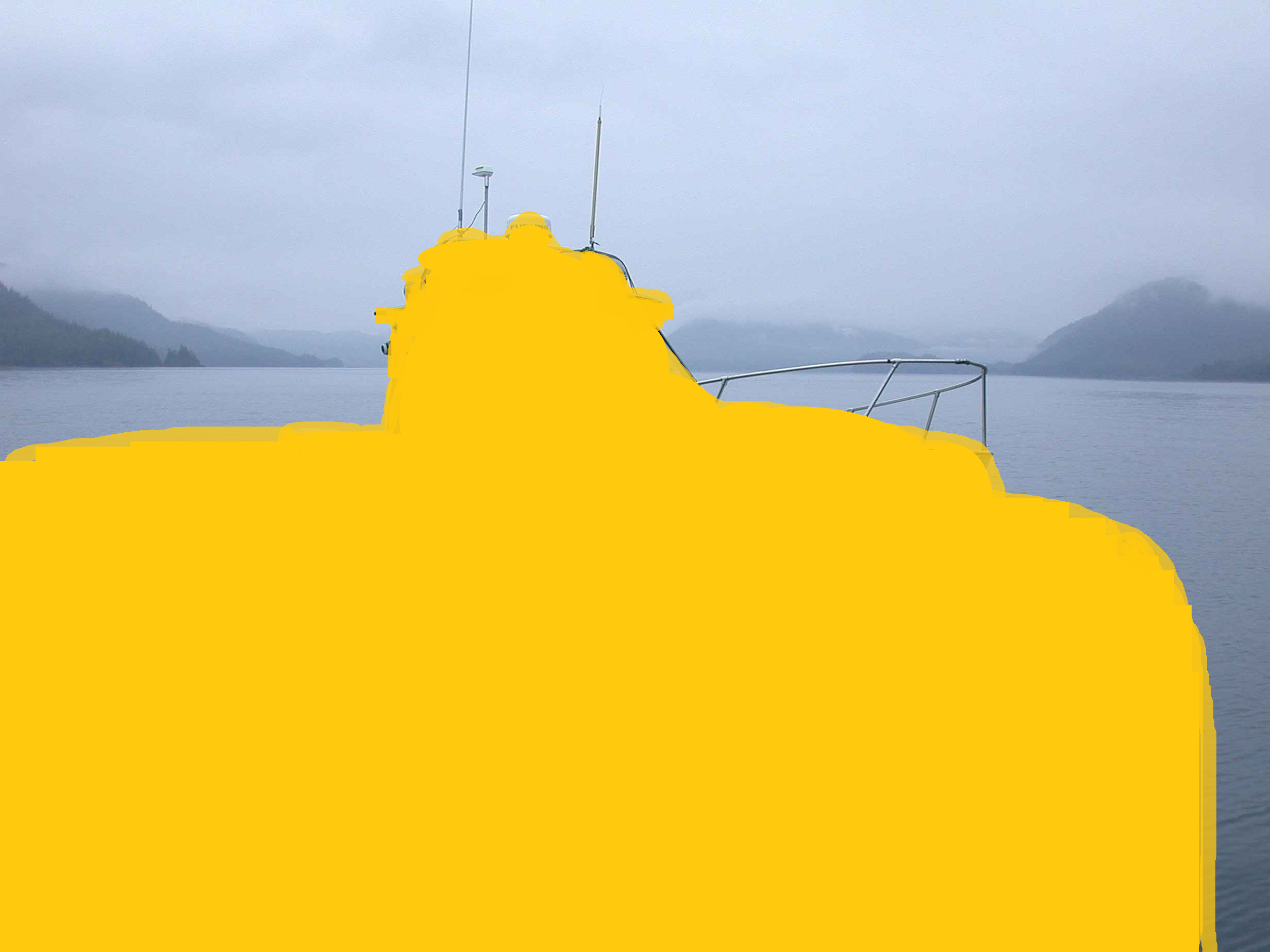} &
        \includegraphics[width=3.5cm, trim={0 0 13cm 0},clip]{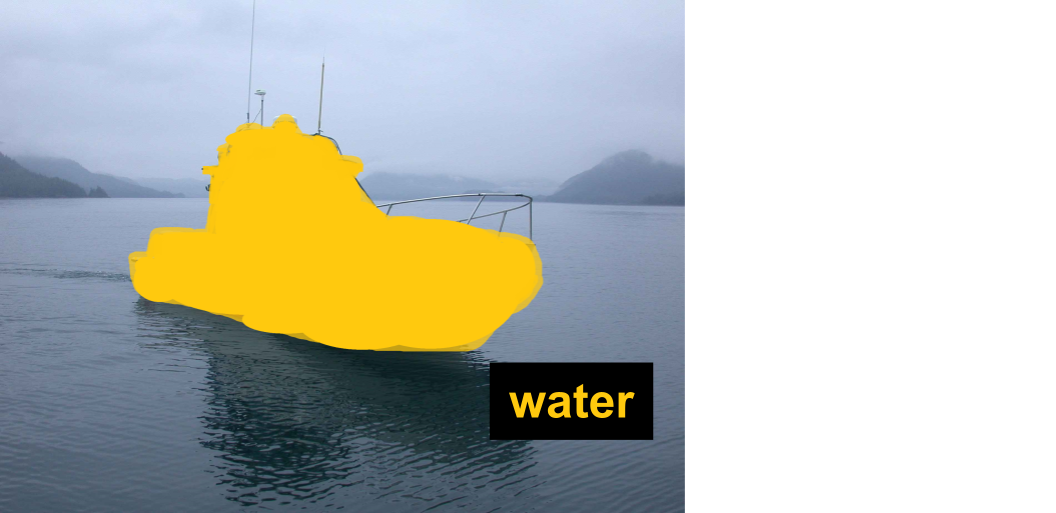}
    \end{tabular}
    % }
    % }  % resizebox
    \caption{
    \textbf{Illustration of our proposed open-world scenario and benefits of 
    contrastive concepts (\cc).} We investigate open-world segmentation, where only one (or a few) visual concepts are to be segmented (2\textsuperscript{nd} column), while
    all concepts that can occur in an image are unknown. 
    Contrasting the query with ``background'' allows us to obtain a coarse segmentation \cite{clippy2022, wysoczanska2023clipdino} (3\textsuperscript{rd} column), but is not enough to catch all pixels \emph{not} corresponding to the query when they are related or co-occur frequently in the VLM training set. Our automatically-generated \emph{contrastive concepts} (\cc) (4\textsuperscript{th} column) help to separate and disentangle pixels of the query (right column, generated \cc\ in text boxes), therefore achieving better segmentation.}
    \label{fig:teaser}
\end{figure}

Most OVSS methods label each pixel with the most probable prompt (or query) among a finite set of prompts provided as input, contrasting concepts with each other. 
This works well for benchmarks that provide a large and nearly exhaustive list of things that can be found in the dataset images, such as ADE20K \cite{zhou2019ade20k} or COCO-Stuff \cite{caesar2018cocostuff}.
However, when given a limited list of queries,
these methods are bound to occasionally suffer from hallucinations \cite{wysoczanska2023clipdino, miller2024open}. 
In particular, common setups do not 
handle cases where only a single concept is queried \cite{cha2022tcl, xu2022groupvit}, which results in classifying all pixels using the same concept. 

To catch such hallucinations, a common strategy consists of using an extra class labeled `background', intended to capture pixels that do not correspond to any visual concept being queried. 
This extra class is already in object-centric datasets, such as Pascal VOC \cite{pascal-voc-2012}.
It provides an easy, generic concept to be used as a negative query, i.e., to be used to contrast with actual (positive) queries but to be discarded from the final segmentation.
However, the notion of background is not well defined as it is context-dependent, therefore providing suboptimal contrasts. 
This strategy also fails when a queried concept (e.g., “tree”) falls in the learned background (which commonly encompasses trees).

In this work, we consider a practical and realistic OVSS task in which only one or a few arbitrary concepts are to be segmented, leaving out the remaining pixels
without prior knowledge of other concepts that may occur in an image. We name this setup \emph{open-world} \footnote{We distinguish our setup from open-world/ open-set setting known from literature~\cite{wu2024towards}, where a segmentation model
identifies novel classes and marks them as “unknown”. Here, we consider the task of \textit{open-world open-vocabulary} segmentation, thus considering only OVSS models, where the goal is to segment queried concepts unknown at test time and leave the remaining pixels in an image with no class. For the full name of our setup, we thus consider \textit{open-world open-vocabulary segmentation} but keep \textit{open-world} throughout the rest of the work for brevity.}.
Given a query, instead of assuming access to a dataset-specific set of classes (a closed-world setup), we propose to automatically suggest contrastive concepts useful to better localize the queried concept, although they can later be ignored. In particular, we focus on predicting concepts likely to co-occur with the queried concept, e.g., ``water'' for the query ``boat'' (as visible in \cref{fig:teaser}), thus leading to better segment boundaries when prompted together. 

Moreover, we argue that this scenario needs to be evaluated to understand the limitations of open-vocabulary segmentation methods better. We therefore propose a new metric to measure such an ability, namely \metric, which considers one query prompt at a time and thus does not rely on the knowledge of potential domain classes. 

To summarize, our contributions are as follows:

\begin{itemize}[itemsep=-1pt,topsep=-6pt]
    \item We introduce the notion of test-time contrastive concepts and discuss the importance of contrastive concepts in open-vocabulary semantic segmentation.
    \item We analyze the usage of ``background'' as a test-time contrastive concept, which has been accepted but not discussed so far. 
     \item We propose a new single-query evaluation setup for open-world semantic segmentation that does not rely on domain knowledge.
    We also propose a new metric to evaluate the grounding of visual concepts. 
    \item We propose two different methods to generate test-time 
    contrastive concepts automatically and show that our approaches consistently improve the results of 7 different popular OVSS methods or backbones.
\end{itemize}

\section{Related work}

\vspace{-5pt}
\paragraph{Open-vocabulary semantic segmentation.}
VLMs trained on web-collected data to produce aligned image-text representations \cite{radford2021learning,jia2021scaling,zhai2023sigmoid} had a major impact on open-vocabulary perception tasks and opened up new avenues for research and practical applications. While CLIP can be used \emph{off-the-shelf} for image classification in different settings, it does not % a major shortcoming is its inability to 
produce dense pixel-level features and predictions, due to its final global attentive-pooling \cite{zhou2022maskclip, jatavallabhula2023conceptfusion}. To mitigate this and produce dense image-text features, several methods fine-tune CLIP with dense supervision~\cite{cho2024catseg, xu2023side, xu2023masq, ding2023maskclip, wu2023clipself}. Other approaches devise new CLIP-like models trained from scratch using a pooling compatible with segmentation. Their supervision comes from large datasets annotated with coarse captions \cite{ghiasi2022openseg, clippy2022,liang2023open, xu2022groupvit,liu2022open,xu2023learning, mukhoti2023pacl, cha2022tcl}, object masks \cite{rao2021denseclip, ghiasi2022openseg, ding2021ZegFormer} or pixel labels \cite{li2022languagedriven, liang2023open}. However, 
when models are fine-tuned, they face feature degradation \cite{jatavallabhula2023conceptfusion}, or require long training cycles on large amounts of images when trained from scratch.

CLIP densification methods have emerged as a low-cost alternative to produce pixel-level image-text features while keeping CLIP frozen \cite{zhou2022maskclip,wysoczanska2023clipdiy,jatavallabhula2023conceptfusion,abdelreheem2023satr,wysoczanska2023clipdino, bousselham2023gem}. The seminal MaskCLIP \cite{zhou2022maskclip} mimics the global pooling layer of CLIP with a $1\,{\times}\,1$ conv layer. The aggregation of features from multiple views and crops \cite{abdelreheem2023satr, kerr2023lerf, wysoczanska2023clipdiy,jatavallabhula2023conceptfusion} also leads to dense features, yet with the additional cost of multiple forward passes. Some methods \cite{shin2022reco,shin2023namedmask, karazija2023diffusion} rely on codebooks of visual prototypes per concept, including per-dataset negative prototypes \cite{karazija2023diffusion}, or leverage self-self-attention to create groups of similar tokens \cite{bousselham2023gem}. The recent CLIP-DINOiser \cite{wysoczanska2023clipdino} improves MaskCLIP features with limited computational overhead thanks to a guided pooling strategy that leverages the correlation information from DINO features \cite{caron2021dino}. 

\paragraph{Prompt augmentation.} 
Prompt engineering is a common practice for adapting Large Language Models (LLMs) to different language tasks \cite{kojima2022large} without updating parameters. This strategy of carefully selecting task-specific prompts also improves the performance of VLMs. 
For instance, in the original CLIP work \cite{radford2021learning}, dataset-specific prompt templates, e.g., ``\relax{a photo of the nice \{$\cdots\!$\hspace{0.5mm}\}}'' were devised towards improving zero-shot prediction performance. Although effective, manual prompting can be a laborious task, as templates must be adapted per dataset and sufficiently general to apply to all classes. Afterwards, different automated strategies were subsequently explored,
e.g., scoring and ensembling predictions from multiple prompts \cite{allingham2023simple}.
Prompts can also be augmented by exploiting semantic relations 
between concepts defined in WordNet \cite{fellbaum1998wordnet} to generate new coarse/fine-grained \cite{ge2023improving} or synonym \cite{lin2023clip} prompts. LLMs can be used as a knowledge base to produce rich visual descriptions adapted for each class starting from simple class names \cite{pratt2023does,menon2023visual}. Prompt features can be learned by 
% taking into account 
considering visual co-occurrences \cite{gupta2019vico}, a connection between training and test distributions \cite{xiao2024any}, mining important features for the VLM \cite{esfandiarpoor2024if} or by test-time tuning on a sample \cite{shu2022test}. Most of these strategies have been designed and evaluated for the image classification task, and their generalization and scalability for semantic segmentation are not always trivial. 
Here, we aim to obtain better prompts for semantic segmentation to separate queried object pixels from their background.
We do this automatically without supervision and without changing the parameters of either the text encoder or the image encoder, leveraging statistics from VLM training data or LLM-based knowledge.

\paragraph{Dealing with \ccword s in OVSS.} 
Our \ccword~discovery is tightly related to \emph{background handling} in the context of open-vocabulary semantic segmentation, since the standard benchmark datasets for this task, originally designed for supervised learning, use \emph{background} 
to describe unlabeled pixels, for example, to cover concepts outside of the dataset vocabulary. There are three main types of approaches to address this problem. The first one is to threshold uncertain predictions \cite{cha2022tcl, bousselham2023gem, xu2022groupvit} with a given probability value \cite{xu2022groupvit, bousselham2023gem} or clip similarities \cite{cha2022tcl}.
The second group of methods leverages the object-centric nature of certain datasets by defining background through visual saliency~\cite{wysoczanska2023clipdiy, wysoczanska2023clipdino}. 
Finally, a significant body of work addresses the same issue by defining dataset-level concepts either by adding handcrafted names of concepts to the background definition \cite{lin2023tagclip, yu2023fcclip, clippy2022, cho2024catseg} or by extracting visual \emph{negative prototypes} with a large diffusion model \cite{karazija2023diffusion}. 
In contrast, in this work, we aim for automatic discovery of \ccword s without prior access to the vocabulary used to annotate the dataset.

\paragraph{Visual grounding} is the task of localizing within images specific objects from text descriptions.
The major instances of visual grounding tasks are \emph{referring segmentation} that produce pixel-level predictions for one \cite{hu2016segmentation,ding2023vlt, wang2022cris} or multiple target objects
\cite{liu2023gres} given a text description, and \emph{referring expression comprehension} \cite{chen2018real,deng2021transvg,liao2020real,liu2023grounding} that detects objects.
Similarly to referring segmentation, we aim to segment specific user-defined objects. In contrast, we do not use supervision to align textual descriptions with object masks and do not focus on text-described relations between objects and mine \ccword s to disentangle target objects from the background.
\section{Open-world open-vocabulary segmentation with test-time contrastive concepts}

We consider the following segmentation task: given an image and a set of textual queries characterizing different visual concepts, the goal is to label all pixels in the image corresponding to each concept, leaving out unrelated pixels, if any. 
Moreover, we want to do so without any prior knowledge of the concepts that could be prompted at the test time.
We do not only want to be \emph{open-vocabulary} in terms of the choice of words for querying, but we also want to be \emph{open-world}, not specialized in a given domain or set of categories. %,
For evaluation purposes, segmenting a specific dataset thus shall not assume anything about the dataset, such as knowledge of represented classes.

\subsection{Introducing the use of test-time contrastive concepts}
\label{sec:cc}

\paragraph{Closed-world vs open-world open-vocabulary semantic segmentation.}

Even when it is open-vocabulary, traditional semantic segmentation is \emph{closed-world} in the following sense. Given an RGB image $\image \,{\in}\, \Reals^{H \times W \times 3}$ and a set of textual queries $\query \,{\in}\, \Queries$, semantic segmentation produces a map
$\closewseg : \{1...H\} \times \{1...W\} \mapsto \Queries$,
where each image pixel has to be assigned one of the queries as a label.
In contrast, \emph{open-world} segmentation considers an additional dummy label `$\rest$' to represent any visual concept that is different from the queries. The segmentation map, in this case, is then 
$\openwseg : \{1...H\} \times \{1...W\} \mapsto \Queries \cup \{\rest\}$
. For instance, to label a boat, it is enough to ask for the ``boat'' segment; other pixels (sky, sea, sand, rocks, trees, swimmers, etc.) are expected to be labeled~$\rest$ and thus ignored.

In the following, we show how to use any open-vocabulary segmenter in an open-world fashion. 
We only assume that the segmenter uses a CLIP-like architecture with a text encoder, noted $\tenc(\cdot)$, used to extract textual features $\tenc(\query)\,{\in}\,\Reals^{d}$ for any query $\query$, where $d$ is the feature dimension. 
Patch-level features $\venc(\image) \in \Reals^{h \times w \times d}$ are generated using the visual encoder, noted $\venc(\cdot)$, 
where 
$h\,{=}\,H / P$, $w\,{=}\,W / P$, and $P$ is the patch size. 
The cosine similarities between each query feature and a patch feature are then used as 
logits when upsampling to 
obtain pixel-level predictions. 
It yields a closed-world segmentation, given our definition above.

From such segmentation, open-world segmentation could be derived by assigning a pixel (or patch) to a query if the cosine similarity between the visual and query embedding is above a given threshold. 
However, in practice, it has been commonly observed that the CLIP space is not easily separable~\cite{miller2024open}, thus making the definition of such a threshold difficult without overfitting the query or datasets~\cite{bousselham2023gem, cha2022tcl}. We further discuss the separability of CLIP patch features in \cref{app:separability_disc}.

\paragraph{Train-time contrastive concepts.}
Cues to separate visual concepts without supervision primarily come from data where these concepts occur separately and are described in their captions. If some concepts always co-occur, they are harder to be told apart.
This applies in particular to OVSS models trained only from captioned images rather than from dense information. 
Sharing a caption pushes their embedding to align on a common textual feature, which in turn tends to bring the visual embeddings closer together.
Still, such frequently co-occurring visual concepts can often be separated
in a closed-world setting:
pixels (or patches) are then just mapped to the query with which they align the most.
However, a problem arises if a visual concept of a query $\query$ can be mistaken for another visual concept present in the image but not queried (e.g., querying ``boat'' but not ``water'' as in \cref{fig:teaser}).

\paragraph{Test-time contrastive concepts.}

To address this problem, we propose to use one or more additional textual queries of visual concepts that are likely to contrast well with~$\query$. For example, when querying ``boat'', we want to add the query ``water''. We name such queries \emph{test-time contrastive concepts} and note them $\CC_\query$. We further propose different solutions to automatically generate $\CC_\query$, and such without assuming prior access to the image domain.
Given prompt queries $\{\query\} \cup \CC_\query$, we perform closed-world segmentation and assign to the dummy label~$\bot$ any patches that are labeled $\CC_\query$.

\paragraph{Multi-query segmentation.}

This principle can be generalized to several simultaneous queries~$\Queries$, with $|\Queries| \,{>}\, 1$, considering the union of their contrastive concepts $\CC_\Queries = \bigcup_{\query \in \Queries} \CC_\query$. Open-world multi-query segmentation consists in segmenting $\Queries \cup \CC_\Queries$, and ignoring pixels not assigned to the queries in $\Queries$, as in the single-query case.
However, some queries in $\Queries$ may already contrast with each other, which puts them in competition with the set of contrastive concepts $\CC_\Queries$ and could lead to their elimination when pixels labeled in $\CC_\Queries$ are discarded. To prevent it, we propose to exclude contrastive concepts $\CC_\Queries$ that are too similar to queries $\Queries$, e.g., with a cosine similarity of text features above some threshold $\beta$: $\CC_\Queries = \bigcup_{\query \in \Queries} \{\query' \in \CC_\query \mid \tenc(\query')\cdot\tenc(q) \leq \beta\}$.
In the following, for simplicity, we only consider the single-query scenario, where $|\Queries| \,{=}\, 1.$

Moreover, to our knowledge, none of the evaluation benchmarks currently used for OVSS allows us to measure the effectiveness of such \cc. 
%To that end, 
We therefore propose a variant of the traditional evaluation metric for semantic segmentation and discuss it in detail in~\cref{sec:metric}.

\subsection{Contrasting with ``background'' \texorpdfstring{$(\CCB)$}~}
\label{sec:background}

In recent work \cite{clippy2022, wysoczanska2023clipdiy, wysoczanska2023clipdino}, the word ``background'' has been used to try to capture a generic visual concept to help segment foreground objects, separating them from their background. In our framework, it amounts to defining ``background'' as a test-time contrastive concept to any query~$\query$. In other words, it defines $\CCB_\query = \{\text{``background''}\}$.

However, if the word ``background'' feels natural to us, it is not obvious why it should also make sense in the CLIP space. This formulation is not contextual, meaning that the contrastive concept is not specific to the query, which might be suboptimal. % 
Worse, the ``background'' samples from which CLIP learned could accidentally include the visual concept of the query, making the query representation close to the background representation and defeating the contrast mechanism.

\begin{figure}[ht!]
    \centering
    % \fbox{
    \begin{subfigure}[b]{0.32\textwidth}
    \centering
    \resizebox{0.95
    \linewidth}{!}{
    \begin{tikzpicture}{every node/.append style={font=\scriptsize}}
        \begin{axis}[
         width=5cm, % changed size a bit
            height=6cm,
            xbar, % not xbar interval
            bar width=9pt,
            xmin=0, 
            xmax=0.01, % reduced this a bit
            % xlabel = Interest,
            axis x line=bottom,
            axis y line=left,
            enlarge y limits=0.14, % no need to enlarge x limits
            xtick={0.0,0.003, 0.006, 0.009},
            symbolic y coords={background,bicycle,bird,boat,car,chair,horse,person,sofa},
            ytick = data,
        %    y tick label style={rotate=45, anchor=east}, % I wouldn't use this
            xticklabel style={
              /pgf/number format/fixed % no scientific notation for the smallest values
            }
        ]
            \addplot[fill=purple,purple] coordinates { 
        (0.007837898,background) 
        (0.000443042,bicycle) 
        (0.001159409,bird)
        (0.000921201,boat)
        (0.004865769899536321,car) 
        (0.0018899729520865532,chair) 
        (0.0008683249613601236,horse) 
        (0.0006233215803709429,person) 
        (0.0007366837326120557,sofa)
     };
        \end{axis}
    \end{tikzpicture}
    }
        
    \caption{Freq. of VOC concepts.}
    \label{fig:background-freq}
    \end{subfigure}
    \hfill
    \begin{subfigure}[b]{0.32\textwidth}
        \centering
        \renewcommand{\arraystretch}{0.}
        \setlength{\tabcolsep}{0.pt}
        \begin{tabular}{ccc}
            \includegraphics[width=0.32\textwidth, height=0.32\textwidth]{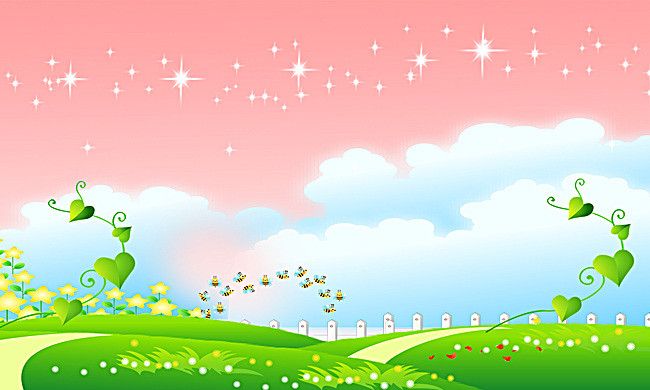} & 
            \includegraphics[width=0.32\textwidth, height=0.32\textwidth]{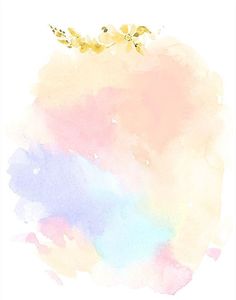} & 
            \includegraphics[width=0.32\textwidth, height=0.32\textwidth]{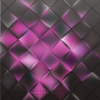}\\ 
            \includegraphics[width=0.32\textwidth, height=0.32\textwidth]{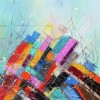} & 
            \includegraphics[width=0.32\textwidth, height=0.32\textwidth]{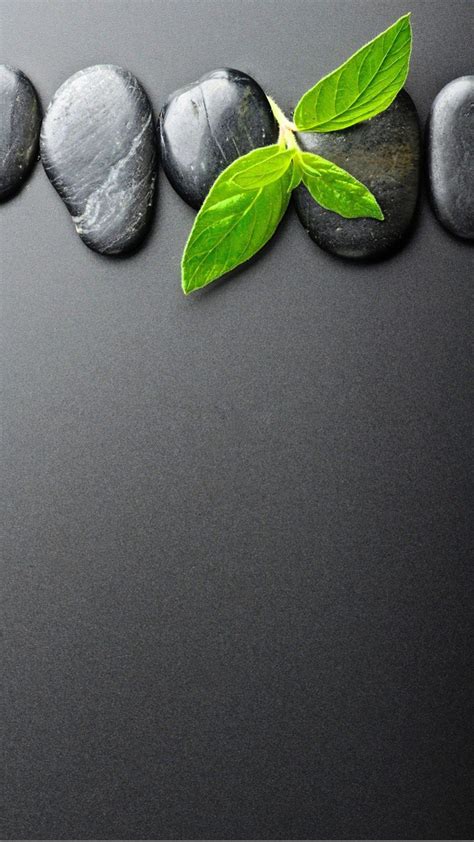} & 
            \includegraphics[width=0.32\textwidth, height=0.32\textwidth]{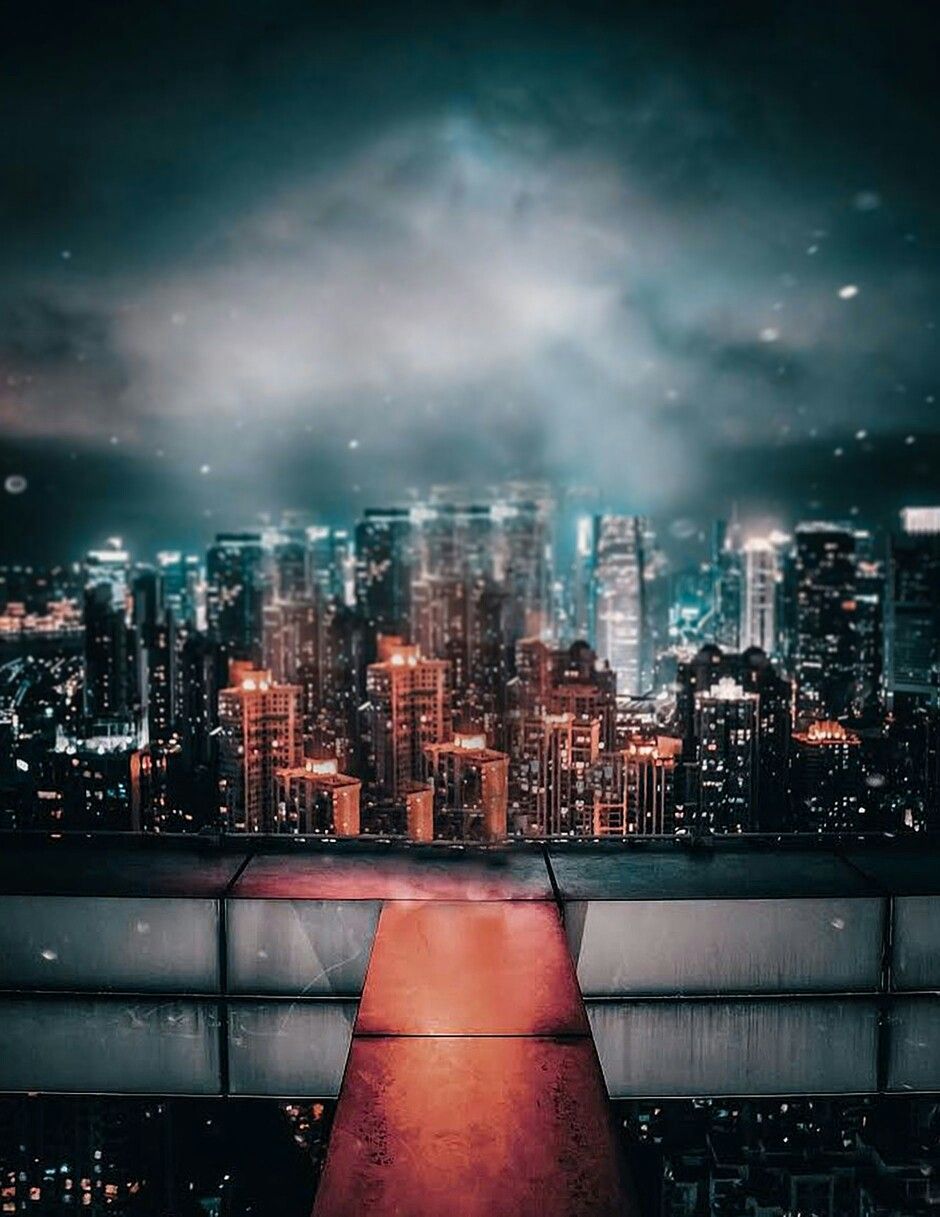} \\
            \includegraphics[width=0.32\textwidth, height=0.32\textwidth]{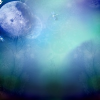} & 
            \includegraphics[width=0.32\textwidth, height=0.32\textwidth]{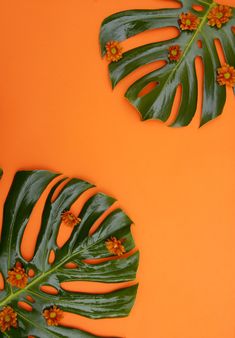} & 
            \includegraphics[width=0.32\textwidth, height=0.32\textwidth]{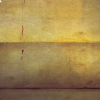} \\
        \end{tabular}
        \caption{``background'' in caption}
        \label{fig:background}
    \end{subfigure}
    \hfill
    \begin{subfigure}[b]{0.32\linewidth}
        \centering
        \renewcommand{\arraystretch}{0.}
        \setlength{\tabcolsep}{0.pt}
        \begin{tabular}{ccc}
            \includegraphics[width=0.32\textwidth, height=0.32\textwidth]{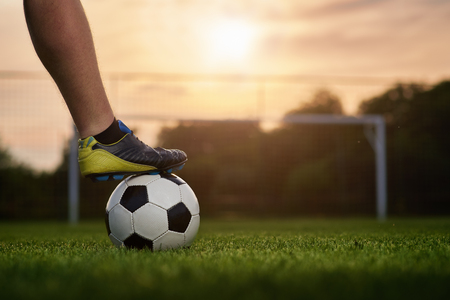} & 
            \includegraphics[width=0.32\textwidth, height=0.32\textwidth]{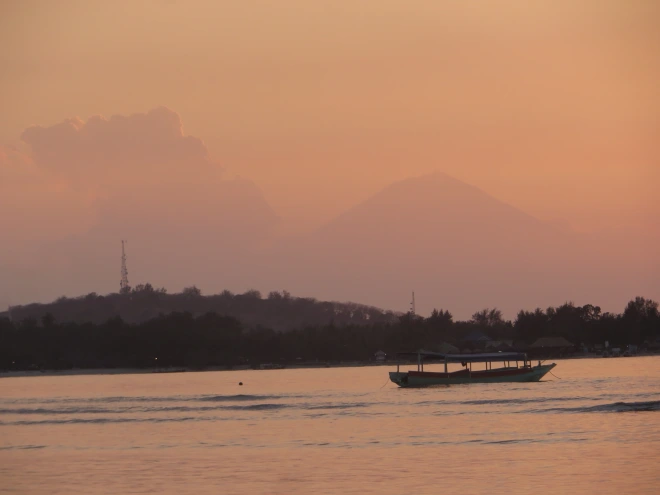} & 
            \includegraphics[width=0.32\textwidth, height=0.32\textwidth]{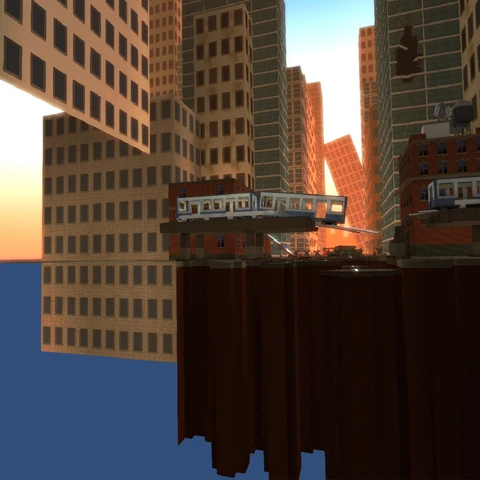}\\ 
            \includegraphics[width=0.32\textwidth, height=0.32\textwidth]{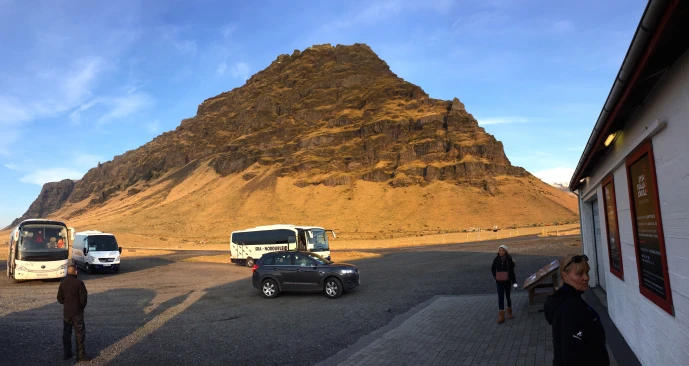} & 
            \includegraphics[width=0.32\textwidth, height=0.32\textwidth]{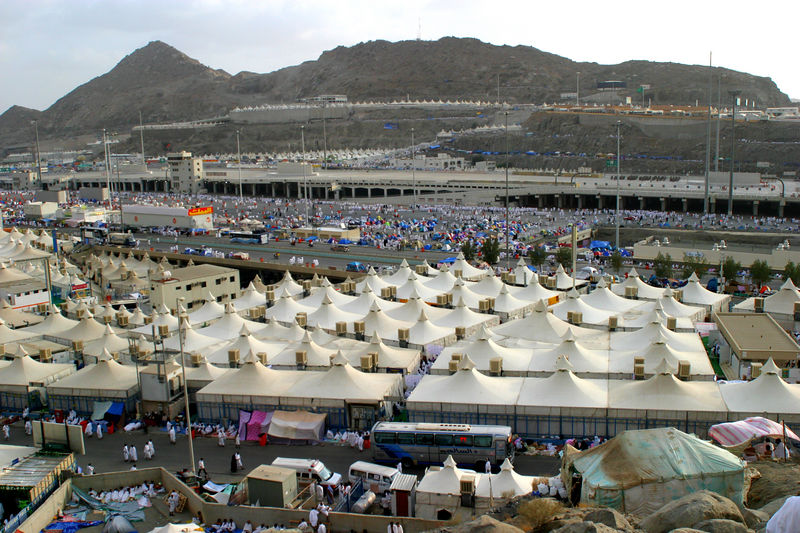} & 
            \includegraphics[width=0.32\textwidth, height=0.32\textwidth]{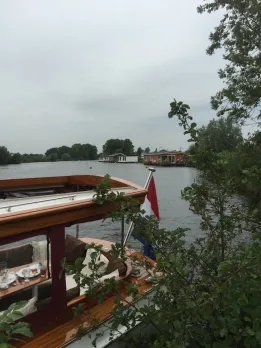} \\
            \includegraphics[width=0.32\textwidth, height=0.32\textwidth]{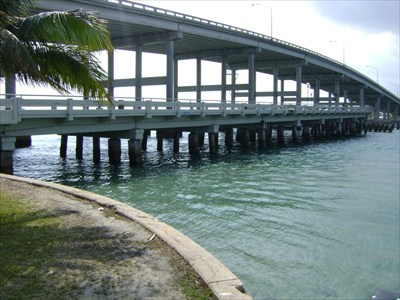} & 
            \includegraphics[width=0.32\textwidth, height=0.32\textwidth]{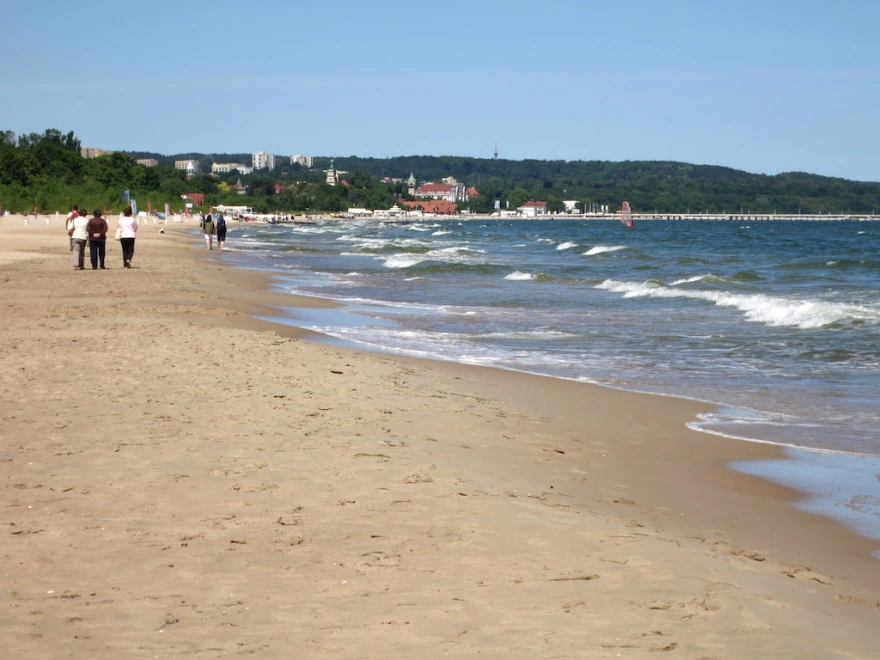} & 
            \includegraphics[width=0.32\textwidth, height=0.32\textwidth]{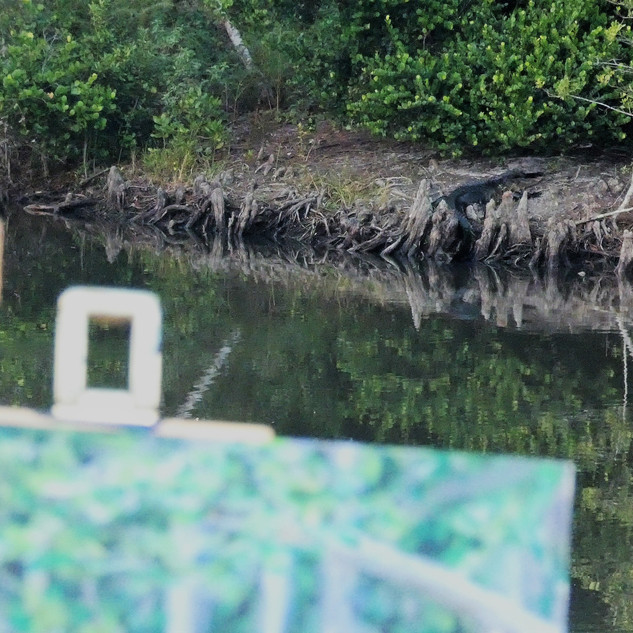} \\
        \end{tabular}
        \caption{``in the background'' in caption}
        \label{fig:inthebackgorund}
    \end{subfigure}
    % }
    \caption{\textbf{Statistics about ``background'' in metadata of web-crawled datasets.} (a) Frequency of some of the concepts from VOC dataset in LAION-400M caption samples. Examples of images in web-crawled data with a caption including the words ``background'' (b) or ``in the background'' (c).}
\end{figure}

We investigate the occurrence of ``background'' in VLM training data to sort it out. First, we use the metadata provided by \cite{udandarao2024zeroshot}, which describes the representation of four thousand common concepts in LAION-400M \cite{schuhmann2021laion}, which is a subset of the web-crawled LAION-2B dataset \cite{schuhmann2022laionb} used to train CLIP. 
In \cref{fig:background-freq}, we plot the frequency of occurrence of ``background'' among other VOC class names. We observe that ``background'' is significantly more frequent than all other words, hinting that it is widely available in CLIP training data and in general web-crawled data.

\cref{fig:background} shows images sampled from the LAION dataset with a caption containing ``background''. We observe that they display a high diversity in colors and textures. Images captioned with ``in the background'' (\cref{fig:inthebackgorund}) appear more photo-oriented. We believe that the combination of a high frequency of the ``background'' word in the dataset and the diversity of associated images make it a good generic contrastive concept and hence make $\CCB$ a baseline. However, superior % state-of-the-art
results have been obtained by applying well-designed tricks to handle the background \cite{wysoczanska2023clipdiy, wysoczanska2023clipdino, cha2022tcl, bousselham2023gem}, emphasizing the necessity of applying something more than simply ``background''.

An option is to define a generic background class list, as done by CLIPpy \cite{clippy2022} or CAT-Seg~\cite{cho2024catseg}, which adds to the concept ``background'' a fixed list of concepts potentially appearing in the background, e.g. ``sky'', ``forest'', ``building'', to be discarded. First, since these visual concepts are intended to be discarded, it would not be possible to query them. Second, such a list is defined at the dataset level, making it domain-specific. As it is impossible to exhaustively describe all visual concepts appearing in any ``background'' (without prior knowledge of the domain or dataset), we propose generating such complements specifically per query, as discussed below.

\subsection{Automatic contrastive concepts \texorpdfstring{$(\CC)$}~ generation} 

\begin{figure}[h]
    \centering
    % \fbox{
    \includegraphics[width=1.\linewidth]{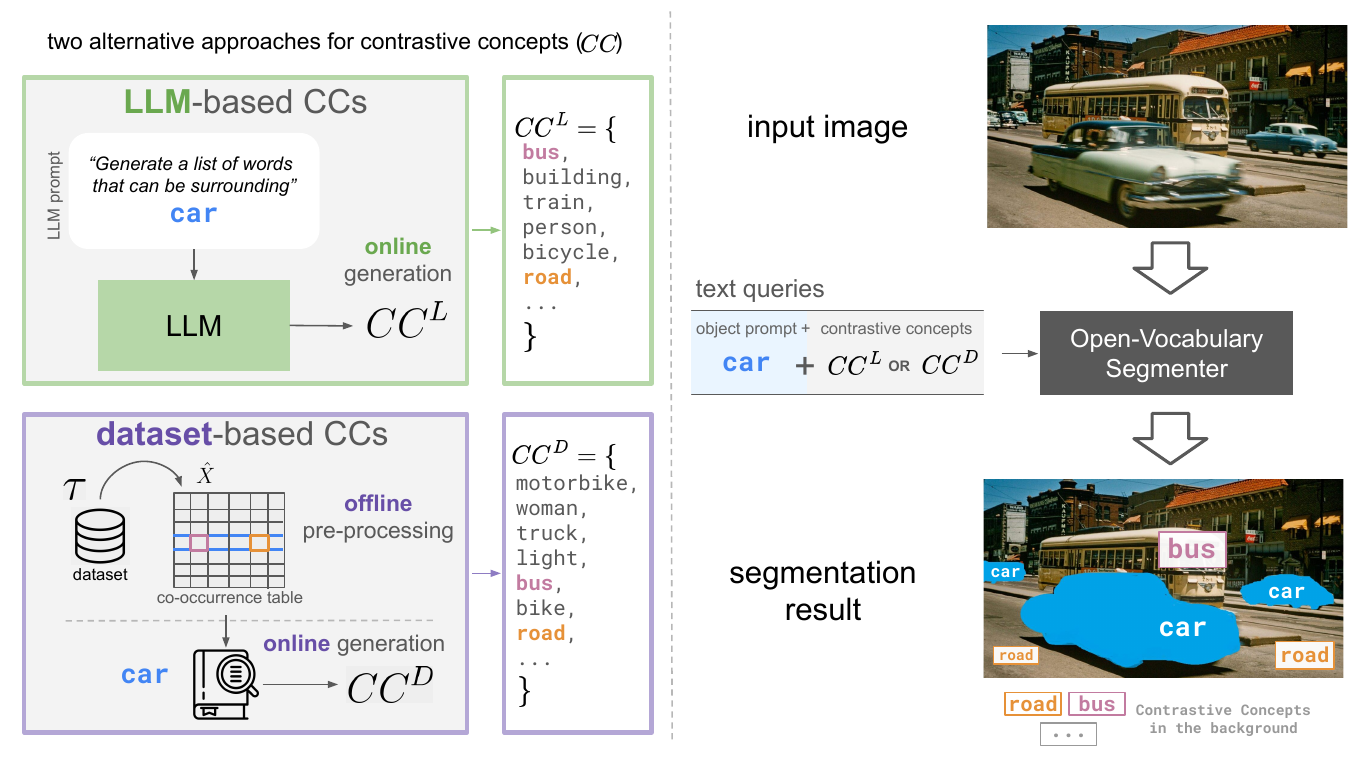}
    % }
    \caption{\textbf{Overview of our method.} We propose two solutions to generate \cc automatically, the first one  (\textit{top-left}) based on LLM prompting (\ccLLM) and the second one, \ccD, that relies on the distribution of co-occurring concepts in a pre-training dataset of a VLM (\textit{bottom-left}). Both methods can be effectively integrated into various open-vocabulary segmentation methods.}
    \label{fig:sup_overview}
    \vspace{-4pt}
\end{figure}

To generate contrastive concepts that are query-specific but also domain-agnostic,
the only data we can then leverage
are (i)~the VLM's training data, or (ii)~unspecific external data. As we focus on text-based contrasts, we can (i)~exploit the large vocabulary of concepts used for VLM training or (ii)~generate prompts via an LLM. Finally, as we want good contrasts, we must find hard negatives. These are concepts that surround queries in images. To gather them, we can (i)~look for word co-occurrences in training data or (ii) ask an LLM to list such concepts. \cref{sec:laion} investigates option~(i), and \cref{sec:ccl}, option~(ii) and \cref{fig:sup_overview} presents a high-level overview of both approaches.

\subsubsection{Mining co-occurrence-based contrastive concepts \texorpdfstring{$(\CCD)$}~} 
\label{sec:laion}
As discussed above,
ambiguity in segmentation for unsupervised approaches arises from co-occurrences in training data. Yet, OVSS does better when prompted to create segments simultaneously for co-occurring concepts.
To list contrastive concepts specific to a given query $\query$, we propose thus to use the information of \emph{co-occurrence} 
in the VLM training captions.
For efficiency, we construct \emph{offline} 
a \emph{co-occurrence dictionary}, built for a large lexicon of textual concepts extracted from the captions. 
We note $\CCD_\query$ the co-occurrence-based contrastive concepts we extract for a query~$\query$ based on this lexicon.

\paragraph{Co-occurrence extraction.} We consider as lexicon a set of textual concepts $\Lexicon$ extracted from captions of the VLM training dataset and construct the co-occurrence matrix $X \in \mathbb{N}^{|\Lexicon| \times |\Lexicon|}$. Concretely, two concepts $\{i,j\} \subset \Lexicon$ co-occur if they appear simultaneously in the caption of an image. $X_{i,j}$ counts the number of times concepts $\{i,j\}$ co-occur in some images. Next, we normalize the symmetric matrix~$X$ row-wise by the number of occurrences of concept~$i$ in the dataset, producing the frequency matrix $\hat{X}$. We then consider only concepts with frequent co-occurrences: for each $i \in \Lexicon$, we select concepts $\Lexicon_i = \{ j \in \Lexicon \mid \hat{X}_{i,j} > \gamma\}$, for some frequency threshold~$\gamma$ 
. 
Selecting only a few contrastive concepts in this way is also consistent with the fact that we target online segmentation: we need to be mindful of computational costs.

\paragraph{Concept filtering.} 
To improve the quality of selected \ccword s\ $\Lexicon_i$, we design a simple filtering pipeline. For each target concept $i \in \Lexicon$ (which can be considered a future query), we remove from $\Lexicon_i$ any concept that might interfere with~$i$ and induce false negatives. First, we discard uninformative words in captions: $\{$``image'', ``photo'', ``picture'', ``view''$\}$. Then, we remove \emph{abstract} concepts, such as ``liberty''. To do so, we ask an LLM whether a given word can be visible or not in an image (more details in~\cref{sec:moredetails:prompts}). We also filter out concepts that are too semantically similar to target concept~$i$, e.g., such that their cosine similarity with $\tenc(i)$ is more than a threshold~$\delta$.
We also consider an alternative approach to filtering, which uses the structured ontology WordNet~\cite{fellbaum1998wordnet} to remove the \cc{}s that possibly interfere with $\query$. However, our experiments, which are discussed in~\cref{sec:wordnetexp}, show that our proposed filtering mechanisms based on dataset statistics are more effective.

\paragraph{Generalization to arbitrary concepts.} So far, we discussed how to select contrastive concepts $\CC^D_i$ for a target concept $i \in \Lexicon$. 
Now, when we are given an arbitrary textual query $\query$, to make the generation of \ccword s truly open-vocabulary, we first find in the CLIP space the nearest neighbor $i$ of $\query$ in $\Lexicon$ and then use for $\query$ the \ccword s of~$i$: $\CC^D_q$ = $\CC^D_i$.

\subsubsection{Prompting an LLM to generate contrastive concepts \texorpdfstring{$(\CCLLM)$}~}
\label{sec:ccl}

\begin{figure}[h]
    \centering
    %\vspace{-8pt}
    \vspace{-4pt}
    \fbox{\includegraphics[width=0.95\linewidth, trim={0.5cm 10.2cm 1.5cm 2.8cm},clip]{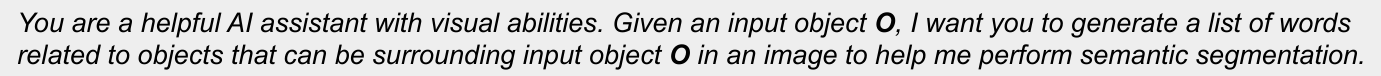}}
    %\vspace{-17pt}
    \vspace{-7pt}
    \caption{\textbf{An abbreviated version of the prompt} we use to generate \ccLLM.}
    \vspace{-8pt}
    \label{fig:prompt}
\end{figure}

Instead of extracting \ccword s from the VLM training set, here we investigate another strategy, generating them using an LLM.
For a given text query~$\query$, we ask an LLM to directly generate contrastive concepts $\CCLLM_\query$, without the need for subsequent filtering. To that end, we design a prompt that excludes potential synonyms, meronyms (e.g., ``wing'' for ``plane''), or possible contents (e.g., ``wine'' for ``bottle''). 
We present a shorter version of the prompt in~\cref{fig:prompt} and include the complete version in~\cref{sec:moredetails:prompts}.

Using an LLM has the benefit of producing specific \ccword s $\CC_\query$ for any target query $\query$, without returning to a fixed and practically limited lexicon. 

\section{Evaluation}

\subsection{Evaluating open-world segmentation}
\label{sec:exp:eval}

We discuss here our evaluation protocols and present our new metric \metric specifically designed to evaluate open-world segmentation.

\paragraph{Evaluation datasets.}
We conduct our experiments on six datasets widely used for the task of zero-shot semantic segmentation \cite{cha2022tcl}, fully-annotated COCO-Stuff \cite{caesar2018cocostuff},
Cityscapes \cite{cordts2016cityscapes} and ADE20K \cite{zhou2019ade20k} and object-centric VOC \cite{pascal-voc-2012}, COCO-Object \cite{caesar2018cocostuff} and Context \cite{mottaghi_cvpr14}, when considering ``background'' pixels. 
We treat the input images following the protocol of \cite{cha2022tcl}, which we detail in~\cref{sec:protocol}.

\paragraph{Our \metric metric.}\label{sec:metric} 

\begin{figure}[h]
    \centering
    % \fbox{
    \includegraphics[width=1.\linewidth,trim={1cm 0 0.5cm 0},clip]{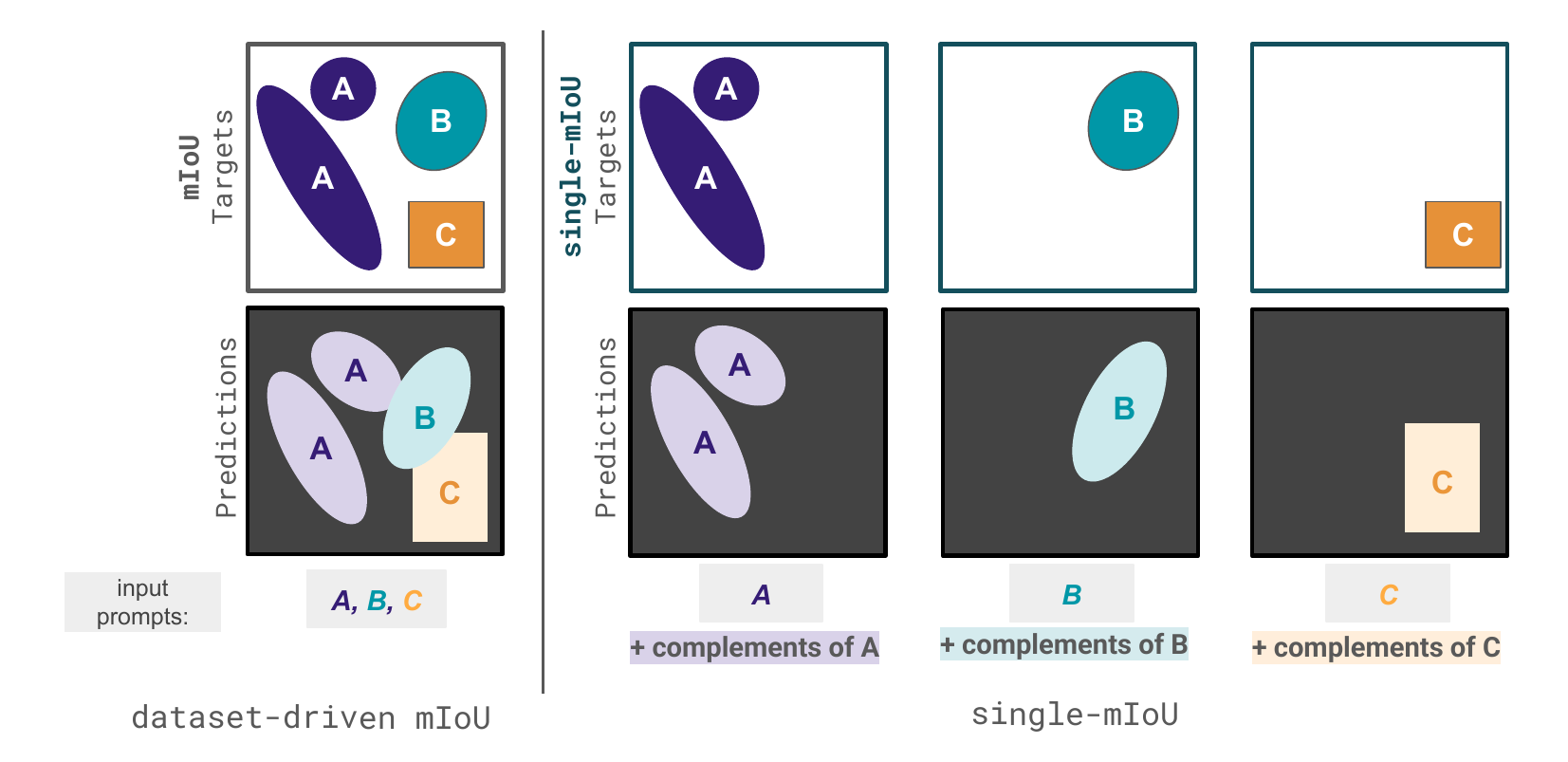}
    % }
    \caption{\textbf{Illustration of \metric metric}. We show the difference with the standard mIoU metric (dataset-driven mIoU), where all the concepts present on an image are considered at once. On the contrary, our \metric considers each of the present concepts separately to measure the single-class segmentation ability of open-vocabulary semantic segmenters.}
    \label{fig:metric}
\end{figure}

To better evaluate the ability
of a method to localize a visual concept when given \emph{no other information}, we propose the \metric metric. It modifies the classic IoU by considering each concept independently and then averaging.
Concretely, we individually segment each class annotated in the dataset for the considered image, thus with $|\Queries| \,{=}\, 1$. 
The \metric is then the average of each IoU with the corresponding ground-truth class segment.
We illustrate this metric in~\cref{fig:metric}, and provide its pseudo-code in~\cref{sup:metric}.
If a dataset contains a \emph{background} class, we do not consider it in the mIoU calculation.

\paragraph{Classic mIoU evaluation.}
We also evaluate the impact of using our \cc\xspace in the classic mIoU scenario on the datasets that consider ``background'' as a class, i.e., VOC and COCO-Object. We prompt at once all dataset classes together with their \cc{}s, using our multiple-query strategy discussed in~\ref{sec:cc}. We then assign pixels that fall into any of the \cc{}s to ``background'', ensuring that none of the concepts competes with the dataset queries. It allows us to verify if our \cc{}s can act as background without hurting the performance on foreground classes.

\subsection{Evaluated methods}
\label{sec:exp:setup}

\paragraph{Test-time contrastive concepts.}
For \ccD generation, we use the statistics gathered by \citet{udandarao2024zeroshot} 
for four thousand common concepts in the LAION-400M dataset, which is a subset of LAION-2B \cite{schuhmann2022laionb} and which is used to train CLIP \cite{radford2021learning}. We filter \ccword s using a low co-occurrence threshold~$\gamma=0.01$ and a high CLIP similarity threshold~$\delta=0.8$. In the classic mIoU scenario, we use a threshold $\beta=0.9$ to account for possible similarities between one query and \ccword s close to the other queries. 
We discuss the selection of these values in~\cref{sec:sup_hyperparams}. To generate \ccLLM, we use the recent Mixtral-8x7B-Instruct model \cite{DBLP:journals/corr/abs-2401-04088}. More details about the setup can be found in~\cref{sec:moredetails:LLM} alongside our designed prompts in~\cref{sec:moredetails:prompts}. 
In our experiments, unless stated otherwise, we include ``background'' in all \cc's: $\mathcal{CC}^D \gets \{\text{``background''}\} \cup  \mathcal{CC}^D$ and $\mathcal{CC}^L \gets \{\text{``background''}\} \cup  \mathcal{CC}^L$.

\paragraph{Baselines.} 
\label{sec:baseline}

To evaluate the impact of using contrastive concepts, we experiment on 6 popular or state-of-the-art methods, one of which (MaskCLIP) uses 3 different backbones, thus resulting in 8 different segmenters, which we believe represent the current OVSS landscape. Concretely,
we study two training-free methods that directly exploit the CLIP backbone, namely MaskCLIP \cite{zhou2022maskclip} and GEM \cite{bousselham2023gem}, where MaskCLIP may exploit different OpenCLIP backbones \cite{openclip} pre-trained either on LAION \cite{schuhmann2022laionb}, MetaCLIP \cite{xu2023metaclip}, or by default on the original OpenAI training data % model}
\cite{radford2021learning}.
We also include TCL \cite{cha2022tcl}, CLIP-DINOiser \cite{wysoczanska2023clipdino} and %fully-
supervised methods: CAT-Seg \cite{cho2024catseg} and SAN~\cite{xu2023side}. Details on the evaluation protocol, including background handling strategies, can be found in~\cref{sec:protocol}.
All compared methods use CLIP ViT-B/16. 

\vspace{5pt}
\subsection{Contrastive concepts generation results}
\label{sec:results}

We first present in~\cref{tab:1vsrest-VOC-paper} results obtained with our \metric metric on 3 datasets, namely ADE20K, Cityscapes and VOC. We compare results when using different \cc's~proposed in this work. 
We also include results when having access to privileged information (\ccP), i.e., the list of concepts present in images as given by the evaluation dataset. 
More results can be found in Appendix~\cref{tab:1vsrest-all}.

\setlength{\tabcolsep}{2.5pt}
\newcolumntype{b}{>{\columncolor{blue!15}}c}
%\vspace{8pt}
\begin{table}[ht!]
    \centering
    \resizebox{\linewidth}{!}{
    % \begin{tabular}{lc|a|cccc}
    \begin{tabular}{lc|a@{~~}c@{~~~}c@{~~~}|a@{~~}ccc|a@{~~}ccc}
    \toprule
     & CLIP & \multicolumn{3}{c}{\bf VOC} & \multicolumn{4}{|c}{\bf Cityscapes} 
     & \multicolumn{4}{|c}{\bf ADE20k} \\
      Method & training data 
      % & Original 
       &  \ccB & \ccLLM  & \ccD & \ccB & \ccLLM &
      \ccD & \deemph{\ccP} & 
        \ccB & \ccLLM  & \ccD & \deemph{\ccP} \\
     \toprule

    MaskCLIP & OpenAI %~\cite{zhou2022maskclip}

    & 44.2 & 52.2 & %51.2 & 
    \bf 53.4 &
    % City
    15.0 & \bf 22.5 & 22.0 & \deemph{30.6} &
    %ADE
    20.2 & 23.5 & \bf 25.2 & \deemph{29.8} \\
    DINOiser & LAION-2B %\cite{wysoczanska2023clipdino}
    & 59.3 & 63.1 & \bf 64.7 &
    %City
    23.2 & \bf 30.6 & 27.3 & \deemph{36.0}  &  
    % ADE
    28.9 & 29.7 & %29.5 & 
    \bf 31.6 & \deemph{35.5}\\ %& \bf 31.5 \\
    
    TCL %+ BG handling 
    & TCL's %~\cite{cha2022tcl}

    & 52.9\rlap* & 52.6\rlap* & \bf 53.6\rlap* &
    % Cityscapes
    9.8 & \bf 26.3 & 22.0 & \deemph{29.7} &
    % ADE20k
    % No BG
     14.9\rlap* & 25.9 & \bf 26.5 & \deemph{32.6} \\
    
    GEM & MetaCLIP %~\cite{bousselham2023gem}

    &  48.6\rlap* & 61.3\rlap* & \bf 64.6\rlap* &  %& 59.9 & \bf 60.6 & 
    % Cityscapes
    14.5\rlap* & \textbf{21.5} & 14.6 & \deemph{20.6} & % 20.6 & 21.2\rlap* & \textbf{21.6} & 14.6
    % ADE20k
     21.5\rlap*& 26.3 & \textbf{29.1} & \deemph{33.0} \\ % & 33.0 & 23.4\rlap*& 26.1 & \textbf{29.1} \\

     SAN & OpenAI %~\cite{cho2024catseg}

    & 50.2 & \bf 73.4 & 69.5 & 
    % Cityscapes
    19.9 & \bf 37.6 & 32.0 & \deemph{44.2} &
    % ADE20k
    24.5 & 35.2 & \bf 35.1 & \deemph{42.8}  \\

    CAT-Seg & OpenAI %~\cite{cho2024catseg}

    & 52.8 & \bf 69.5 & 67.7 & 
    % Cityscapes
    -- & -- & -- & -- &
    % ADE20k
    25.7 & 38.4 & \bf 39.7 & \deemph{46.8}  \\

    \bottomrule
    \end{tabular}
    }
    %\vspace{5pt}
    \caption{\textbf{Benefits of \cc{} measured in \underline{\metric}.}
    `*' indicates that the method's original background handling is applied, if any, and provided it gives the best results.
    Note that CAT-Seg input resolution is 640x640, whereas it is 448x448 for all the other methods. We note \ccP~the unrealistic setup where we have access to all of the dataset classes and use them as systematic contrastive concepts (except for VOC, as its annotations do not cover all pixels). Please note that \colorbox{lightgray}{\ccB} is our baseline.} 
    \label{tab:1vsrest-VOC-paper}
\end{table}

\paragraph{``Background'' is not enough.} We start by analyzing the overall impact of our proposed \cc{}s. In all cases, we observe a significant improvement when using
contrastive concepts \ccD\ and \ccLLM\ compared to the \ccB. Even for object-centric VOC where \ccB already provides a strong baseline, our proposed \cc~ generation methods bring significant gains ranging from 0.7 to 16.7 points. Interestingly, test-time \cc{}s also work well for supervised CAT-Seg, showing that our method is beneficial for open-vocabulary segmenters with all levels of supervision.

\paragraph{\ccLLM~ generalize better to domain-specific datasets.}
For both VOC and ADE20K, the co-occurrence-based \ccD\ outperforms most of the time the LLM-based \ccLLM, with a margin ranging from 0.6 to 2.8 points. %  
However, this trend does not hold for Cityscapes, where \ccLLM gives the best results for all methods. In particular, Cityscapes is a dataset of urban driving scenes that contains images depicting a few recurring concepts. 
This may suggest that LLMs can produce better results than \ccD~ for such domain-specific tasks.
We also note that \ccLLM generally produces fewer \cc{}s, but we do not observe a correlation between segmentation performance and |\cc|, as shown in~\cref{app:concepts_vs_performance}.  

\paragraph{Test-time concepts are different from train-time concepts.}
We also observe that \ccP 
results overall do not exceed
50\% mIoU. The segmentation quality might thus be limited by the VLM capacity or by a mismatch between the dataset classes and the training data. Well-designed prompt engineering could help address this issue \cite{roth2023waffling} and improve segmentation results.

\newcommand{\resultsrownobgpaper}[4]{#1 & #2 & #3 & #4}
\begin{wraptable}[9]{R}{4.cm}
    % \vspace*{-2mm}
    \centering
    \resizebox{\linewidth}{!}{
    \begin{tabular}{l|c|cc}
    % \centering
    \toprule
    
    \resultsrownobgpaper
    {Method}
    {Bkg.}
    {Object}
    {VOC}
    \\
    
    \midrule
    
    \resultsrownobgpaper
    {\multirow{3}{*}{MaskCLIP}}
    {\ccB}
    {17.8} % Object
    {35.1} % VOC
    \\ %ADE 
    
    \resultsrownobgpaper
    { }
    {\ccLLM}
    {25.9} % Object
    {46.2} % VOC
    \\ %ADE 
    
    \resultsrownobgpaper
    { }
    {\ccD}
    {25.1} % Object
    {46.4} % VOC
    \\ %ADE 
    
    \midrule

    \resultsrownobgpaper
    {\multirow{3}{*}{GEM}}
    % {GEM}
% {threshold}
{threshold}
% {MetaCLIP}
{27.4} % Object
{46.6} % VOC
\\ %ADE 

\resultsrownobgpaper
% {GEM}
% {\cc}
{}
{\ccLLM}
% {MetaCLIP}
% {-}
{\bf 35.7 } % Object
{60.0} % VOC
\\ %ADE 

\resultsrownobgpaper
% {GEM}
{}
{\ccD}
% {MetaCLIP}
% {-}
{35.5} % Object
{\bf 60.5} % VOC

    % \resultsrownobgpaper
    % {\multirow{3}{*}{DINOiser}}
    % {sal.}
    % {34.8} % Object
    % {62.1} % VOC
    % \\ %ADE 
    
    % \resultsrownobgpaper
    % {}
    % {\ccB}
    % {29.5} % Object
    % {54.0} % VOC
    % \\ %ADE 
    
    % \resultsrownobgpaper
    % {}
    % {\ccLLM}
    % {35.0} % Object
    % {60.8} % VOC
    % \\ %ADE 
    
    % \resultsrownobgpaper
    % {}
    % {\ccD}
    % {33.3} % Object
    % {60.4} % VOC
    
    % \resultsrownobgpaper
    % {\multirow{3}{*}{DINOiser}}
    % {sal.}
    % {34.8} % Object
    % {62.1} % VOC
    % \\ %ADE 
    
    % \resultsrownobgpaper
    % {}
    % {\ccB}
    % {29.5} % Object
    % {54.0} % VOC
    % \\ %ADE 
    
    % \resultsrownobgpaper
    % {}
    % {\ccLLM}
    % {35.0} % Object
    % {60.8} % VOC
    % \\ %ADE 
    
    % \resultsrownobgpaper
    % {}
    % {\ccD}
    % {33.3} % Object
    % {60.4} % VOC
    \\ %ADE 
    \bottomrule
    \end{tabular}
    }
    \vspace*{-2mm}
    \caption{
    % \textbf{Results w/ mIoU.} 
    \textbf{mIoU results.} 
    }
    \label{tab:miou}
    \vspace*{-6mm}
\end{wraptable}

\paragraph{Classic mIoU evaluation.} 

Additionally, in~\cref{tab:miou}, we present results with the standard mIoU for MaskCLIP (with LAION-2B backbone) and GEM.
We report results with various contrastive concepts (\cc) and the original background handling strategy when applicable. We observe that in all cases, the results with \ccD\ and \ccLLM\ are better than baseline \ccB. We also notice that for GEM the results are better than when applying the background handling strategy originally proposed in ~\cite{bousselham2023gem}. This shows that integrating our contrastive concepts does not hurt or can even improve performance in the classic mIoU setup.
We provide more results in~\cref{table:miou_all} in Appendix.

\subsection{Ablation studies}
\label{sec:ablate}

%\vspace{8pt}

\begin{table}[ht!]
\centering
\setlength{\tabcolsep}{1.8pt}
\begin{subtable}[t]{0.33\linewidth}
    \centering
    \resizebox{\linewidth}{!}{
    \begin{tabular}{c|c|c|ccc}
        \toprule
         \emph{co-} & \emph{no} & \emph{sem.} & Mask & TCL & DINO
         % & GEM\\ 
         \\
         \emph{occ.} & \emph{abs.} & \emph{sim.} & CLIP &  & iser 
         % & GEM\\ 
         \\
        \midrule
         \cmark &  & &  20.2 &  22.4 & 23.9
        % 22.7 \\
        \\
         \cmark & \cmark & & 20.9 & 23.2 & 25.5
                 \\
                  & \cmark & \cmark & 18.4 & 20.0 & 26.3\\
        % & 23.6 \\
        \cmark & \cmark & \cmark &  \bf 25.2 & \bf 26.0 & \bf 31.6
        \\
        % \textbf{28.9} \\
        \bottomrule
    \end{tabular}
    }
    \caption{\textbf{Impact of filtering in \ccD} on ADE20K (\%\metric).}
    \label{tab:abl-filtering}
\end{subtable}
\hfill
\begin{subtable}[t]{0.325\linewidth} 
    \centering
    \resizebox{\linewidth}{!}{
    \begin{tabular}{l|cc|cc}
        \toprule
          & \multicolumn{2}{c|}{\relax{Cityscapes}} & \multicolumn{2}{c}{\relax{ADE20k}} \\ 
         Method & 
         w/o & w/ & w/o & w/ 
         \\
          \midrule
         MaskCLIP & 22.3 & \bf 22.5 & 22.5 & \bf 23.5\\ % OpenAI version
         DINOiser & 30.3 & \bf 30.6 & 27.5 & \bf 29.7 \\
         TCL  & 26.0 & \bf 26.2 & 25.4 & \bf 26.3 \\ 
         GEM  & 21.3 & \bf 21.4 & 25.7 & \bf 26.1 \\ 
        \bottomrule
    \end{tabular}
    }
    \caption{\textbf{Adding ``background'' or not} to our LLM-based \ccLLM.}
    \label{tab:abl-bkg}
\end{subtable}
\hfill
\begin{subtable}[t]{0.315\linewidth} 
    \centering
    \resizebox{\linewidth}{!}{
    \begin{tabular}{l|@{\hspace{0.5em}}c@{\hspace{0.5em}}c@{\hspace{0.5em}}c}
    
    \toprule
    MaskCLIP & \multicolumn{3}{c}{VOC}  \\
    w/ CLIP &   \\[-1mm]
    training set & \ccB & \ccLLM  & \ccD  \\
    \toprule
    LAION-2B  & 47.9 & 51.8 & \bf 53.8 \\
    OpenAI  & 44.2 & 52.2 & \bf 53.4 \\
    MetaCLIP  & 46.8 & \bf 50.6 & 50.0 \\
    \bottomrule
    \end{tabular}
    }
    \caption{\textbf{Impact of pre-training dataset} on VOC (\%\metric). % \textbf{Different backbones}.
    }
    \label{tab:ablation_maskclip}
\end{subtable}
\caption{\textbf{Ablation studies.}
(a) % We study the 
The impact of filtering steps: `co-occ.' is the co-occurrence-based filtering; `no abs.' is the removal of abstract concepts; `sem. sim.' is the semantic-similarity filtering.
% when considering different methods on the ADE20K dataset.  We note as 'co-occ.' the co-occurrence-based filtering, \emph{abstract} concept removal as 'abstract' and \emph{semantic similarity} filtering 'sem. sim.'.
(b) Relevance of adding ``background'' to \ccLLM. (c) Varying the pre-training dataset.}
%\renaud{(b) Impact of combining the LLM contrastive concepts with word ``background''.}
%\renaud{(c) Performance of MaskCLIP on VOC with different CLIP pre-training datasets.}
% (b) We ablate MaskCLIP~\cite{zhou2022maskclip} performance with different CLIP pre-training datasets on VOC.
\vspace{-10pt}

\label{tab:abl}
\end{table}

\paragraph{\ccD\ concept filtering.}
In~\cref{tab:abl-filtering}, we analyze the impact of the different filtering steps discussed in~\cref{sec:laion} on the challenging ADE20K dataset. We observe that each step boosts
results by removing noisy or detrimental concepts. 
The largest gain is obtained when filtering highly similar (`sem.\,sim.') concepts. % 
We also note that the improvement is consistent for all methods. 
We report the performance without the
co-occurrence thresholding (w/o `co-occ.') and observe a significant degradation. 
More experiments in \cref{sec:wordnetexp} suggest that ontology-based filtering (e.g., using WordNet) does not help and can even be harmful. %

\paragraph{Adding ``background'' to \ccLLM.}
In~\cref{tab:abl-bkg}, we study the influence of adding the word ``background'' to the set of contrastive concepts \ccLLM\  generated with the LLM.
We observe that it is always beneficial, in most cases with little gain, except on ADE20k where the gain is up to 2.2 \metric pts.

\paragraph{Impact of the pre-training dataset.} % 
\cref{tab:ablation_maskclip} shows the results of MaskCLIP with different
datasets used to train CLIP. We observe that using \ccD always gives a boost over using ``background'' alone (\ccB) across all pre-training datasets, including on the highly-curated MetaCLIP. 
However, we notice that for MetaCLIP, \ccLLM gives even better results, suggesting that leveraging LLMs can also be more profitable with backbones pre-trained on carefully curated datasets. 

\subsection{Qualitative results}
\label{sec:quali}

\begin{figure}[!ht]
    \centering
    \resizebox{\linewidth}{!}{
    \setlength\extrarowheight{-5pt}
    % \fbox{\
    \begin{tabular}{c@{}c@{}c@{}c@{}c@{}c@{}c}
        & GT & \ccB & \ccD (single) & \ccD (all) & \ccLLM (single) & \ccLLM (all) \\

         % other example
         \raisebox{1.75\normalbaselineskip}[0pt][0pt]{\rotatebox{90}{\makecell{\promptstyle{bed}{bed}}}} & \includegraphics[width=0.165\linewidth]{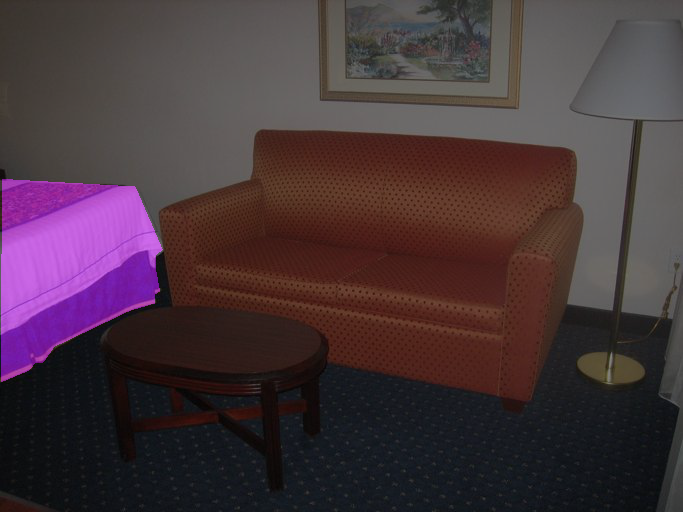} & 
         \includegraphics[width=0.165\linewidth]{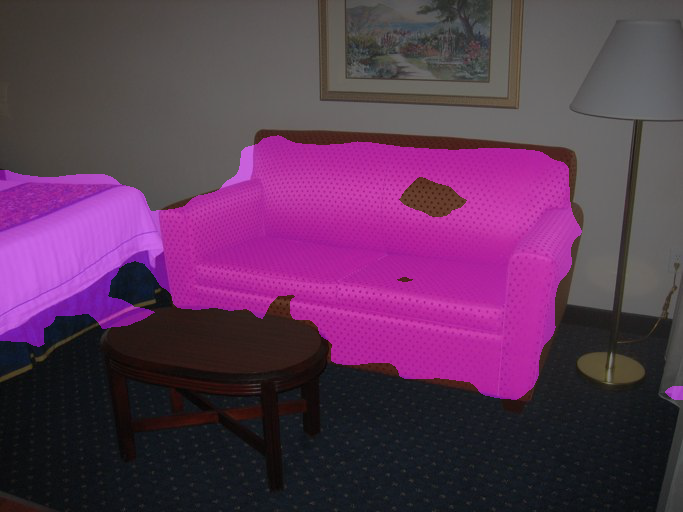} & \includegraphics[width=0.165\linewidth]{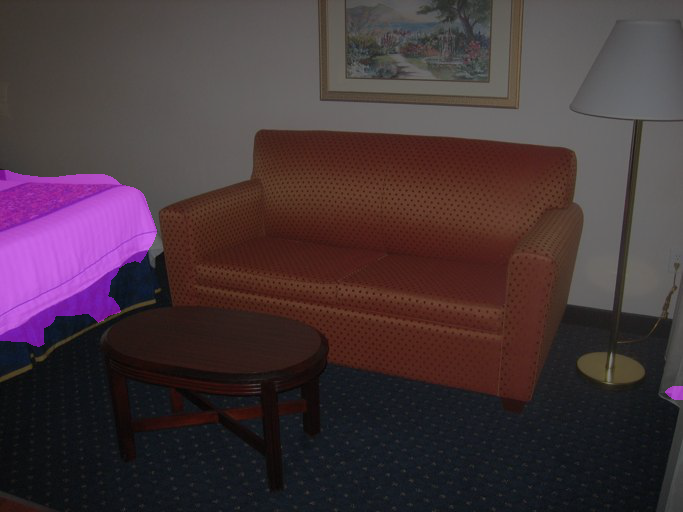} & \includegraphics[width=0.165\linewidth]{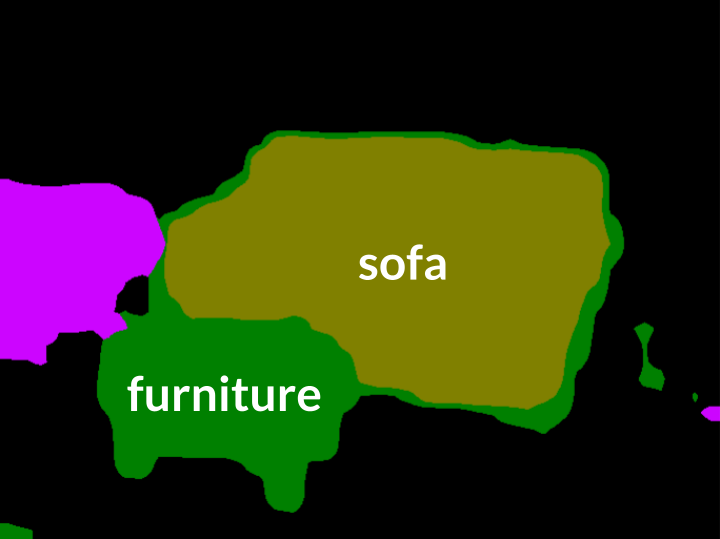} & \includegraphics[width=0.165\linewidth]{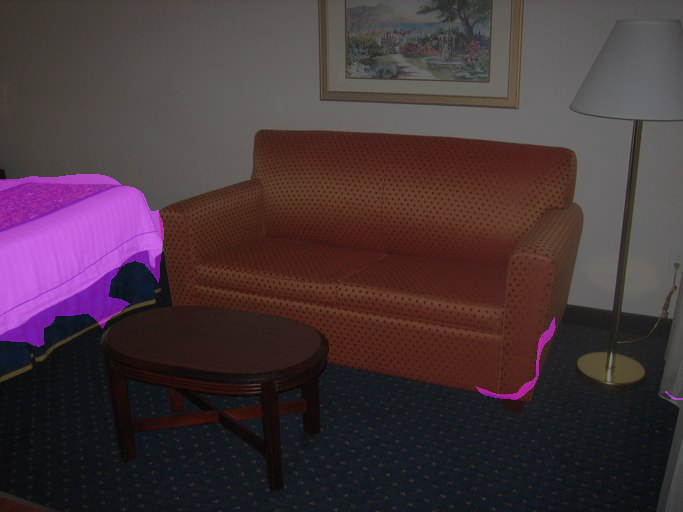} & \includegraphics[width=0.165\linewidth]{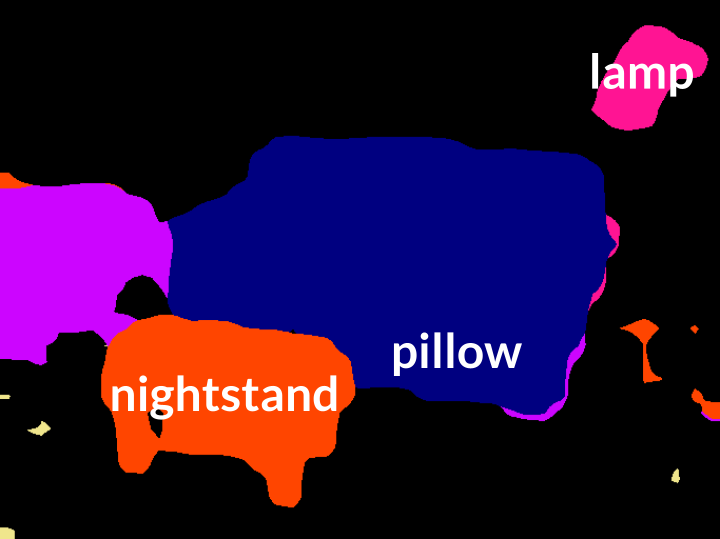} \\
                  % other example
         \raisebox{.9\normalbaselineskip}[0pt][0pt]{\rotatebox{90}{\makecell{\promptstyle{ceiling}{ceiling}}}} & \includegraphics[width=0.165\linewidth]{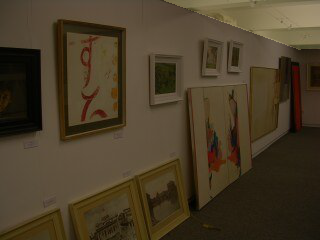} & 
         \includegraphics[width=0.165\linewidth]{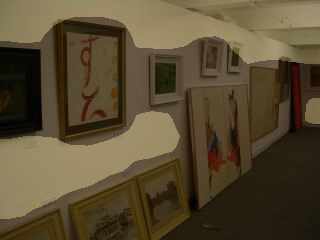} &
         \includegraphics[width=0.165\linewidth]{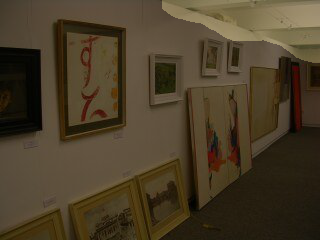} & 
         \includegraphics[width=0.165\linewidth]{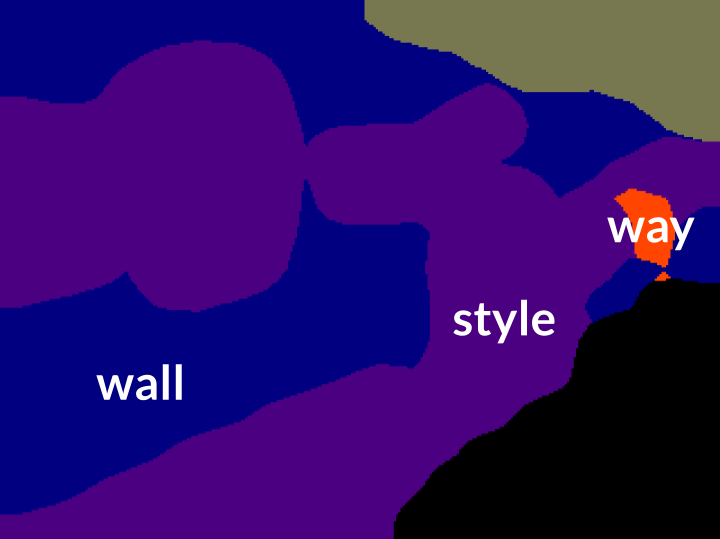} & 
         \includegraphics[width=0.165\linewidth]{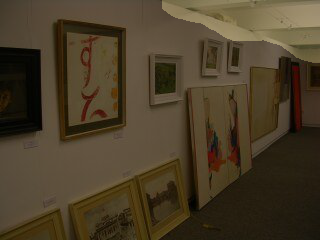} & 
         \includegraphics[width=0.165\linewidth]{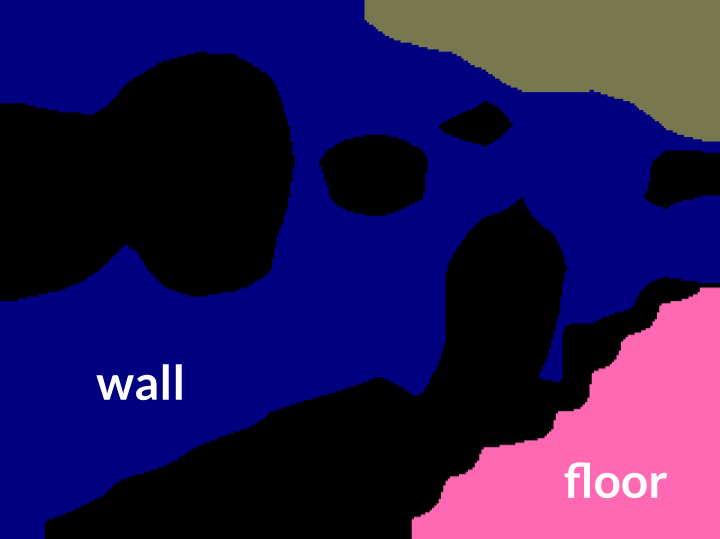} \\
        % other example
         \raisebox{1.3\normalbaselineskip}[0pt][0pt]{\rotatebox{90}{\makecell{\promptstyle{bird}{bird}}}} & \includegraphics[width=0.165\linewidth]{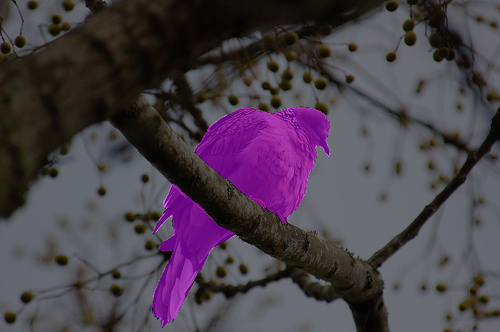} &
         \includegraphics[width=0.165\linewidth]{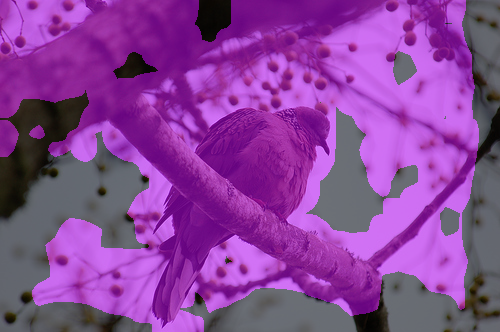} &
         \includegraphics[width=0.165\linewidth]{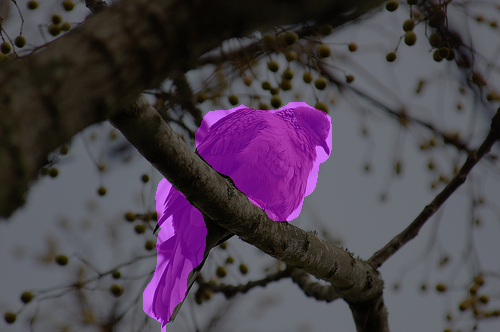} &
         \includegraphics[width=0.165\linewidth]{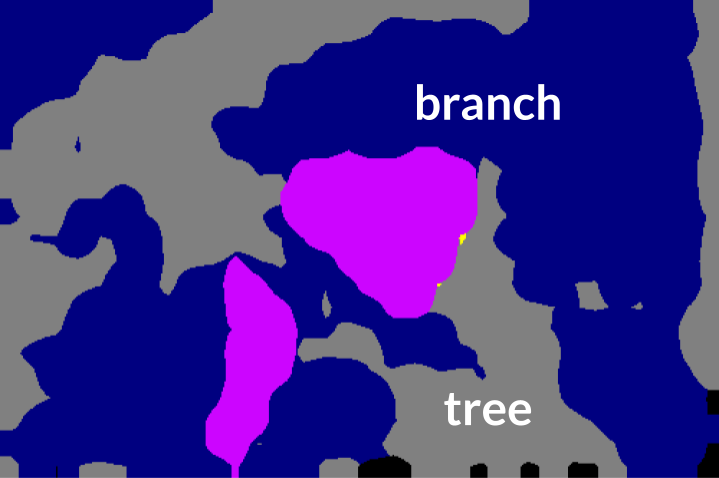} &
         \includegraphics[width=0.165\linewidth]{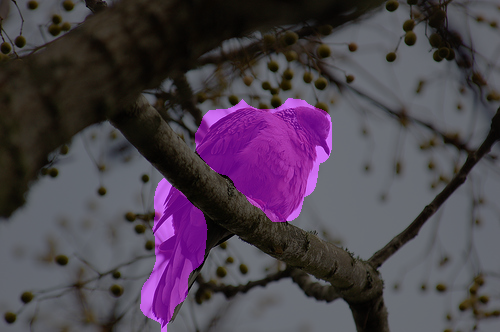} &
         \includegraphics[width=0.165\linewidth]{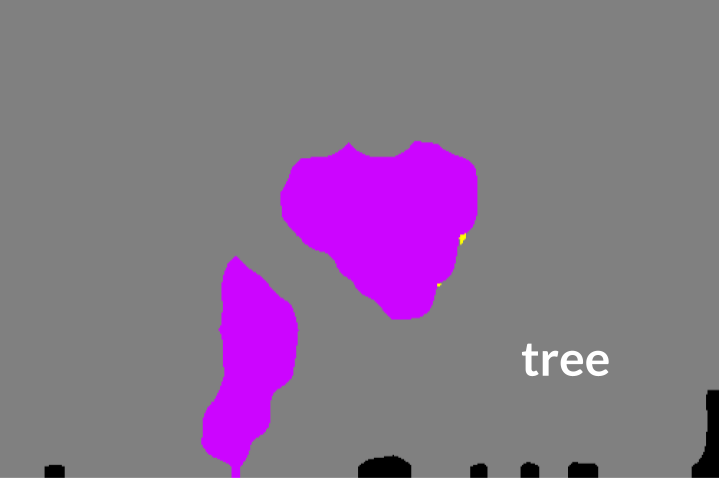} \\
        % other example
         \raisebox{.8\normalbaselineskip}[0pt][0pt]{\rotatebox{90}{\makecell{\promptstyle{aeroplane}{aeroplane}}}} & \includegraphics[width=0.165\linewidth]{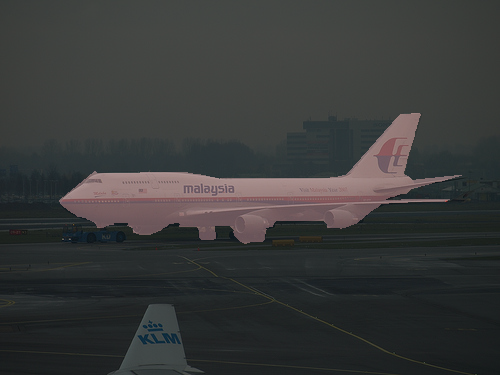} &
         \includegraphics[width=0.165\linewidth]{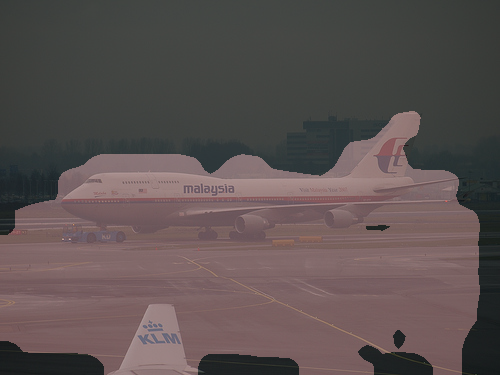} &
         \includegraphics[width=0.165\linewidth]{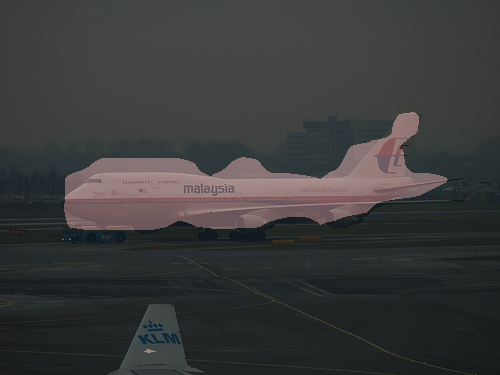} &
         \includegraphics[width=0.165\linewidth]{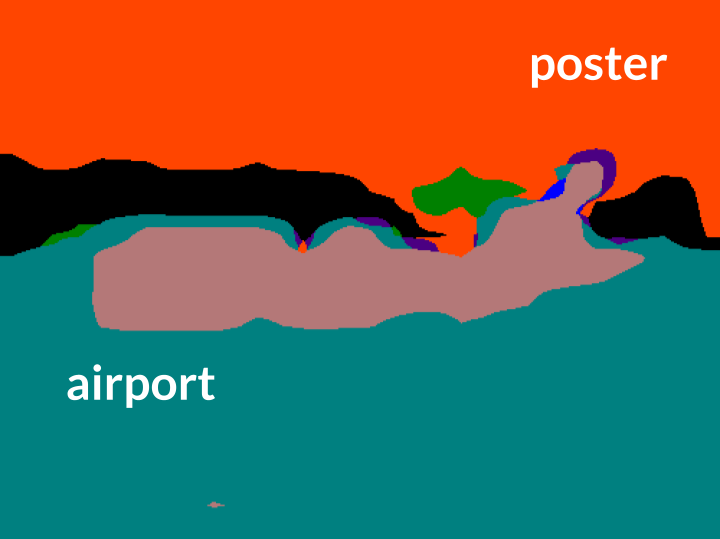} &
         \includegraphics[width=0.165\linewidth]{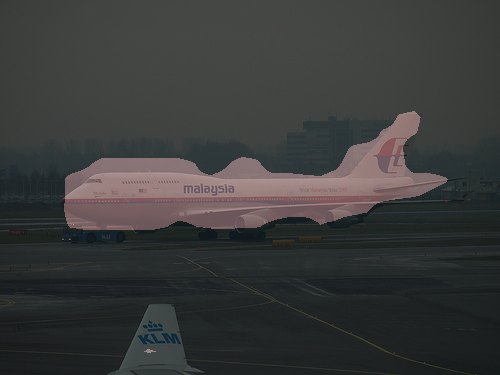} &
         \includegraphics[width=0.165\linewidth]{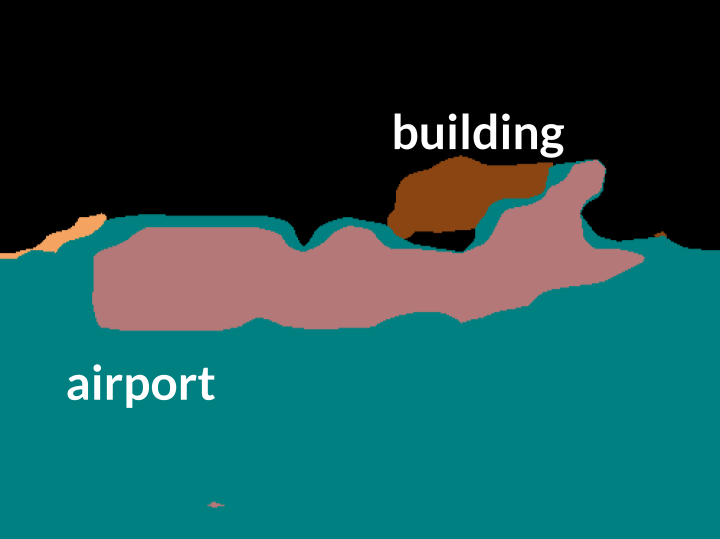} \\
    \end{tabular}
    % }
    }
    % \vspace{-4pt}
    \caption{\textbf{Qualitative results.} We show segmentation examples from ADE20K (1\textsuperscript{st} and 2\textsuperscript{nd} row) and Context (3\textsuperscript{rd} and 4\textsuperscript{th} row), with segments % and the segmentation masks 
    produced by CLIP-DINOiser. For \ccD\ and \ccLLM, we additionally show the joint segmentation of all contrastive classes (all). 
    }
    \label{fig:qualitative}
\end{figure}

In \cref{fig:qualitative}, we present qualitative examples when using different \ccword s proposed in this work. We compare \ccLLM and \ccD with ground truth (GT) and baseline \ccB. 
For both \ccLLM and \ccD, we present the output segmentation mask for the queried concept together with its \ccword s (noted \emph{all}) as well as the single queried concept (noted \emph{single}), where \cc{}s are discarded. 
We observe that the output masks produced by our methods are more accurate, removing the noise from related concepts, e.g. ``tree'' for the bird or ``sofa'' for the ``bed''.

\vspace{-5pt}
\paragraph{Generalization to arbitrary concepts.}
\newlength{\wildwidth}
\newlength\resultsheight
\setlength{\wildwidth}{0.25\textwidth}
\setlength{\resultsheight}{2.6cm}
\newcommand{\addresult}[2]{%
\begin{minipage}[b]{#1}
     \centering
     \includegraphics[width=\textwidth, height=\resultsheight]{figures/examples/#2.png}
\end{minipage}}
\begin{figure}[t]
\centering
\renewcommand{\arraystretch}{0}
\setlength\tabcolsep{0pt} 

% \rowcolors{7}{lightgray}{white}
\begin{center}
% \fbox{
\begin{tabular}{
>{\centering\arraybackslash}m{0em} 
>{\centering\arraybackslash}m{\wildwidth}
>{\centering\arraybackslash}m{\wildwidth}
>{\centering\arraybackslash}m{\wildwidth}
>{\centering\arraybackslash}m{\wildwidth}
% >{\centering\arraybackslash}m{0.22\textwidth}
}
~ & \multicolumn{2}{c}{MaskCLIP} & \multicolumn{2}{c}{CLIP-DINOiser} \\
~ & \ccD (single) & \ccD (all) & \ccD (single) & \ccD (all) \\
% \raisebox{1.5\normalbaselineskip}[0pt][0pt]
~& 
\addresult{\wildwidth}{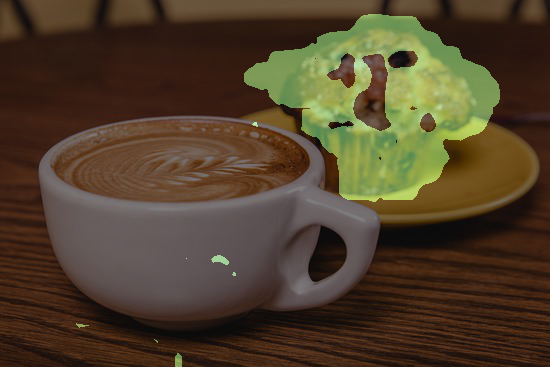} &
\addresult{\wildwidth}{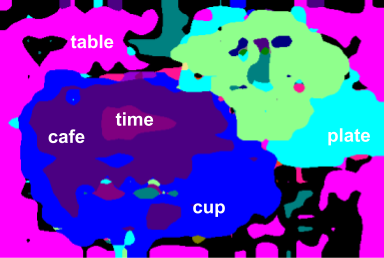} &
\addresult{\wildwidth}{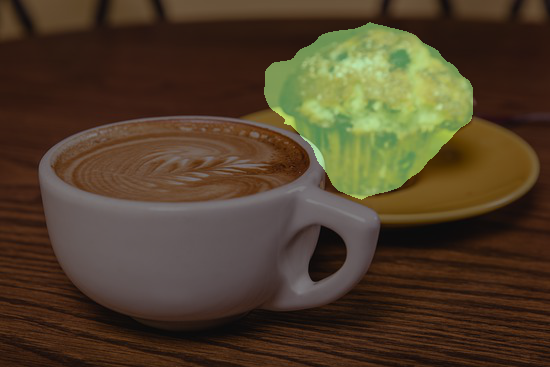} &
\addresult{\wildwidth}{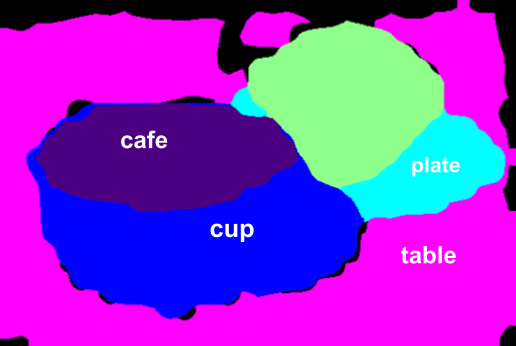}
\\
\multicolumn{5}{>{\centering\arraybackslash}m{1.\linewidth}}{\cellcolor[gray]{0.7}\makecell{$q$: \promptstyle{muffin}{muffin} $\rightarrow$ $i \in \mathcal{T}$: \promptstyle{muffin}{pastry}}}
\\
~ & 
\addresult{\wildwidth}{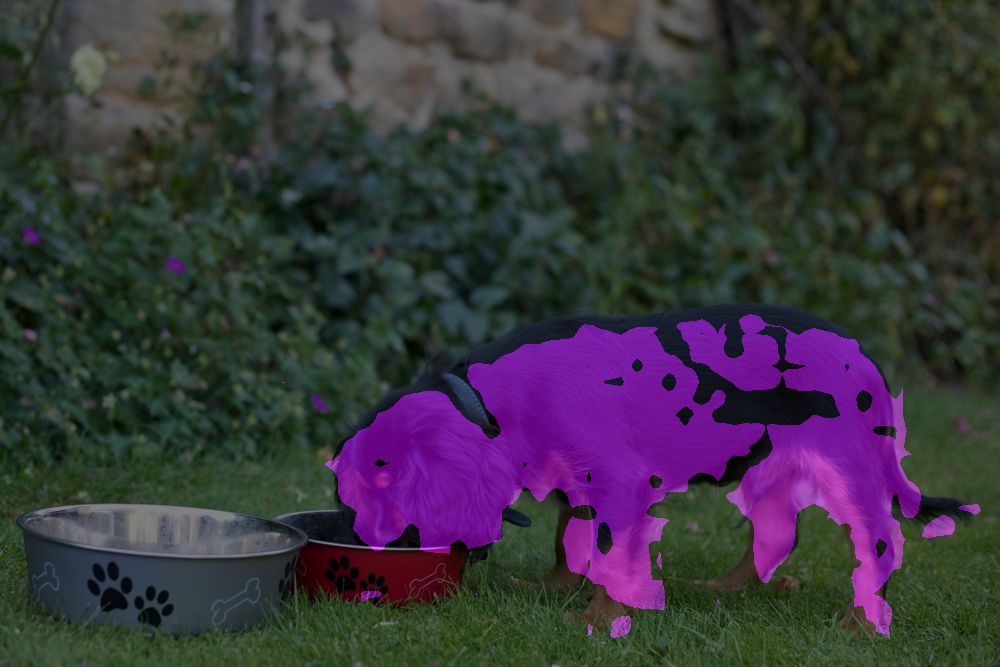} &
\addresult{\wildwidth}{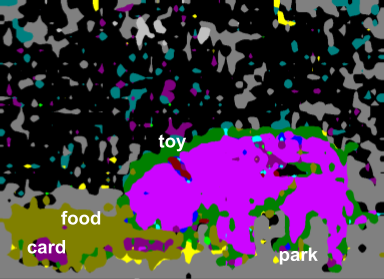} &
\addresult{\wildwidth}{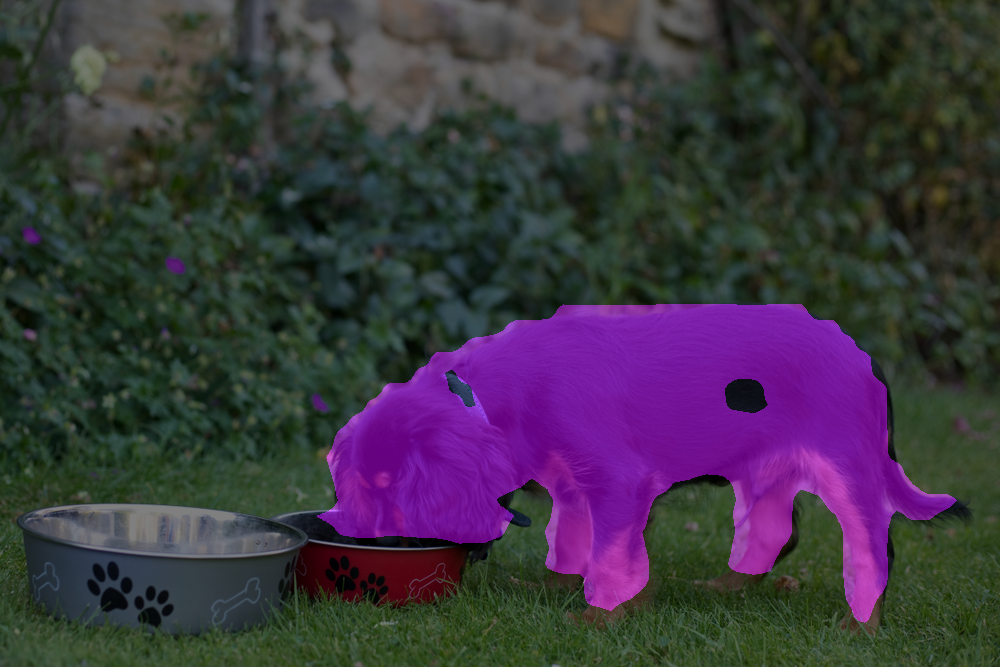} &
\addresult{\wildwidth}{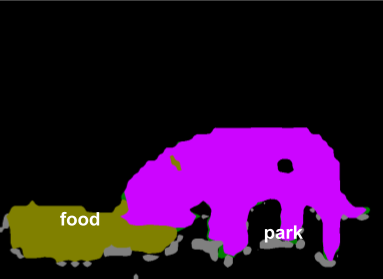}
\\
\multicolumn{5}{>{\centering\arraybackslash}m{1.\linewidth}}{\cellcolor[gray]{0.7}\makecell{$q$: \promptstyle{ bird}{cavalier} $\rightarrow$ $i \in \mathcal{T}$: \promptstyle{bird}{dog}}}
\\
\end{tabular}
% }
\end{center}
    \vspace{-5pt}
    \caption{\textbf{In the wild examples}. We visualize results for MaskCLIP and CLIP-DINOiser for query concepts beyond $\mathcal{T}$. The closest neighbor to a query is presented below each example (grey row).}
    \label{fig:wild}
    % \vspace{-5pt}
\end{figure}
\cref{fig:wild} presents results when prompting queries that are not included in the subset of concepts $\mathcal{T}$ extracted from the VLM training dataset, such as ``muffin'' or ``cavalier'' (a dog breed). We show the closest neighbor for the query $q$ below each example and visualize masks for both MaskCLIP and CLIP-DINOiser. We observe that the \ccD generation method leveraging statistics from pre-training datasets is also robust to examples outside of the co-occurrence dictionary by accurately mapping $q$ to its closest concept in $\mathcal{T}$, e.g., mapping ``cavalier'' to ``dog''.

\section{Conclusion}
\label{sec:conclusion}
In this work, we identify limitations of the current evaluation setup for open-vocabulary semantic segmentation tasks, which are inherited from close-world evaluation benchmarks.
To bridge the gap between closed- and open-world setups, we propose the single-class segmentation scenario. We study the limitations of current state-of-the-art models when we assume no prior access to in-domain classes and propose to automatically discover \ccword s \cc~that are useful to better localize any queried concept.  
To do so, we propose two methods leveraging either the distribution of co-occurrences in the VLM’s training set or an LLM to generate such \cc. Our results show the generalizability of our proposed method across several setups.

\paragraph{Broader Impact Statement.} In this work, we leverage statistics extracted from the training set of CLIP. While Vision-Language Models offer powerful capabilities for visual understanding, their reliance on large-scale internet-scraped datasets introduces significant risks and ethical concerns. These models can perpetuate and amplify societal biases present in their training data.
% For instance, a VLM trained on biased image-text pairs might associate certain professions or attributes with specific demographic groups, leading to unfair treatment in applications like visual search or content moderation. 
Researchers and practitioners must, therefore, carefully consider these ethical implications when developing and deploying VLM-based systems, implementing mitigation strategies, and being transparent about the limitations and potential risks of their applications.

% \clearpage

% Despite its known data-collection problems, CLIP trained with LAION~\cite{schuhmann2021laion} is still widely adopted. Therefore, we also base our analysis on LAION dataset by leveraging the thorough work of \cite{udandarao2024zeroshot} and their metadata.

\section*{Acknowledgments}
This work was supported by the National Centre of Science (Poland) Grant No. 2022/45/B/ST6/02817 and by the grant from NVIDIA providing one RTX A5000 24GB used for this project. We would also like to thank the authors of~\cite{udandarao2024zeroshot} for sharing their metadata.

\bibliography{main}
\bibliographystyle{tmlr}

\appendix
\clearpage
\section*{Appendix}

In this appendix,
\begin{itemize}
\item we start by providing details on the evaluation in Sec.~\ref{sec:protocol}: evaluation protocol (Sec.~\ref{sec:eval_protocol}), approaches to the background handling of the considered baselines (Sec.~\ref{app:back-handling}), and details of the IoU-Single metric (Sec.~\ref{sup:metric}).
\item In Sec.~\ref{sup:quantitative}, we present additional results including classic mIoU results (Sec.~\ref{app:sota_miou}), and further quantitative (Sec.~\ref{app:quantitative_res}) and qualitative (Sec.~\ref{app:qualitative_res}) results. We also discuss failure cases (Sec.~\ref{app:limitations}).

\item Sec.~\ref{app:additional_analysis} presents an additional analysis of our method, particularly: hyperparameter selection (Sec.~\ref{sec:sup_hyperparams}), (Sec.~\ref{app:concepts_vs_performance}) performance vs. the number of contrastive concepts when considering \ccD and \ccLLM, (Sec. ~\ref{app:separability_disc}) CLIP's patch-level separability and how our method addresses this issue, (Sec.~\ref{app:sigmoid}) alternative to \cc{} scenario based on the sigmoid operation, and our experiments with filtering based on WordNet ontology (Sec.~\ref{sec:wordnetexp}).
% , \change{and analysis of the recall of the generated concepts (Sec.~\ref{sec:recall_stats})}.

\item In Sec.~\ref{app:llm_prompting} we provide details about LLM and the used prompts, together with examples of LLM-generated contrastive concepts.

\item Finally, in Sec.~\ref{sec:efficiency_analysis} we present an efficiency analysis of the proposed approach regarding the computation cost of generating the contrastive concepts and of employing them at segmentation time.

\end{itemize}

\section{Details on the evaluation}
\label{sec:protocol}

\subsection{Evaluation protocol}
\label{sec:eval_protocol}

Our experiments follow the evaluation protocol of \citet{cha2022tcl}. We use MMSegmentation implementation \cite{mmseg2020} with a sliding window strategy and resize input images to have a shorter side of 448.
In the case of CAT-Seg, we retain the original model framework and integrate \metric into Detectron~\cite{wu2019detectron2}. We also use its evaluation protocol, meaning that the input images differ from other evaluated methods,
i.e., with an input image size of 640x640.
Regarding the text prompts, we keep the native prompting of each method to stay as close as possible to the methods.

\subsection{Background handling of baselines}
\label{app:back-handling}

We detail here the different strategies employed in the methods that we evaluate to handle the background.

\begin{description}[labelindent=0.6cm]
\item[TCL] \cite{cha2022tcl} applies thresholding and considers pixels with maximal logit $\le 0.5$ to be in the background, where the logits are the cosine similarities of the visual embedding with the embedding of queries. 

\item[GEM] \cite{bousselham2023gem} applies a background handling strategy only for Pascal VOC. It only predicts the foreground classes. The background is obtained by thresholding the softmax-normalized similarity between the patch tokens and the text embedding of each class name. The threshold is fixed (set to~0.85). In our experiments with VOC, we explore the performance of GEM both with and without background handling and report each time a better score.
For other datasets than VOC, we apply only our methods.

\item[MaskCLIP] \cite{zhou2022maskclip} does not use any dedicated mechanism for background. Therefore, we do not report the original setup for it. 

\item[CLIP-DINOiser] \cite{wysoczanska2023clipdino} leverages a foreground/background saliency strategy which focuses on foreground pixels. In that case, the foreground/background is defined following FOUND \cite{simeoni2023found}, which focuses on objectness and mainly discards pixels corresponding to stuff-like classes, which might also be of interest.

\item[CAT-Seg]~\cite{cho2024catseg} does not apply any background handling strategy. Instead, for VOC they create a list of potential background classes and use them as "dummy" classes. This approach is closest to what we propose. In practice, for the VOC dataset, the authors use class names from the Context dataset, an extension of VOC with +40 class names.

\item[SAN]~\cite{xu2023side} does not design any background handling strategy and does not evaluate datasets with "background" class.

\end{description}

% \subsection{The overview of our approach}
% We present in ~\cref{fig:sup_overview} the overview of our method. As detailed in \cref{sec:cc}, we propose two solutions to generate \cc, one based on LLM (\ccLLM) and \ccD that relies on a distribution of co-occurring concepts in a pre-training dataset of a VLM. Our methods can be effectively integrated into various open-vocabulary segmentation methods.

% \begin{figure}
%     \centering
%     \includegraphics[width=0.8\linewidth]{figures/CC_visual_overview.pdf}
%     \caption{\textbf{Overview of our method.}}
%     \label{fig:sup_overview}
% \end{figure}

\subsection{About the \metric metric}
\label{sup:metric}

% \begin{figure}
%     \centering
%     \includegraphics[width=0.9\linewidth]{figures/metric_v4.pdf}
%     \caption{\textbf{Illustration of \metric metric}.}
%     \label{fig:metric}
% \end{figure}

% We present in \cref{fig:metric} an illustration of our proposed metric \metric. We show the difference between the standard mIoU metric (dataset-driven mIoU), where all the concepts present on an image are considered at once. On the contrary, our \metric considers each of the present concepts separately to measure the single-class segmentation ability of open-vocabulary semantic segmenters.

We present a pseudo-code of our metric in \cref{alg:metric}.

\begin{algorithm}[t]

\SetKwInOut{Input}{input}
\SetKwInOut{Output}{output}
\SetKwComment{Comment}{right mark}{left mark}
% \hspace{-2.5mm}
\Input{%
    ~$I$ -- input image: $I \in \Reals^{H \times W \times 3}$ \\
    ~$Y$ -- ground-truth annotations of $I$: $gt \in \Naturals ^{H \times W \times 1}$ \\
    ~$T$ --  ground-truth text labels \\
    ~$CC$ -- a dictionary of contrastive concepts per query \\
    ~$\texttt{model}$ -- segmenter producing pixel-level predictions given text queries}
\Output{~mean \metric, a mIoU score for a single-query scenario for a given image}
\textbf{procedure} $\texttt{IoUsingle}(I, Y)$:
\\
$\quad$ // \textit{Get unique classes from $Y$}
\\
$\quad$ $gt_{cls} \gets \texttt{unique}($Y$)$
\\
$\quad$ scores $\gets \emptyset$
\\
$\quad$ \textbf{for} $i \in gt_{cls}$ \textbf{do}
\\
$\qquad$ $q \gets T_{i}$
\\
$\qquad$ // \textit{Text prompts include query $q$ and contrastive concepts of $q$}
\\
$\qquad$ $t_{q} \gets q \cup CC_{q}$
\\
$\qquad$ // \textit{Get model predictions for given prompt set}
\\
$\qquad$ $\hat{y} \gets \texttt{model}(I, t_{q})$
\\
$\qquad$ // \textit{Get binarized version of predicted mask}
\\
$\qquad$ $\hat{y} \gets \texttt{binarize}(\hat{y}, i)$
\\
$\qquad$ // \textit{Get ground-truth binary mask for gt class $i$}
\\
$\qquad$ $y \gets \texttt{binarize}(Y, i)$
\\
$\qquad$ // \textit{Record corresponding IoU}
\\
$\qquad$ scores $\gets$ scores $\cup\; \texttt{IoU}(\hat{y}, y)$
\\
$\quad$ \textbf{end for}
\\
\textbf{return} $\texttt{mean}($scores$)$

\caption{\metric}
\label{alg:metric}
\end{algorithm}

\section{Additional results}
\label{sup:quantitative}

\subsection{More quantitative results}
\label{app:sota_miou}
\newcommand{\resultsrownobg}[8]{#1 & #2 & #3 & #4 & #5 & #6 & #7 & #8}

\vspace{-8pt}
\begin{table}
% \centering
\small
\renewcommand{\arraystretch}{1.}
\setlength{\tabcolsep}{1.4pt}
\resizebox{\linewidth}{!}{
\begin{tabular}
{l|cccc|ccc}
\toprule

& \textbf{Background} &
{\textbf{Type of}} &
{\textbf{CLIP}} &
{\textbf{Training}} &
\multicolumn{3}{c}{\textbf{Dataset}} 
\\
\resultsrownobg
{\textbf{Methods}}
{\textbf{handling}}
{\textbf{$\mathcal{CC}$}}
{\textbf{backbone}}
{\textbf{dataset}}
{Context}
{Object}
{VOC} \\

\midrule
\resultsrownobg
{GroupViT}
{threshold}
{$\varnothing$}
{scratch}
{CC12M+RedCaps}
{18.7} % Context
{27.5} % Object 
{50.4} % VOC
\\ %ADE 

\resultsrownobg
{CLIP-DIY}
{saliency}
{$\varnothing$}
{LAION-2B}
{-}
{19.7} % Context
{31.0} % Object 
{59.9} % VOC
\\ %ADE 

\resultsrownobg
{TCL}
{threshold}
{$\varnothing$}
{OpenAI} {CC12M+CC3M}
{24.3} % Context
{30.4} % Object 
{51.2} % VOC
\\

\resultsrownobg
{MaskCLIP\dag}
{$\varnothing$}
{$\varnothing$}
{OpenAI}
{-}
{21.1} % Context
{15.5} % Object 
{29.3} % VOC
\\ %ADE 

\resultsrownobg
{MaskCLIP$^{*}$}
{$\varnothing$}
{$\varnothing$}
{LAION-2B}
{-}
{22.9} % Context 
{16.4} % Object
{32.9} % VOC
\\ %ADE 

\resultsrownobg
{MaskCLIP$^{*}$ (\emph{+keys})}
{$\varnothing$}
{$\varnothing$}
{LAION-2B}
{-}
{24.0} % Context
{21.6} % Object
{41.3} % VOC
\\ %ADE

\resultsrownobg
{CLIP-DINOiser}
{$\varnothing$}
{$\varnothing$}
% {IN (random 1k im.)}
{LAION-2B}
{ImageNet (1k im.)}
{32.4} % Context
{29.9} % Object
{53.7} % VOC
\\ %ADE

\resultsrownobg
{GEM }
{$\varnothing$}
{$\varnothing$}
% {IN (random 1k im.)}
{MetaCLIP}
{-}
{-} 
{-} % Object
{46.8} % VOC
\\ %ADE
\midrule

\resultsrownobg
{\multirow{4}{*}{CLIP-DINOiser }}
{saliency}
{$\varnothing$}
{LAION-2B}
{ImageNet (1k im.)}
{--} % Context
{34.8} % Object
{\textbf{62.1}} % VOC
\\ %ADE 

\resultsrownobg
{}
{\cc}
{\ccB}
{LAION-2B}
{ImageNet (1k im.)}
{\bf 32.4} % Context
{29.5} % Object
{54.0} % VOC
\\ %ADE 

\resultsrownobg
{}
{\cc}
{\ccLLM}
{LAION-2B}
{ImageNet (1k im.)}
{31.3} % Context
{\bf 35.0} % Object
{60.8} % VOC
\\ %ADE 

\resultsrownobg
{}
{\cc}
{\ccD}
{LAION-2B}
{ImageNet (1k im.)}
{31.8} % Context
{33.3} % Object
{60.4} % VOC
\\ %ADE 

\midrule

\resultsrownobg
{\multirow{3}{*}{MaskCLIP}}
{\cc}
{\ccB}
{LAION-2B}
{-}
{23.6} % Context
{17.8} % Object
{35.1} % VOC
\\ %ADE 

\resultsrownobg
{ }
{\cc}
{\ccLLM}
{LAION-2B}
{-}
{\textbf{22.5}} % Context
{\bf 25.9} % Object
{46.2} % VOC
\\ %ADE 

\resultsrownobg
{ }
{\cc}
{\ccD}
{LAION-2B}
{-}
{23.2} % Context
{25.1} % Object
{\bf 46.4 } % VOC
\\ %ADE 

\midrule

\resultsrownobg
{GEM}
{threshold}
{$\varnothing$}
{MetaCLIP}
{-}
{\textbf{33.4}\rlap*} % Context
{27.4\rlap*} % Object
{46.6\rlap*} % VOC
\\ %ADE 

\resultsrownobg
{GEM}
{\cc}
{\ccLLM}
{MetaCLIP}
{-}
{31.6} % Context
{\bf 35.7 } % Object
{60.0} % VOC
\\ %ADE 

\resultsrownobg
{GEM}
{\cc}
{\ccD}
{MetaCLIP}
{-}
{32.1} % Context
{35.5} % Object
{\bf 60.5} % VOC
\\ %ADE 
\bottomrule
\end{tabular}
}
% \vspace{5pt}
\caption{\textbf{Results with standard mIoU metric} when employing different contrastive concept generation strategies. '*' denotes our implementation, `\dag' denotes results from~TCL \cite{cha2022tcl}, and 'MaskCLIP (+keys)' denotes keys refinement proposed in the original paper~\cite{zhou2022maskclip}. Training datasets include CC12M~\cite{changpinyo2021conceptual}, RedCaps~\cite{desai2021redcaps}, ImageNet~\cite{imagenet}, CC3M~\cite{cc2018}.}
\label{table:miou_all}
\end{table}

\paragraph{State-of-the-art results under classic mIoU.}
In~\cref{table:miou_all}, we report the results under the classic mIoU metric for selected state-of-the-art methods on open-vocabulary semantic segmentation. For each of the methods, we detail the specific background handling techniques (if any), the CLIP backbone used as well as additional datasets used for training. 

Extending the dataset vocabulary with our generated contrastive concepts does not hurt the overall performance under a normal setup when all dataset labels are considered prompts. For GEM and MaskCLIP we observe significant improvements over their original setups on VOC. This holds for both contrastive concept generation methods \ccD\ and \ccLLM. Looking at the results of CLIP-DINOiser, we observe that saliency is still more effective in the object-centric scenario.

\paragraph{More open-world evaluation results.}
\label{app:quantitative_res}
\begin{table}
    \centering
    \def\cite#1{}
    %\resizebox{0.9\linewidth}{!}{
    \begin{tabular}{lc|d|accc}
      Method   & CLIP dataset & Original & \ccP & \ccB & \,\ccLLM\,  & \,\ccD\,  \\
     \toprule
    \rowcolor{blue!20!white}\multicolumn{7}{l}{VOC} \\
    \multirow{3}{*}{MaskCLIP~\cite{zhou2022maskclip}} & LAION-2B  & -- & 49.9 & 47.9 & 51.8 & \bf  53.6 \\
    & OpenAI  & -- & 47.1 & 44.2 & 52.2 & \bf 53.4 \\
   & MetaCLIP  & -- &47.9 & 46.6 & \bf 50.6 & 50.1 \\
    \midrule
    CLIP-DINOiser \cite{wysoczanska2023clipdino} & LAION-2B & 63.8\rlap* & 61.0 & 59.3 & 63.1 & \bf 64.7    \\
    TCL~\cite{cha2022tcl} %+ BG handling 
    & TCL's \cite{cha2022tcl} & 52.9\rlap* & 53.0\rlap* &  52.9\rlap* & 52.6\rlap* & \bf 53.6\rlap* \\
    GEM & MetaCLIP&  -- & -- & 48.6\rlap* & 61.3\rlap* & \bf 64.6\rlap* \\
    CAT-Seg & OpenAI &  -- & -- & 52.8 & \textbf{69.5} & 67.7 \\
    \midrule
    \midrule
    \rowcolor{blue!20!white}\multicolumn{7}{l}
    {Cityscapes} \\
    \multirow{3}{*}{MaskCLIP~\cite{zhou2022maskclip}} & LAION-2B  & -- & 32.2 & 16.2 & \bf 27.2 & 24.0 \\
    % & OpenAI  & -- & 30.6 & 15.0 & \bf 23.1 & 22.1 \\
    & OpenAI  & -- & 30.6 & 15.0 & \bf 22.5 & 22.0 \\
   & MetaCLIP  & -- & 30.0 & 13.6 & \bf 24.6 & 23.3 \\
    \midrule
    CLIP-DINOiser \cite{wysoczanska2023clipdino} & 
    LAION-2B & 20.8\rlap & 36.0 & 23.2 & \bf 30.6 & 27.3 \\
    % TCL~\cite{cha2022tcl} & TCL's \cite{cha2022tcl} & 18.6\rlap* & 29.7 & 9.8 & \bf 26.5 & 22.0 \\
    TCL~\cite{cha2022tcl} & TCL's \cite{cha2022tcl} & 18.6\rlap* & 29.7 & 9.8 & \bf 26.3 & 22.0 \\
    GEM &  MetaCLIP & -- & 20.6 & 14.5\rlap* & \textbf{21.5} & 14.6 \\
    \midrule
    \midrule
    \rowcolor{blue!20!white}\multicolumn{7}{l}{COCO-Stuff } \\
    \multirow{3}{*}{MaskCLIP~\cite{zhou2022maskclip}} & LAION-2B  & -- & 34.1 & 26.4 & 28.8 & \bf 29.5 \\
    & OpenAI  & -- &33.6& 24.1 & 28.4 & \bf 28.8 \\
   & MetaCLIP   & -- &34.0& 25.8 & \bf 28.1 & \bf 28.1 \\
    
    CLIP-DINOiser \cite{wysoczanska2023clipdino} & LAION-2B & 28.0\rlap* & 35.3 & 32.4 &  33.9 & \bf 34.4 \\
    TCL~\cite{cha2022tcl} & TCL's \cite{cha2022tcl} & 25.0\rlap* & 34.7 & 17.4 & 29.5 & \bf 30.6 \\
    GEM &  MetaCLIP & -- & 38.3 & 22.9\rlap* & 32.2 & \bf 33.6\\
    \midrule
    \midrule

% ADE %%%%%%%%%%%%%%%%%%%
    \rowcolor{blue!20!white}\multicolumn{7}{l}{ADE20k} \\
    \multirow{3}{*}{MaskCLIP~\cite{zhou2022maskclip}} & LAION-2B  & -- & 33.2 & 22.7 & 26.8 & \bf 27.8 \\
    & OpenAI  & -- & 29.8 & 20.2 & 23.5 & \bf 25.2 \\
   & MetaCLIP   & -- & 32.1 & 21.5 & 24.7 & \bf 26.0 \\
    \midrule
    CLIP-DINOiser \cite{wysoczanska2023clipdino} & LAION-2B & 28.8\rlap* & 35.3 &  28.9 & 29.7 & \bf 31.6 \\
    TCL~\cite{cha2022tcl} & TCL's \cite{cha2022tcl} & 14.8\rlap* & 32.6 % No BG
    & 14.9\rlap* & 25.9 & \bf 26.5\\
    GEM & MetaCLIP &  -- & 33.0 & 21.5\rlap*& 26.3 & \bf 29.1\\
    CAT-Seg & OpenAI &  -- & 46.8 & 25.7 & 38.4 & \textbf{39.7} \\
    \midrule
    \midrule
% Coco Object %%%%%%%%%%%%%%%%%%%
    \rowcolor{blue!20!white}\multicolumn{7}{l}{COCO-Object 
    } 
    \\
    \multirow{3}{*}{MaskCLIP~\cite{zhou2022maskclip}} & LAION-2B  & -- &32.1 & 27.7 & \bf 33.7 & 32.9\\
    & OpenAI  & -- & 31.3&  24.3 & \bf 34.5 & 33.3 \\
   & MetaCLIP   & -- & 30.9 & 27.4 & \bf 32.2 & 31.1 \\
    \midrule
    CLIP-DINOiser \cite{wysoczanska2023clipdino} & LAION-2B & 38.8\rlap* & 38.9 & 35.5 & \bf 41.6 & 39.9 \\
    TCL~\cite{cha2022tcl} & TCL's \cite{cha2022tcl} & 37.1\rlap* &38.1 & 37.2\rlap* & \bf 38.1\rlap* & 37.2\rlap* \\
    GEM & MetaCLIP  & -- & -- & 31.4 & 39.7 & \bf 40.1\\
    \midrule
    \midrule
    % Context %%%%%%%%%%%%%%%%%%%
    \rowcolor{blue!20!white}\multicolumn{7}{l}{Pascal Context  
    } 
    \\
    \multirow{3}{*}{MaskCLIP~\cite{zhou2022maskclip}} & LAION-2B  & -- &40.5&34.4 &  35.2 & \bf 37.4 \\
    & OpenAI  & -- & 41.1 & 32.9 & 34.7 & \bf 36.8  \\
   & MetaCLIP   & -- & 41.1 & 32.6 & 34.2 & \bf 35.8 \\
    \midrule
    CLIP-DINOiser \cite{wysoczanska2023clipdino} & LAION-2B & 33.9\rlap* & 45.8 & 41.5 & 41.6 & \bf 44.2 \\
    TCL~\cite{cha2022tcl} & TCL's \cite{cha2022tcl} & 29.7\rlap* & 41.7 & 29.7\rlap* & 36.8 & \bf 38.2 \\
    GEM & MetaCLIP  & -- & -- & 26.9 & 40.1 & \bf 42.1 \\
    \bottomrule
    \end{tabular}
    \caption{Results on all datasets with our \metric metric defined in \cref{sec:metric}. `*' denotes the result when the original background handling gives the best results.}
    \label{tab:1vsrest-all}
\end{table}

\cref{tab:1vsrest-all} extends \cref{tab:1vsrest-VOC-paper} and completes the results obtained with the \metric on all the datasets that we considered.

\subsection{Failure case analysis}
\label{app:limitations}

\begin{figure}[h!]
    \centering
    \resizebox{\linewidth}{!}{
    \renewcommand{\arraystretch}{0.1}
    % \fbox{
    \begin{tabular}{c@{}c@{}c@{}c@{}c@{}c}
        & GT &  \ccD (single) & \ccD (all) & \ccLLM (single) & \ccLLM (all) \\

         % other example
         \raisebox{1.5\normalbaselineskip}[0pt][0pt]{\rotatebox{90}{\makecell{\promptstyle{bed_sumpat}{bed}}}} & \includegraphics[width=0.165\linewidth]{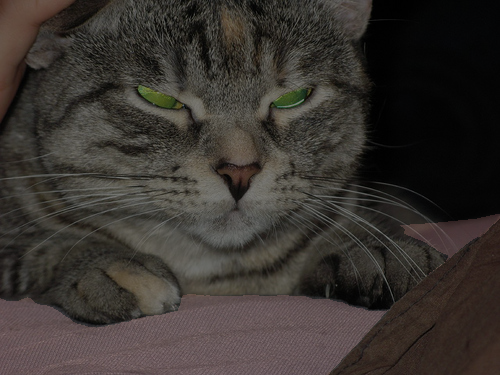} & 
         \includegraphics[width=0.165\linewidth]{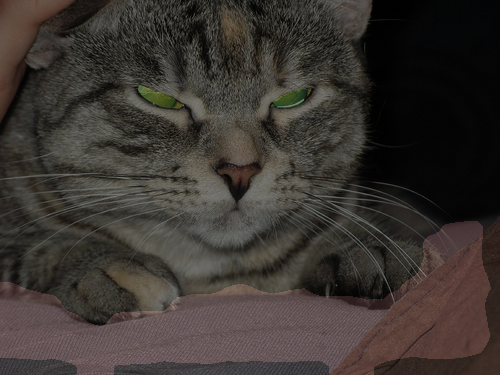} & \includegraphics[width=0.165\linewidth]{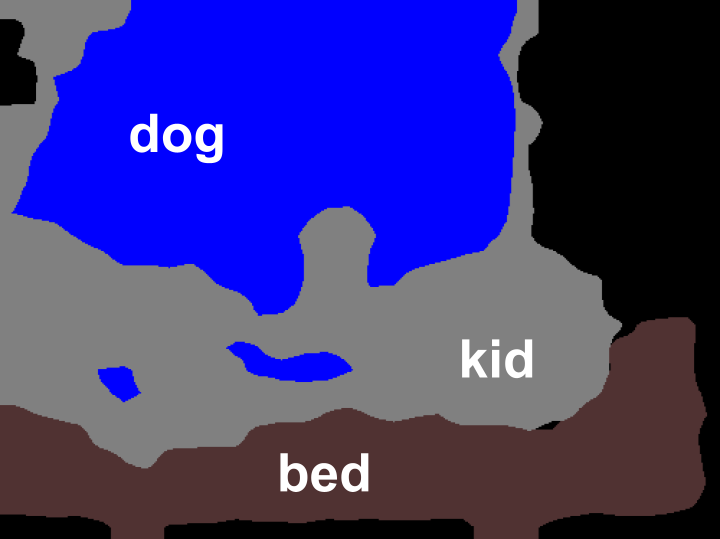} &          \includegraphics[width=0.165\linewidth]{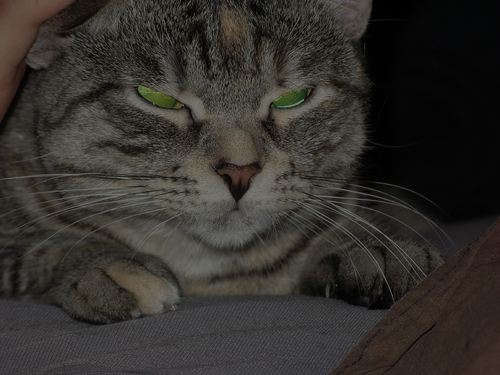} & 
         \includegraphics[width=0.165\linewidth]{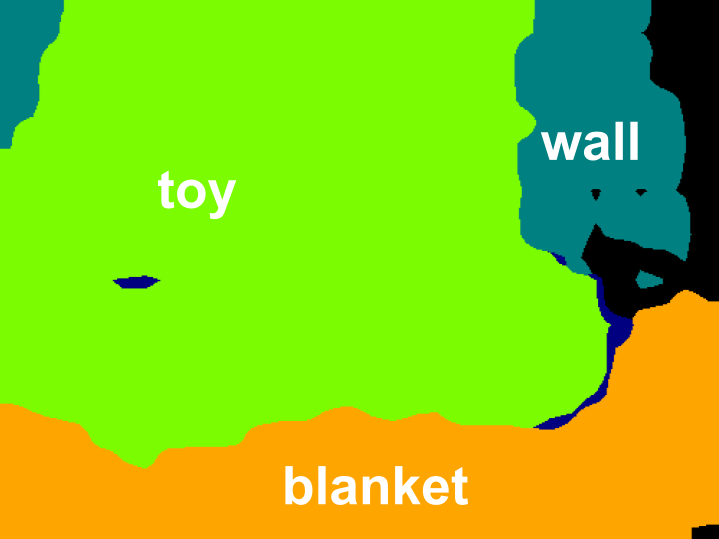} \\
                  % other example
         \raisebox{1.4\normalbaselineskip}[0pt][0pt]{\rotatebox{90}{\makecell{\promptstyle{rug}{rug}}}} & \includegraphics[width=0.165\linewidth]{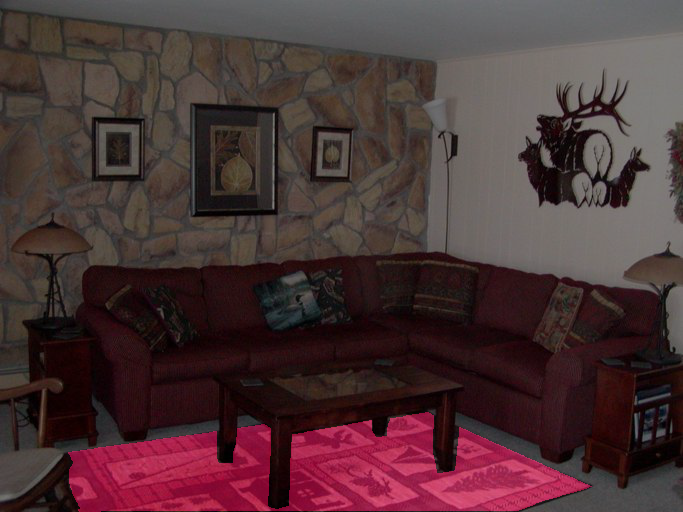} 
         
         % & 
         % \includegraphics[width=0.165\linewidth]{figures/qualitative/supmat/context/2008_006185/2008_006185_bg_boat_blend (1).png} 
         &
         \includegraphics[width=0.165\linewidth]{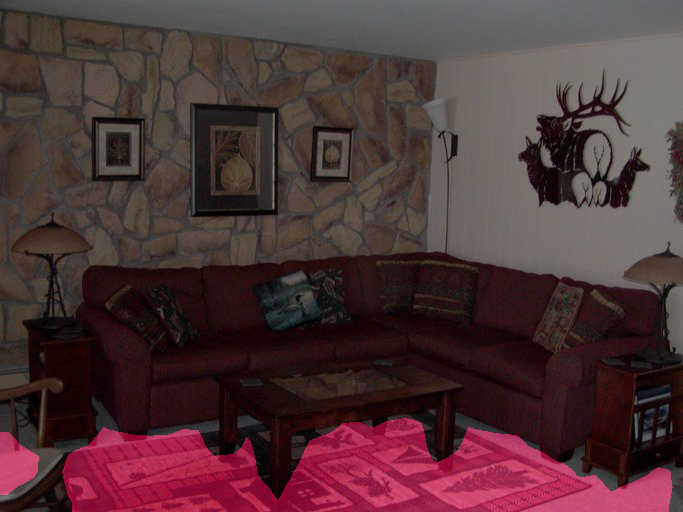} & 
         \includegraphics[width=0.165\linewidth]{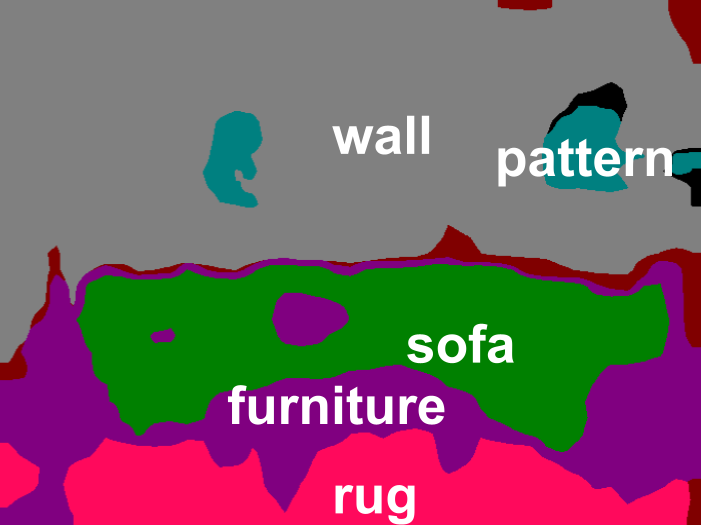} & 
         \includegraphics[width=0.165\linewidth]{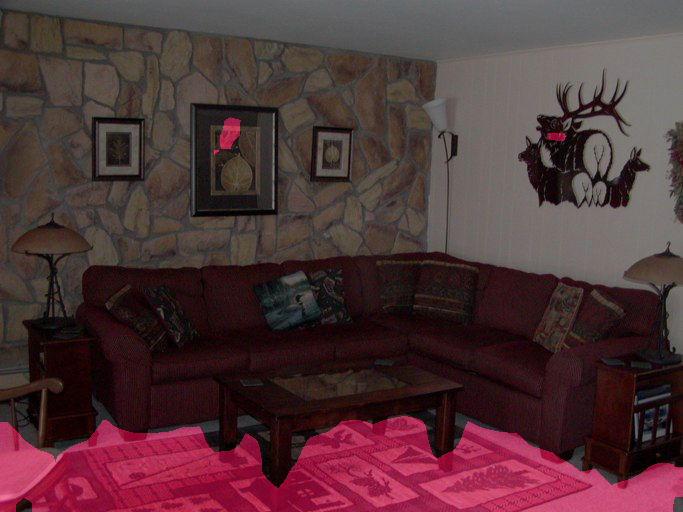} & 
         \includegraphics[width=0.165\linewidth]{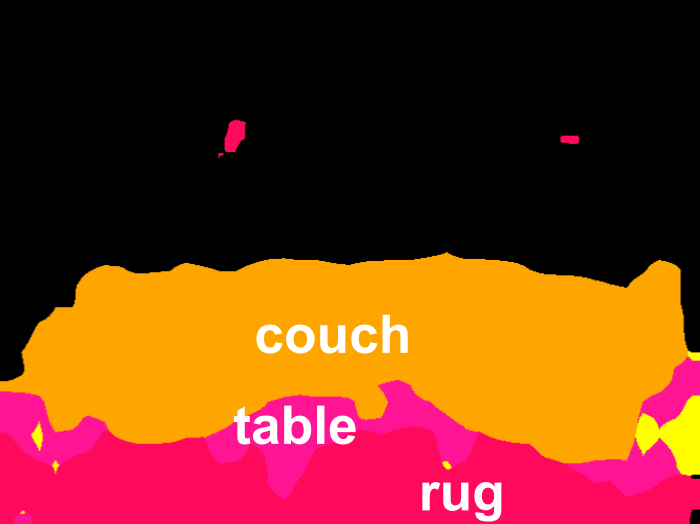} \\
                          % other example
         \raisebox{0.01\normalbaselineskip}[0pt][0pt]{\rotatebox{90}{\makecell{\promptstyle{skyscraper}{bedclothes}}}} & \includegraphics[width=0.165\linewidth]{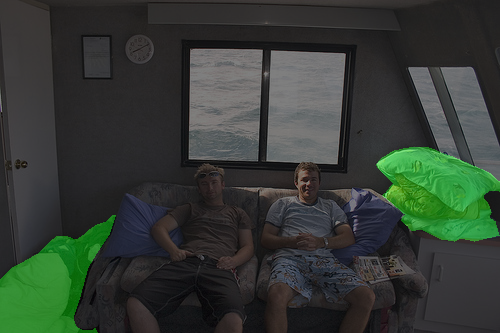} & 
         \includegraphics[width=0.165\linewidth]{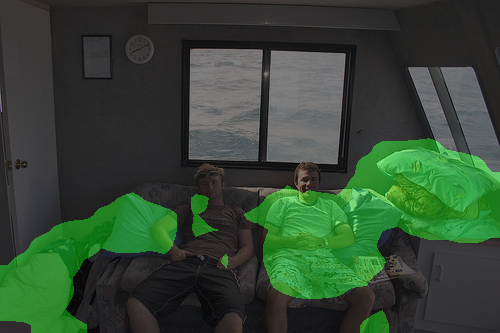} & 
         \includegraphics[width=0.165\linewidth]{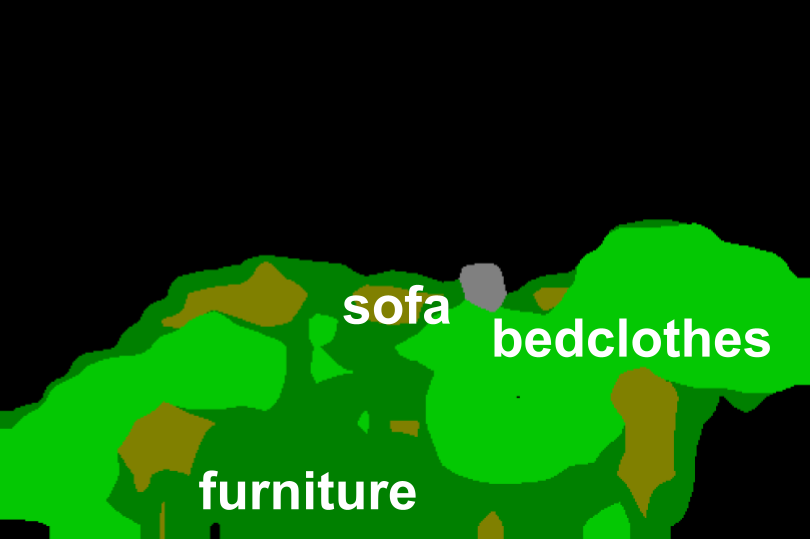} & 
         \includegraphics[width=0.165\linewidth]{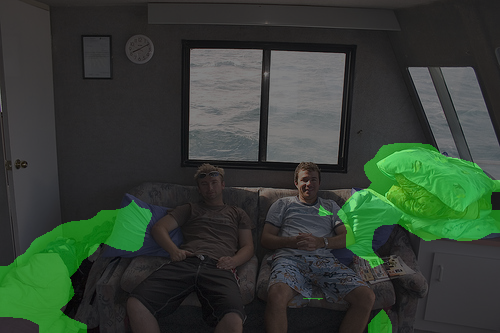} & 
         \includegraphics[width=0.165\linewidth]{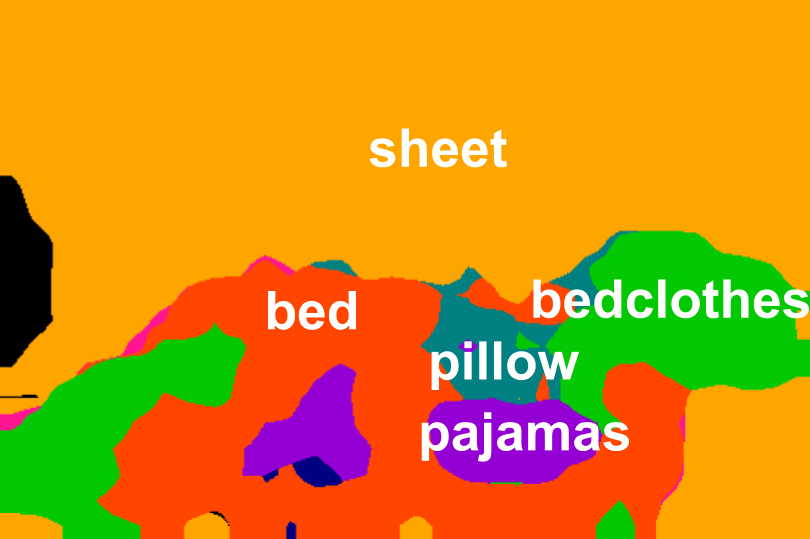} \\

    \end{tabular}}
    % }
    %\vspace{-5pt}
    \caption{\textbf{Failure cases of our method.} We show examples of CLIP-DINOiser when one of the methods fails to generate accurate \cc. In the first example \ccLLM suggests ``blanket" for ``bed" which typically covers the query concept. In the second row, both methods fail to provide ``floor" to contrast with ``rug". Finally, in the third example, both methods fail to generate ``person" to contrast with ``bedclothes", however, \ccLLM suggest ``pyjamas", which results in a better segmentation. }
    \label{fig:failures}
\end{figure}

We present some failure cases of our approach in \cref{fig:failures}. Precisely, we show examples of CLIP-DINOiser when one of the generation methods fails.
The first example (first row), \ccLLM suggests “blanket" for “bed", which typically covers the query concept. One of the potential improvements would be to instruct an LLM to ignore potentially occluding objects.
In the second row, both methods fail to provide “floor" to contrast with “rug". We notice that \ccLLM tend to be more oriented towards objects, as opposed to stuff-like classes. We also observe that in the example, a small part of a painting on the wall is segmented as "rug". This suggests that \cc{}s might not give a complete set of.
Finally, in the third example, both methods fail to generate “person" to contrast with “bedclothes". However, \ccLLM includes “pyjamas", which results in a better segmentation overall. Image-conditioned generation (e.g., with VLMs) could be a candidate solution to this problem, but we leave it for future work.

\vspace{5pt}

\subsection{More qualitative results}
\label{app:qualitative_res}

\begin{figure}[h!]
    \centering
    \resizebox{\linewidth}{!}{
    \renewcommand{\arraystretch}{0.1}
    % \fbox{
    \begin{tabular}{c@{}c@{}c@{}c@{}c@{}c@{}c}
        & GT & \ccB & \ccD (single) & \ccD (all) & \ccLLM (single) & \ccLLM (all) \\

         % other example
         \raisebox{1.5\normalbaselineskip}[0pt][0pt]{\rotatebox{90}{\makecell{\promptstyle{bed_sumpat}{bed}}}} & \includegraphics[width=0.165\linewidth]{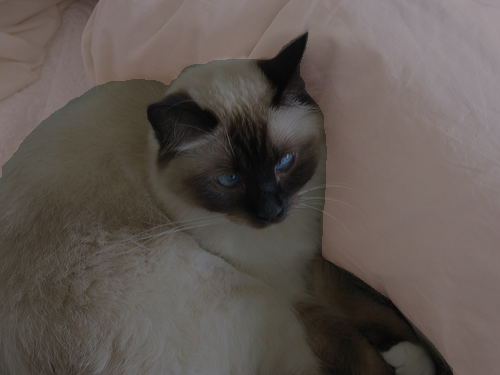} & 
         \includegraphics[width=0.165\linewidth]{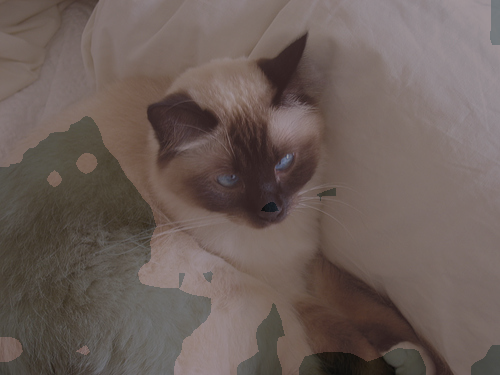} & \includegraphics[width=0.165\linewidth]{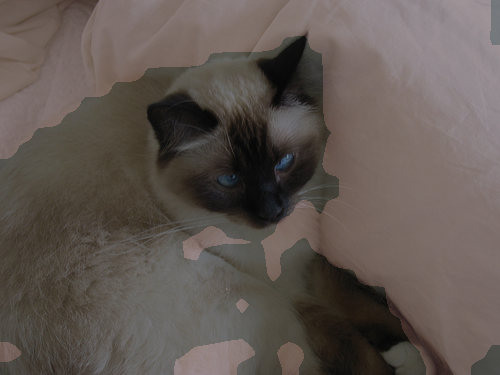} & \includegraphics[width=0.165\linewidth]{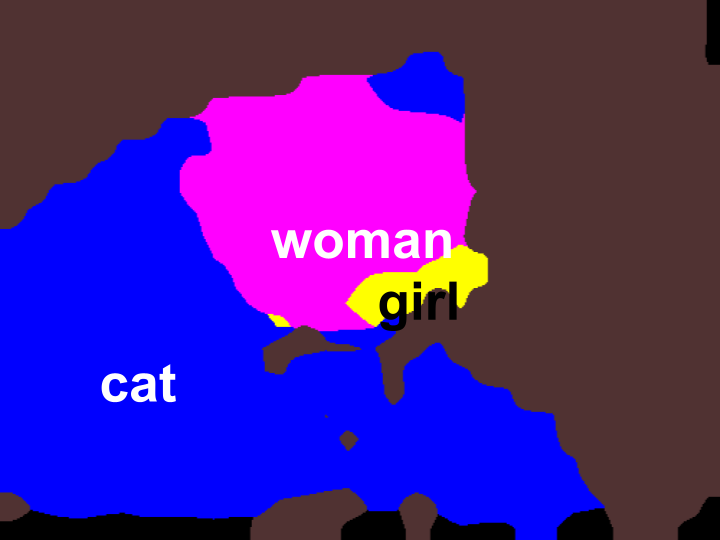} &          \includegraphics[width=0.165\linewidth]{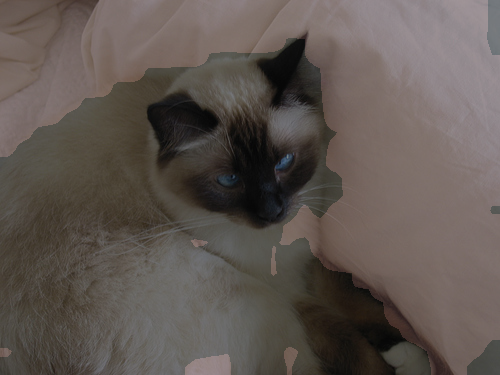} & 
         \includegraphics[width=0.165\linewidth]{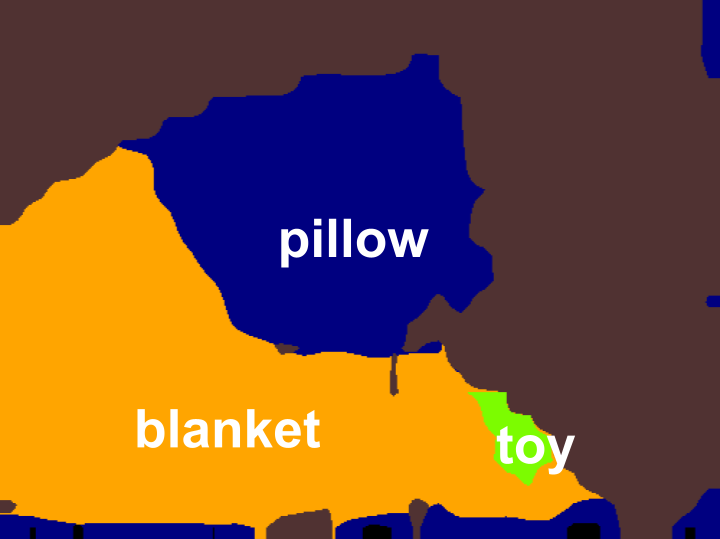} \\
                  % other example
         \raisebox{1.4\normalbaselineskip}[0pt][0pt]{\rotatebox{90}{\makecell{\promptstyle{boat}{boat}}}} & \includegraphics[width=0.165\linewidth]{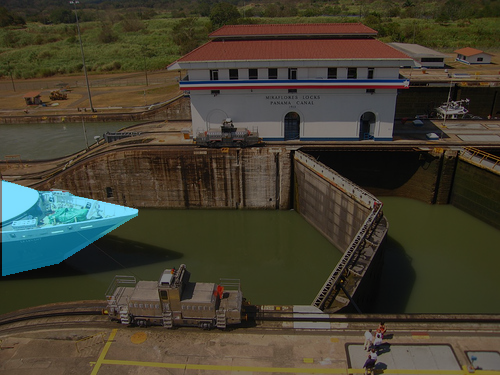} & 
         \includegraphics[width=0.165\linewidth]{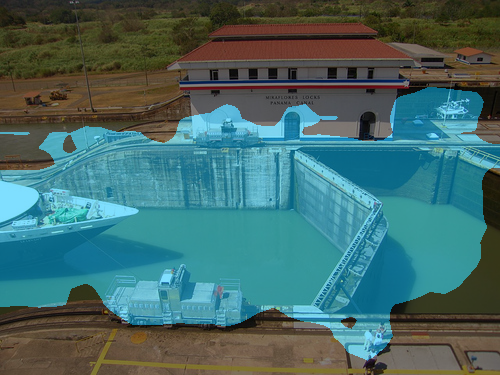} &
         \includegraphics[width=0.165\linewidth]{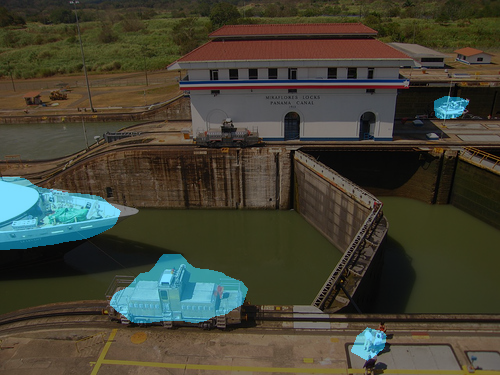} & 
         \includegraphics[width=0.165\linewidth]{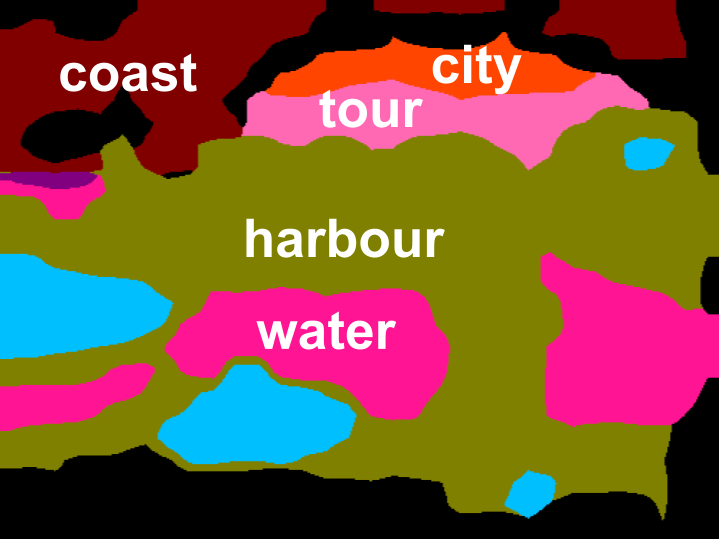} & 
         \includegraphics[width=0.165\linewidth]{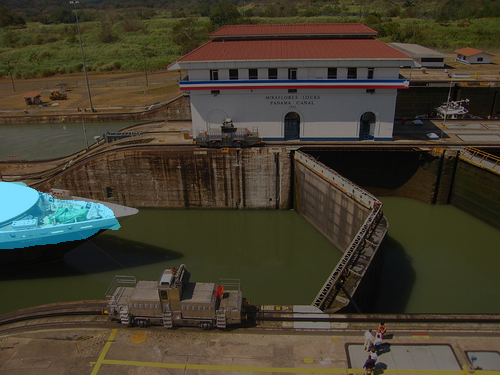} & 
         \includegraphics[width=0.165\linewidth]{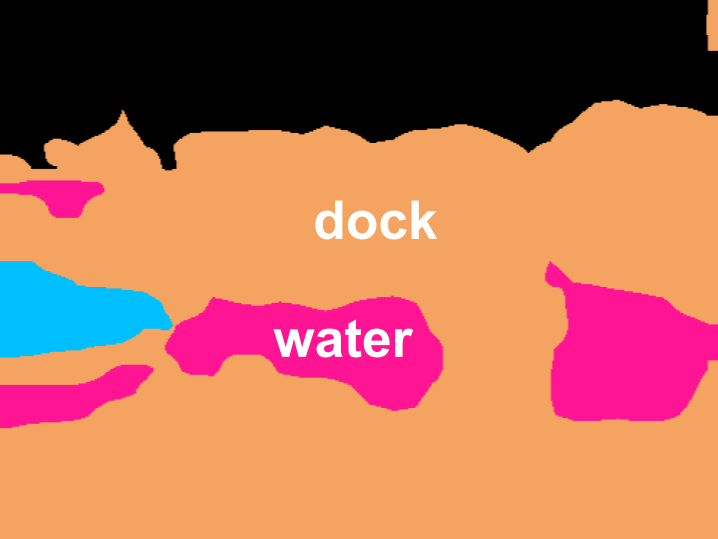} \\
                          % other example
         \raisebox{0.01\normalbaselineskip}[0pt][0pt]{\rotatebox{90}{\makecell{\promptstyle{skyscraper}{skyscraper}}}} & \includegraphics[width=0.165\linewidth]{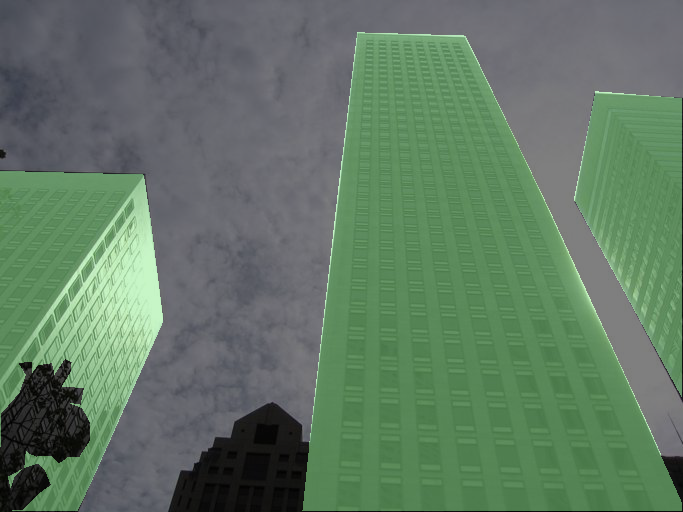} & 
         \includegraphics[width=0.165\linewidth]{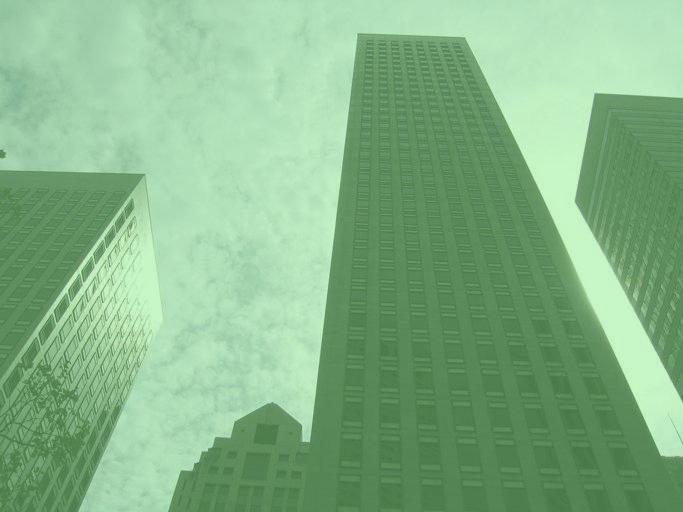} &
         \includegraphics[width=0.165\linewidth]{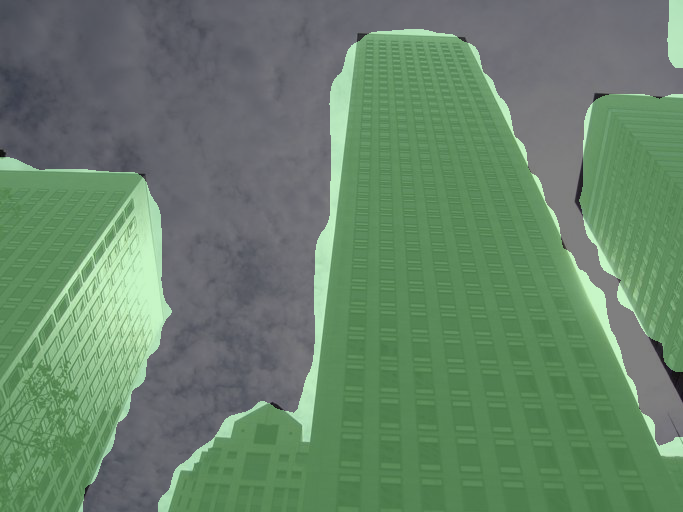} & 
         \includegraphics[width=0.165\linewidth]{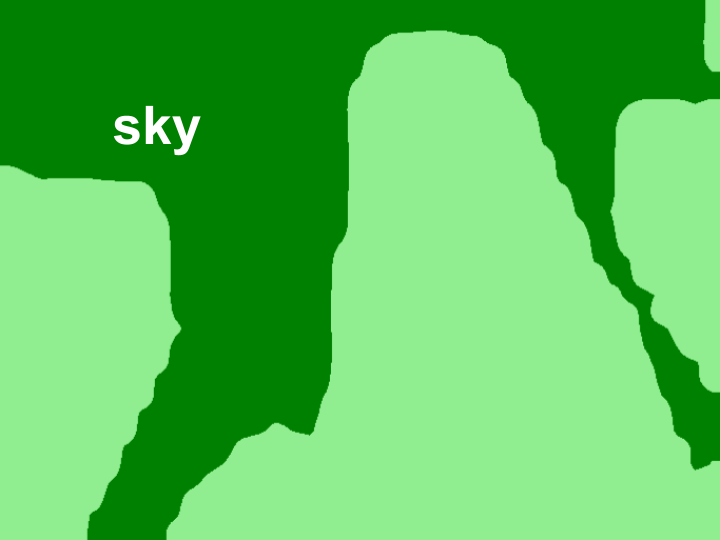} & 
         \includegraphics[width=0.165\linewidth]{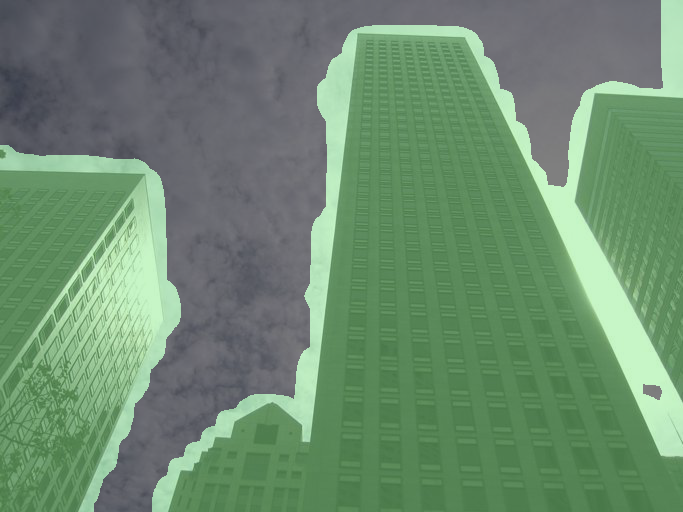} & 
         \includegraphics[width=0.165\linewidth]{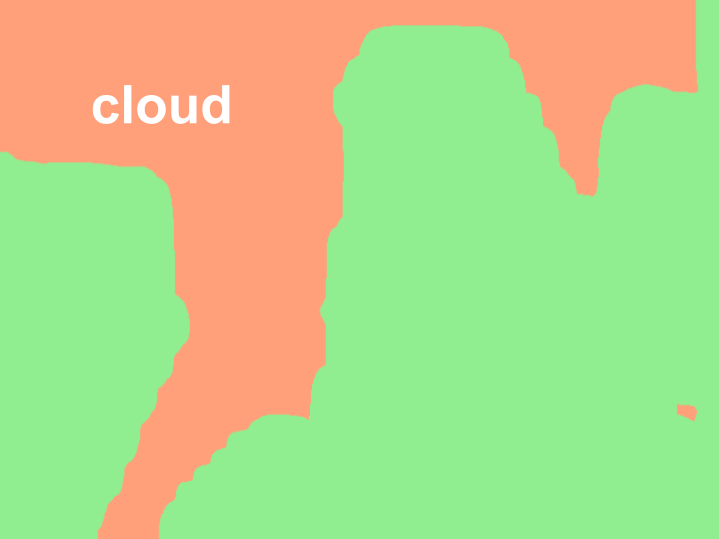} \\

    \end{tabular}}
    % }
    %\vspace{-5pt}
    \caption{\textbf{More qualitative results} of CLIP-DINOiser with different \cc. Here we 
    focus on cases where \ccD{} and $\CCLLM$ give different results. 
    For ``boat" (2\textsuperscript{nd} row), $\CCLLM$ gives a better result providing a good \cc\xspace (``dock"). On the other hand, for ``skyscraper" (3\textsuperscript{rd} row), \ccD\xspace yields slightly better results suggesting ``sky" and not ``cloud". Note that in this last example, \ccB\xspace completely fails, possibly due to a difficult (uncommon) angle of view.}
    \label{fig:qualitative2}
\end{figure}

More qualitative results are provided in \cref{fig:qualitative2}, comparing \ccD\ to \ccLLM.
\vspace{5pt}

\section{Additional analyses}
\label{app:additional_analysis}

\subsection{Hyperparameter selection}
\label{sec:sup_hyperparams}

% \newcolumntype{a}{>{\columncolor{white!80!black}}c}
% \vspace{8pt}
\begin{table}[h]
\centering
\setlength{\tabcolsep}{2.5pt}
    \begin{tabular}{lc|ccacc|cccac}
    \toprule
     & & \multicolumn{5}{c}{values of $\gamma$} & \multicolumn{5}{|c}{values of $\delta$} \\
      Method & CLIP tr. data 
      % & Original 
      & 0.001 &0.005 & 0.01 &  0.015& 0.02 % gamma
      & 0.95 & 0.9 & 0.85 & 0.8 & 0.75  \\ % delta \\
     \toprule
     \multirow{3}{*}{MaskCLIP
     % ~\cite{zhou2022maskclip}
     }

     & OpenAI
    &  24.4	&26.0 &	24.8	&24.4&	23.2 
    & 19.9	&21.0&	23.0&	24.4	& 22.8\\
    
    & Laion2B 
     & 25.8&	27.8	&27.4&	26.0&	25.4  
     & 23.0	&24.1	&26.4	&27.4	&24.6 \\
     
    & MetaCLIP 
     & 22.0 &	24.1	&24.4	&23.8&	23.4 
     & 22.7	&23.7	&25.9	&27.2	&23.7 \\
     \midrule
     DINOiser
     % ~\cite{wysoczanska2023clipdino} 
     & Laion2B 
     &  24.4	&27.2	&27.9&	27.9	&27.7 
     &  23.5&	24.6	&26.4&	27.9	&26.9 \\
    \bottomrule
    \end{tabular}
    % \vspace{5pt}
    \caption{\textbf{Parameter study of $\gamma$ and $\delta$}. Selection (marked in grey) of the hyperparameters $\gamma$ and $\delta$ with \metric on 100 randomly-selected images in ADE20k training dataset.}
    \label{tab:sup_ade_hyper}
\end{table}

This section discusses the selection of hyperparameters for our \cc\ generation. For the frequency threshold $\gamma$ and the cosine similarity threshold $\delta$, we randomly select 100 images from the training set of the ADE20K dataset and report \metric on this subset --- which we observed was enough to select the values. We report in \cref{tab:sup_ade_hyper} a parameter study of both hyperparameters and mark in grey selected values, i.e., $\gamma=0.01$ and $\delta=0.8$.
For~$\gamma$, we observe that values $\gamma < 0.005$ are too low, most likely introducing too much noise in selected contrastive concepts.

% \newcolumntype{a}{>{\columncolor{white!80!black}}c}
%\vspace{15pt}
\begin{table}[t!]
    \setlength{\tabcolsep}{2.5pt}
    \centering
    \begin{tabular}{lc|ccacc}
    \toprule
      Method & CLIP training data 
      & 1.0 & 0.95 & 0.9 & 0.85 & 0.8 \\
     \toprule
     \multirow{3}{*}{MaskCLIP}
     & OpenAI 
    &  26.0 & 40.4&	41.1	&39.1	&32.1 \\
    & Laion2B 
     & 35.3 & 43.7&	44.0	&44.6	&42.2 \\
    & MetaCLIP 
     & 24.4	&39.1	&40.3&	34.3	&30.6  \\
     \midrule
    DINOiser & Laion2B 
     & 51.3&	57.8	&58.6&	58.8	&55.2 \\
    TCL & TCL's %~\cite{cha2022tcl}
    & 37.2&	47.6	&47.7	&47.1&	47.7
    \\
    \bottomrule
    \end{tabular}
    %\vspace{4pt}
    \caption{\textbf{Selection of $\beta$ with classic mIoU} on 100 randomly-selected images in the VOC training dataset. Results are reported for \ccLLM. 
    }
    
    \label{tab:sup_beta} 
\end{table}

\cref{tab:sup_beta} presents a parameter study of the cosine similarity of text queries $\beta$ in multi-query segmentation. Here, we randomly select 100 images from the VOC training set and report classic mIoU for different $\beta$ values. We select $\beta=0.9$ because it gives the best result for most methods. We also note that controlling the similarity between query concepts and contrastive concepts in the multiple-query scenario is necessary. Not including this step (see results for $\beta=1.0$) greatly degrades performance.

\subsection{Average number of contrastive concepts vs performance}
\label{app:concepts_vs_performance}

We present in \cref{fig:noccvsperf} a scatterplot of performance vs the number of contrastive concepts when considering \ccD\ (\cref{fig:noccvsperf}(a)) and \ccLLM\ (\cref{fig:noccvsperf}(b)). The points correspond to the \metric\ scores per class obtained with CLIP-DINOiser on all the datasets we evaluate. We do not observe a strong correlation between the number of contrastive concepts and performance, although there is a small mode of around 20 concepts when using \ccD. We also observe that, on average, $|\CCD| > |\CCLLM|$.

\begin{figure*}[!h]
    \vspace*{10pt}
    \centering
    \begin{subfigure}{0.45\textwidth}
    \centering
    \input{figures/evaluation/fig_corr_ccl}
    \caption{\ccLLM}        
    \end{subfigure}
    % \hfill
    \begin{subfigure}{0.45\textwidth}
    \centering
    \input{figures/evaluation/fig_corr_ccd}
    \vspace{-5pt}
    \caption{\ccD}        
    \end{subfigure}
    \vspace{-5pt}
    \caption{\textbf{Number of \cc\ vs performance.} We compare the number of \cc\ against the performance of CLIP-DINOiser for each class used in our evaluations (considering all datasets). Performance is reported with per class \metric\%. }
    \label{fig:noccvsperf}
\end{figure*}

\subsection{On separability of CLIP patch-features}
\label{app:separability_disc}

\begin{figure*}[!h]
    % \vspace*{10pt}
    \centering
    \begin{subfigure}{0.45\textwidth}
    \centering
    \begin{tikzpicture}
\begin{axis}[
    width=6.2cm,
    height=5cm,
    title={},
    title style={font=\large\sffamily},
    xlabel={Cosine similarity},
    xlabel style={font=\sffamily},
    ylabel={$\times 10^5$},
    ylabel style={font=\sffamily},
    ymin=0,
    ymax=6e5,
    xmin=-1,
    xmax=1,
    xtick={-1.0,-0.5,0,0.5,1.0},
    ytick={0,2e5,4e5,6e5,8e5},
    yticklabels={0,2,4,6,8},
    ymajorgrids=false,
    xmajorgrids=false,
    axis lines=box,
    bar width=0.025,
    enlarge x limits=false,
    scaled y ticks=false
]
% Histogram data points (approximated from the image)
\addplot[
    ybar interval,
    fill=blue!70,
    draw=black,
    line width=0.1pt,
]
coordinates {
(0.04686367, 685)
(0.078053236, 17770)
(0.109242804, 140258)
(0.14043237, 406461)
(0.17162193, 502529)
(0.20281151, 318353)
(0.23400107, 110409)
(0.26519063, 39941)
(0.29638022, 11170)
(0.32756978, 1111)
};

\end{axis}
\end{tikzpicture}
    \caption{VOC}        
    \end{subfigure}
    % \hfill
    \begin{subfigure}{0.45\textwidth}
    \centering
    \begin{tikzpicture}
\begin{axis}[
    width=6.2cm,
    height=5cm,
    title={},
    title style={font=\large\sffamily},
    xlabel={Cosine similarity},
    xlabel style={font=\sffamily},
    ylabel={$\times 10^6$},
    ylabel style={font=\sffamily},
    ymin=0,
    ymax=9e6,
    xmin=-1,
    xmax=1,
    xtick={-1.0,-0.5,0,0.5,1.0},
    ytick={0,2e6,4e6,6e6,8e6},
    yticklabels={0,2,4,6,8},
    ymajorgrids=false,
    xmajorgrids=false,
    axis lines=box,
    bar width=0.025,
    enlarge x limits=false,
    scaled y ticks=false
]

\addplot[
    ybar interval,
    fill=blue!70,
    draw=black,
    line width=0.1pt,
] coordinates {
    (0.05,0.3e6)
    (0.10,1.6e6)
    (0.15,6.5e6)
    (0.20,8.7e6)
    (0.25,3e6)
    (0.30,0.4e6)
    (0.35,0.1e6)
};

\end{axis}
\end{tikzpicture}
    % \vspace{-5pt}
    \caption{ADE20K}        
    \end{subfigure}
    \vspace{-5pt}
    \caption{\textbf{Distribution of maximum patch similarities with text prompts.} We plot histograms for 100 images of VOC (a) and ADE20K (b) of patch similarities in MaskCLIP.
    }
    \label{fig:sep_disc}
\end{figure*}
In~\cref{fig:sep_disc}, we present an analysis of the patch-level CLIP space using MaskCLIP features. The figure shows histograms of patch-level maximum text similarities (in cosine similarity) across 100 randomly sampled images from VOC (a) and ADE20k (b). We notice an overall concentration of cosine similarity scores in [0.1,0.3], suggesting that the feature space is not easily separable.

% \vspace{5pt}
\begin{figure}
    \centering
    \renewcommand{\arraystretch}{0.1}
    \centering
    \subfloat[][\textbf{Input} image]{\includegraphics[width=0.55\linewidth]{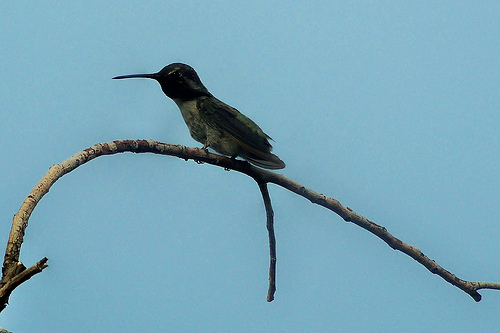}} 
    \\ 
     \subfloat[][]{
    \begin{tabular}{c@{}c@{}}
        % \cc = \varnothing & \ccB & \ccD & \ccLLM \\
        \cc = $\varnothing$ & \ccB \\

         % other example
         % \raisebox{2.5\normalbaselineskip}[0pt][0pt]{\rotatebox{90}{\makecell{\promptstyle{bird}{arm}}}} & 
         \includegraphics[trim={1.2cm, 1.2cm, 1.2cm, 1.2cm},clip, width=0.5\linewidth]{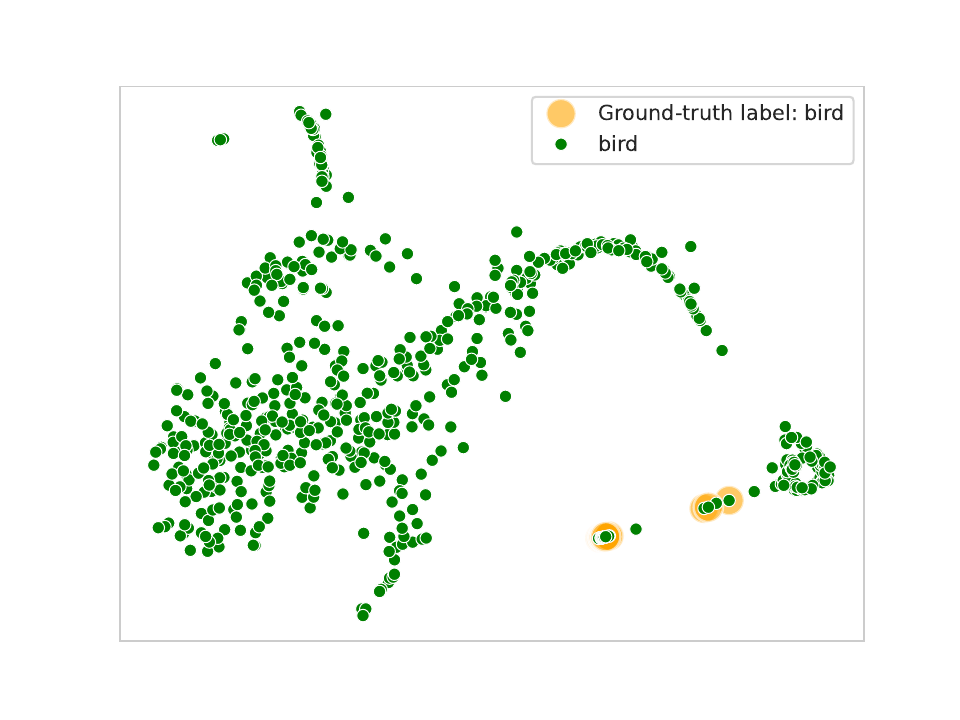} & 
         \includegraphics[trim={1.2cm, 1.2cm, 1.2cm, 1.2cm},clip, width=0.5\linewidth]{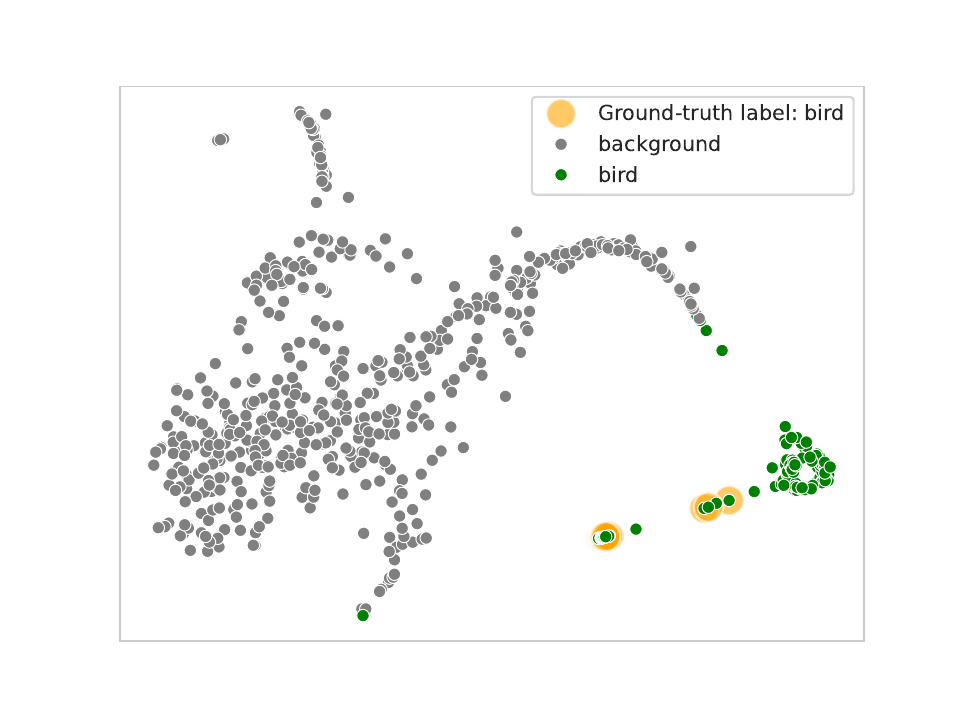} \\    
         \ccD & \ccLLM \\
         \includegraphics[trim={1.2cm, 1.2cm, 1.2cm, 1.2cm},clip, width=0.5\linewidth]{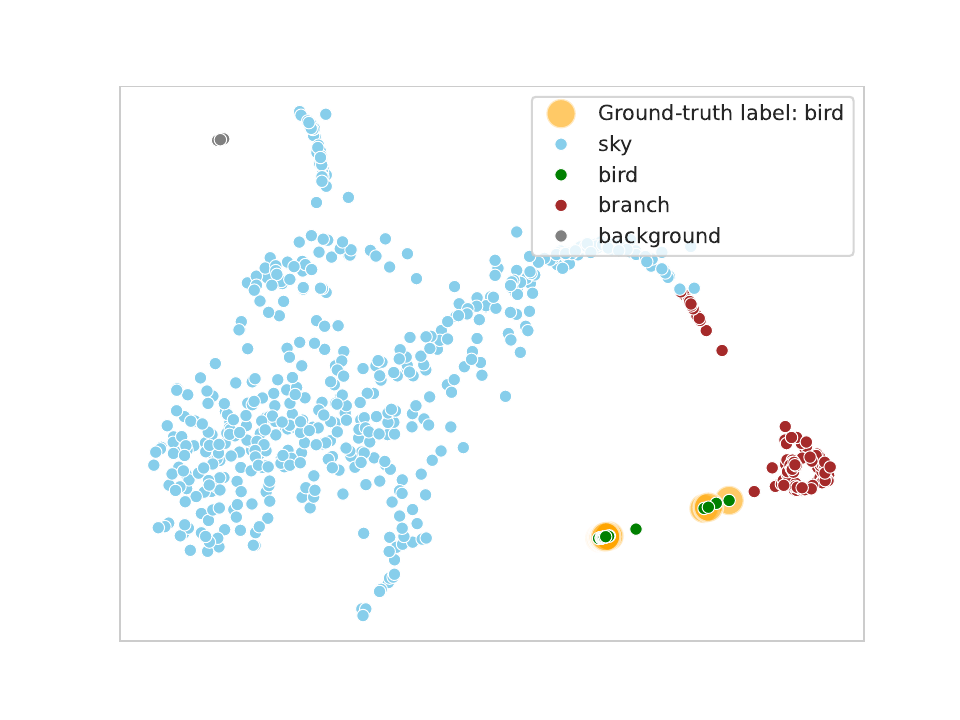} & 
         \includegraphics[trim={1.2cm, 1.2cm, 1.2cm, 1.2cm},clip, width=0.5\linewidth]{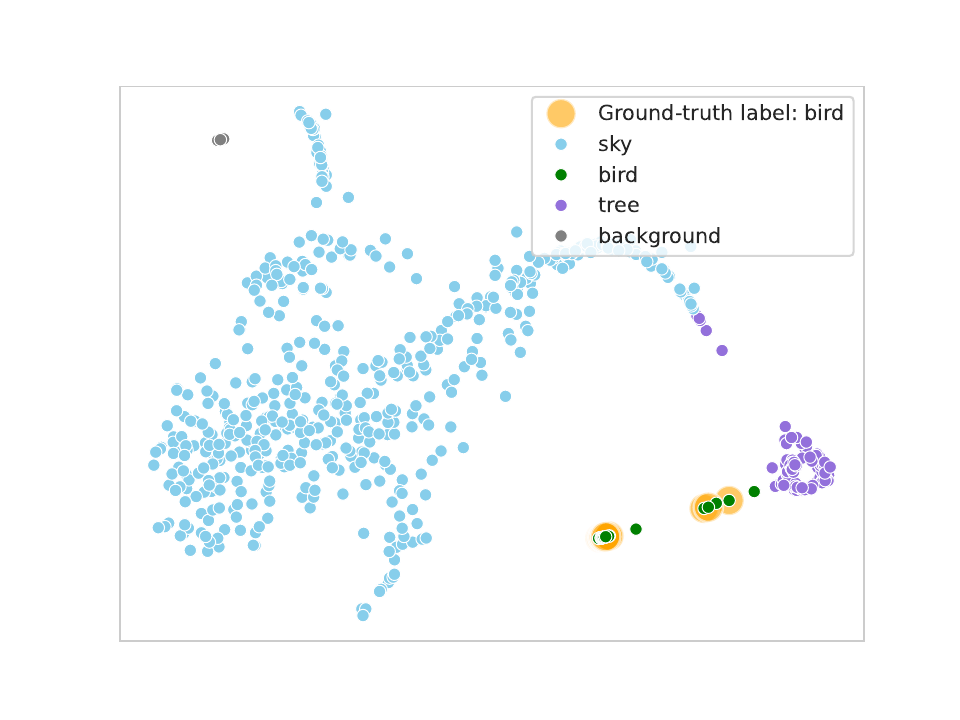} \\
    \end{tabular}
    }
    %\vspace{-5pt}
    \caption{\textbf{t-SNE analysis of patch features for different \cc of an image $q$ = ``bird".} We present patch features with their predicted closest text embedding coded in color. Text embeddings are corresponding \cc of $q$ = ``bird". We also mark the ground truth labels in orange. The sample is from VOC dataset. }
    \label{fig:tsne}
\end{figure}

To illustrate how our approach overcomes this issue we present in ~\cref{fig:tsne} a t-SNE analysis over patch features from an image of VOC dataset for the $q$ = "bird". We plot the result of classification for each separate set of \cc{}s. We highlight in orange the patches that belong to the ground truth mask of class "bird". We observe that \ccB already helps to separate the space of background concepts from "bird" patches. However, we notice that only with \ccLLM or \ccD we can separate one visible cluster left, possibly belonging to the patch features of a branch in the image, by providing "branch" in the case of \ccD or "tree" in \ccLLM. Both of our proposed methods improve the final segmentation result.

\subsection{Replacing \cc~ with sigmoid operation}
\label{app:sigmoid}

\begin{figure}
    \centering
\begin{tikzpicture}
\begin{axis}[
    width=8cm, %\textwidth,
    height=5cm,
    grid=both,
    xlabel={threshold},
    ylabel={single-IoU\%},
    legend style={at={(0.5,1.25)}, anchor=north, legend columns=-1}, % Legend position
    legend entries={thresholded sigmoid, $CC^{BG}$} % Legend labels
]

\addplot[color=blue, thick] coordinates {
    (0.512742579, 0.18312433)
    (0.5155154467, 0.18312427)
    (0.5182883143, 0.18312417)
    (0.5210611224, 0.18317117)
    (0.5238339901, 0.18322766)
    (0.5266068578, 0.1835422)
    (0.5293797255, 0.18416256)
    (0.5321525931, 0.18538046)
    (0.5349254608, 0.18778671)
    (0.5376982689, 0.19279334)
    (0.5404711366, 0.20114931)
    (0.5432440042, 0.21549122)
    (0.5460168719, 0.2371439)
    (0.5487897396, 0.27412823)
    (0.5515626073, 0.32192856)
    (0.5543354154, 0.37634808)
    (0.557108283, 0.4382375)
    (0.5598811507, 0.48912516)
    (0.5626540184, 0.5197644)
    (0.5654268861, 0.52549875)
    (0.5681997538, 0.49970445)
    (0.5709725618, 0.439114)
    (0.5737454295, 0.34883144)
    (0.5765182972, 0.24587929)
    (0.5792911649, 0.14369884)
    (0.5820640326, 0.06063196)
    (0.5848369002, 0.018464137)
    (0.5876097083, 0.0045066657)
    (0.590382576, 0.0003216077)
    (0.5931554437, 0)
};

% Horizontal dashed line
\addplot[color=black, dashed, line width=2pt] coordinates {
    (0.512, 0.593) (0.594, 0.593)
};

\end{axis}
\end{tikzpicture}
    \caption{\textbf{Sigmoid experiments.} We replace softmax with sigmoid applied on individual patch-to-query prompt similarities. We show the variation of single-IoU\% wrt. the threshold that is applied after sigmoid to decide on a positive vs. "background" class.
    To get the thresholds, we find the minimum and maximum values of the features after sigmoid and linearly sample $30$ values in this range.
    We can see that the result is sensitive to the threshold value and does not reach the baseline of $CC^{BG}$.
    }
    \label{fig:clipdinoiser_voc_sigmoid}
\end{figure}

Trying to separate a query from its background using a binary criterion is a natural alternative direction to consider, and this could be implemented with a sigmoid.

We test using a sigmoid on CLIP similarity scores. We show the results of an experiment with CLIP-DINOiser on VOC in ~\cref{fig:clipdinoiser_voc_sigmoid}. We make the following observations: (1) none of the thresholds allow us to reach the performance of \ccB, and (2) the performance is very sensitive to the threshold value. We believe this is because the CLIP space is not easily separable, as discussed in~\cref{app:separability_disc}.

\subsection{Ontology-based filtering with WordNet}
\label{sec:wordnetexp}

\begin{table}[h!]
\centering
\setlength{\tabcolsep}{1.8pt}
    \centering

    \resizebox{0.6\linewidth}{!}{
    \begin{tabular}{l|cccc}
        \toprule
         Method & MaskCLIP & TCL & DINOiser 
        \\
        \midrule
        \ccD &  \bf 25.2 & 26.0 & \bf 31.6 & 
        \\
        % \textbf{28.9} \\
        \ccD + \emph{WordNet} & \bf 25.2 & \bf 26.4 & 26.3 \\
        \ccD + \emph{WordNet} $-$ \emph{sem.\ sim.} & 21.0 & 23.4 & 25.8 \\
        \bottomrule
    \end{tabular}
    }
    \caption{\textbf{Ontology-based (WordNet) filtering} out synonyms, meronyms, hyponyms and hypernyms (at depth 1) from \ccD. Results are reported on ADE20K, as \%\metric.}
    \label{tab:sup_wordnet}
\end{table}

Here, we discuss our experiments using the WordNet ontology~\cite{fellbaum1998wordnet} for \ccD\ filtering. 
We extract synonyms, meronyms, hyponyms, and hypernyms for each query concept in-depth 1 in the WordNet ontology.
From the results in~\cref{tab:sup_wordnet}, we observe that adding such filtering on top of our \emph{semantic similarity} filtering brings little to no improvement, suggesting that \emph{semantic filtering} removes most of the contrastive concepts that interfere with a query concept. Furthermore, replacing \emph{semantic similarity} with WordNet-based filtering yields significantly worse results than our proposed \ccD.

\section{Prompting the LLM}
\label{app:llm_prompting}

In this section, we provide more details about the LLM and the prompts used.

\subsection{The LLM model}
\label{sec:moredetails:LLM}

We use the recent Mixtral-8x7B-Instruct model \cite{DBLP:journals/corr/abs-2401-04088}, a sparse mixture of experts model (SMoE), finetuned for instruction following and released by Mistral AI. More precisely, we rely on the v0.1 version of its open weights available via the Hugging Face transformers library. We run the LLM in 4-bit precision with flash attention to speedup inference.

\subsection{The prompts used for contrastive concepts}
\label{sec:moredetails:prompts}

We provide in \cref{fig:promptllm} the prompt used to generate the contrastive concepts \ccLLM\ and in \cref{fig:promptvisible} the prompt used to predict whether a concept can be seen in an image or not in order to filter \ccD.

In these prompts, we indicate the inserted input text as $\{q\}$. We follow Mixtral-8x7B Instruct's prompt template. In particular, we use \textit{<s>} as the beginning of the string (BOS) special token, as well as \textit{[INST]} and \textit{[/INST]} as string markers to be set around the instructions.

For the generation of \ccLLM, we also integrate a light post-processing step, ensuring that all generated lists have a unified format with coma separation. We do not apply any filtering or cleaning step to the LLM-generated results.

\begin{figure}[h]
\fbox{\begin{minipage}{\linewidth}
\centering\begin{minipage}{0.9\linewidth}
%\begin{quote}
    <s> [INST] You are a helpful AI assistant with visual abilities.\medskip
    
    Given an input object O, I want you to generate a list of words related to objects that can be surrounding input object O in an image to help me perform semantic segmentation.\medskip

    For example:\medskip
    
    * If the input object is 'fork', you can generate a list of words such as '["bottle", "knife", "table", "napkin", "bread"]'.\medskip
    
    * If the input object is 'child', you can generate a list of words such as '["toy", "drawing", "bed", "room", "playground"]'.\medskip

    You should not generate synonyms of input object O, nor parts of input object O.\medskip

    Generate a list of objects surrounding the input object $\{q\}$ without any synonym nor parts, nor content of it. Answer with a list of words. No explanation.\medskip
        
    Answer: [/INST] 
%\end{quote}
    \end{minipage}
    \end{minipage}}
    \caption{Prompt for \ccLLM\ contrastive concept generation.}
    \label{fig:promptllm}
\end{figure}

\begin{figure}[h]
\fbox{\begin{minipage}{\linewidth}
\centering\begin{minipage}{0.9\linewidth}
%\begin{quote}
    <s> [INST] Please specify whether $\{q\}$ is something that one can see.\medskip
    
    Reply with 'yes' or 'no' only. No explanation.\medskip
    
    Answer: [/INST] 
%\end{quote}
    \end{minipage}
    \end{minipage}}
    \caption{Prompt for \ccLLM\ visibility prediction.}
    \label{fig:promptvisible}
\end{figure}

\subsection{Example of generated \texorpdfstring{\ccLLM}~}
We present in \cref{tab:llm_cityscapes} the example of \ccLLM generated for Cityscapes dataset. We provide \cc for each query $q$ in separate rows.

\begin{table}[!h]
    \centering\vspace{10pt}
    
\vspace{3pt}
\begin{tabular}{l|l}
\toprule
Query $\query$ & $\CCLLM_\query$ \\
\midrule
road& building, tree, car, pedestrian, sky, streetlight, sidewalk, bicycle, parked car, traffic sign\\
sidewalk & building, street, car, tree, people, bike, road, park, sky, lane\\
building& sky, tree, road, car, park, people, lane, fence, house, field\\
wall& door, window, floor, ceiling, painting, light, chair, table, carpet, curtain\\
fence& grass, tree, house, car, path, post, gate, field, flowers, animals\\
pole& building, wire, tree, street, sky, fence, cable, road, banner, light\\
traffic light& road, car, building, pedestrian, sky, streetlight, traffic sign, parking meter\\
traffic sign& road, street, pole, vehicle, building, sky, pedestrian, curb, lane, light\\
vegetation& soil, tree, grass, water, animal, fence, field, sky, rock, sun\\
terrain& tree, sky, building, road, mountain, river, field, fence, vehicle, person\\
sky& tree, building, cloud, sun, bird, airplane, mountain, sea, sunset, cityscape\\
person& bike, road, car, tree, building, park, cityscape, nature, animal, sports equipment\\
rider& bicycle, road, nature, park\\
car& road, tree, building, person, parking\\
truck& road, car, building, tree, parking\\
bus& road, tree, building, sky, person, car, traffic light, bicycle, parking meter, street sign\\
train& track, grass, sky, building, platform, tree, sign, person, car, road\\
motorcycle& road, person, bike, car, traffic, building, nature, parking, city, scenery\\
bicycle& road, tree, person, park, building, grass, basket, helmet, traffic, path\\
\bottomrule
\end{tabular}
    \caption{Example of LLM-generated \ccLLM\ for Cityscapes.}
    \label{tab:llm_cityscapes}
\end{table}

\subsection{Part removal via LLM-prompting}

\begin{figure}[h]
\vspace{5pt}
    \centering
    \resizebox{\linewidth}{!}{
    \setlength\extrarowheight{-5pt}
    % \fbox{
    \begin{tabular}{c@{}c@{}c@{}c@{}c@{}c@{}}
        & GT & \ccD  & \ccD & \ccD$-$parts  & \ccD$-$parts\\
        &  &  (single) &  (all) & (single) &  (all) \\

         % other example
         \raisebox{.3\normalbaselineskip}[0pt][0pt]{\rotatebox{90}{\makecell{\stackbox{\mbox{\vphantom{X}}\\ MaskCLIP}}}} & \includegraphics[width=0.165\linewidth]{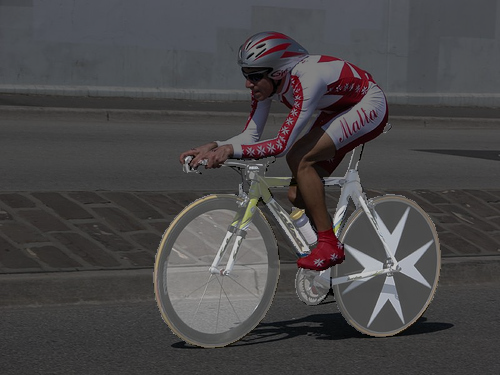} & 
         \includegraphics[width=0.165\linewidth]{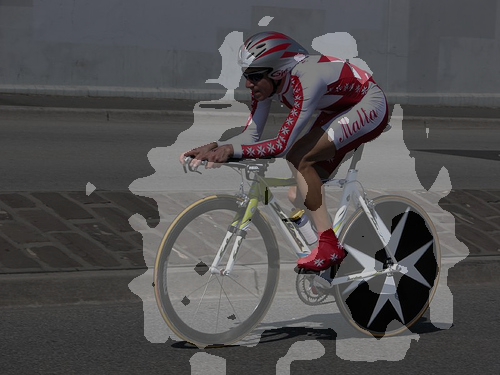} & \includegraphics[width=0.165\linewidth]{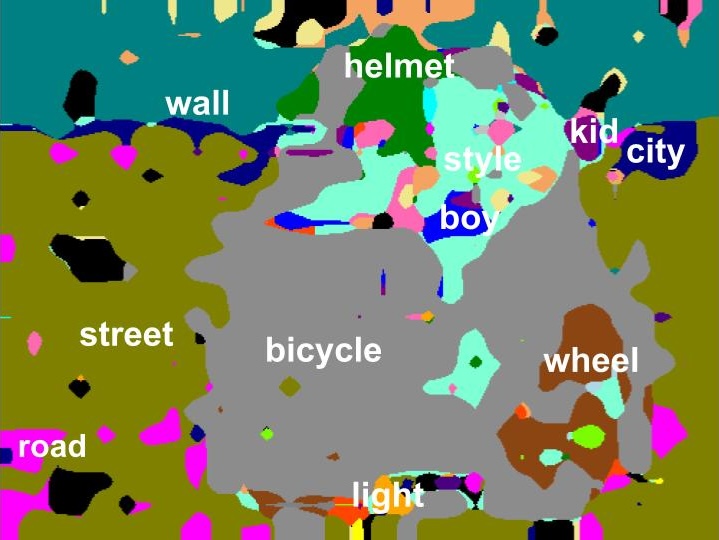} & \includegraphics[width=0.165\linewidth]{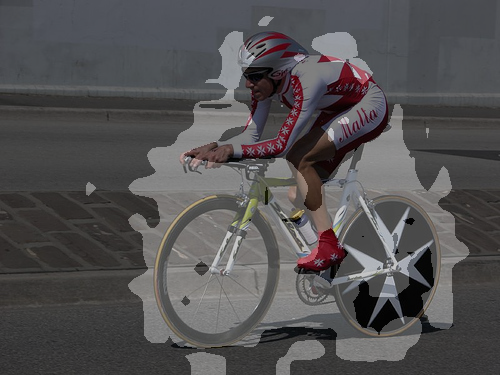} & \includegraphics[width=0.165\linewidth]{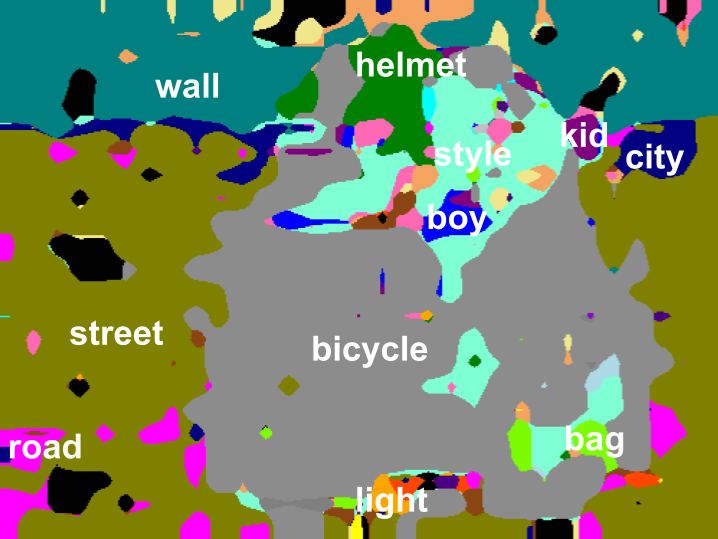}
         \\         % other example
         \raisebox{.3\normalbaselineskip}[0pt][0pt]{\rotatebox{90}{\stackbox{CLIP- \\ \makecell{DINOiser}}}} & \includegraphics[width=0.165\linewidth]{figures/examples/2008_008337_llm_bicycle_gt_blend.png} & 
         \includegraphics[width=0.165\linewidth]{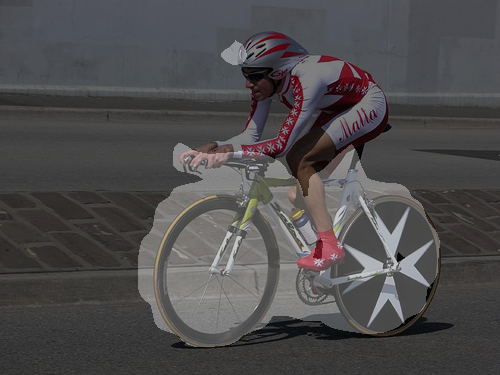} & \includegraphics[width=0.165\linewidth]{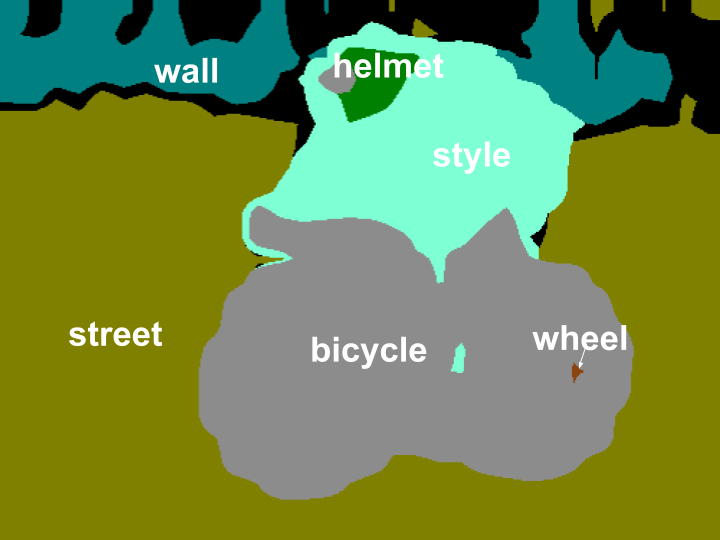} & \includegraphics[width=0.165\linewidth]{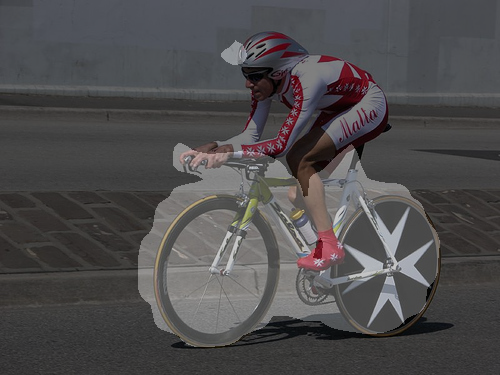} & \includegraphics[width=0.165\linewidth]{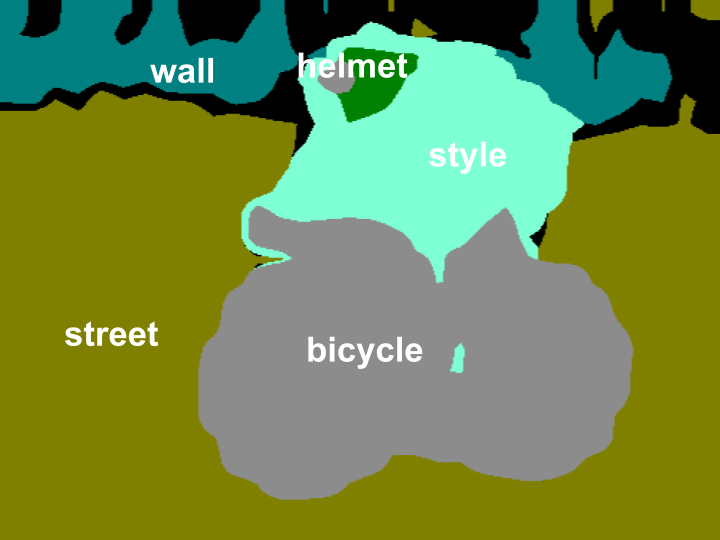}
    \end{tabular}}
    % }
    \caption{\textbf{Part removal.} We consider  an example from Pascal Context with $q$\,{=}\,\promptstyle{bicycle}{bicycle}. We show the segmentation masks produced by MaskCLIP and CLIP-DINOiser for \ccD, as well as for \ccD when parts of objects are removed (\ccD$-$parts). 
    }
    \label{fig:parts}
\end{figure}

We also explore the possibility of removing suggested contrastive concepts that can be \emph{parts} of query concepts. Note that in \ccLLM, we explicitly do it in the prompt itself (\cref{fig:promptpart}). \cref{fig:parts} presents one of such examples when removing ``wheel'' from the \ccD of query ``bicycle'' gives a slight improvement for MaskCLIP segmentation. However, we do not notice a particular improvement in the case of other segmentation methods since, typically, they refine the masks or feature maps to include localization priors. For example, in \cref{fig:parts}, the second row presents the same example for CLIP-DINOiser (DINOiser), where the improvement is marginal. Finally, we observe little or no quantitative improvement when applying part removal filtering on entire datasets. Therefore, we do not include it in our final method.

\begin{figure}[h!]
\fbox{\begin{minipage}{\linewidth}
\centering\begin{minipage}{0.9\linewidth}
%\begin{quote}
    <s> [INST] You are a helpful AI assistant with visual abilities.\medskip
    
    Given an input object O, I want you to generate a list of words that are parts of an object O.\medskip
    
    For example:\medskip
    
    * If the input object is 'rabbit', you can generate a list of words such as '["paw", "tail", "fur", "ears", "muzzle"]'.\medskip
    
    * If the input object is 'building', you can generate a list of words such as '["door", "window", "wall", "hall", "floor"]'.\medskip

    Generate a list of parts of the input object $\{q\}$. Answer with a list of words. Do not give any word that is not a part of the input object. No explanation.\medskip
        
    Answer: [/INST] 
%\end{quote}
    \end{minipage}
    \end{minipage}}
    \vspace{-5pt}
    \caption{Prompt for part prediction.}
    \label{fig:promptpart}
\end{figure}

\section{Efficiency analysis}
\label{sec:efficiency_analysis}
We first discuss the computational cost of generation and then the cost of employing generated \cc at segmentation time.

\subsection{Computational cost of \ccLLM}
We use the HuggingFace implementation of Mixtral-8x7B-Instruct-v0.1 through the $\texttt{transformers}$ library. Using 4-bit quantization and flash attention on an A100 GPU, the LLM requires $25.5$ GB of GPU memory. The average inference time required to generate a complete list of contrastive concepts for a given input query (averaged over 20 Pascal VOC queries) is $5.4$ s.

We also note that new competing LLMs, of comparable or smaller sizes than Mixtral 8x7B, are regularly being released, such as Llama 3 8B Instruct \cite{grattafiori2024llama} (Apr 2024), or Gemma-2 9B Instruct \cite{team2024gemma} (Jun 2024), Gemma-3 4B \cite{team2025gemma} (Mar 2025). For these LLMs, the GPU memory requirements are more than $3.5\times$ smaller, and the generation time more than $3.5\times$ faster.

\subsection{Computational cost of \ccD}
\paragraph{Offline cost.} In order to obtain a matrix of co-occurrences of concepts the main cost lies in the construction of the co-occurrence matrix $X$ where we iterate over 400M samples of LAION. However, we only need to go through captions, and not images, and we do it once and offline. Finally, this can be efficiently implemented by leveraging modern libraries for multiprocessing.

\paragraph{Online cost.} At runtime, the generation of contrastive concepts is fast. We provide below runtimes of all the online steps required for \ccD extraction, computed on a machine equipped with Intel(R) i7 CPU and a Nvidia RTX A5000 GPU:

\begin{itemize}
    \item Computing the CLIP embedding for a query $q$: $24.4$ ms.
    \item Mapping of query $q$ to the closest concept in $T$: $0.001$ ms.
    \item Retrieving \ccD of the closest concept in $T$: marginal cost (look-up table).
\end{itemize}

\subsection{Segmentation efficiency}
The computational cost of employing our proposed methods strictly depends on the OVSS. We present in \cref{tab_app:runtime_comparison} a comparison of runtimes between \ccB and \ccD with CLIP-DINOiser. We split the comparison into two sub-processes, text embedding extraction, and segmentation, where typically the former is stable across different OVSS methods. We observe that the main difference in runtimes stems from the embedding extraction phase, due to the different number of text prompts, that is, $|\{q\} \cup \mathcal{CC}^{BG}| = 1 + 1 = 2$, while $|\{q\} \cup \mathcal{CC}^D| = 21$ on average. (We discuss the average number of \ccD and \ccLLM in \cref{app:concepts_vs_performance}.)
% $q$ + \ccB = 2 ($|$\ccB$|$ = 1), $q +$ \ccD $\approx 21$ (we discuss the average number of \ccD and \ccLLM in \cref{app:concepts_vs_performance}. 
However, we note that this extraction time could be effectively reduced with caching mechanisms and an increase in memory consumption.

For the segmentation forward pass, we report the runtime of a single forward pass on images of size 448 x 448 when using pre-extracted text embeddings for final segmentation. We observe a negligible increase in the runtime between \ccB and \ccD.

\begin{table}
\centering\tabcolsep 5pt
\begin{tabular}{l|cc}
    \toprule
    Method & \ccB & \ccD \\
    \midrule 
Text Embeddings extraction & 49.2 $\pm$ 0.7 ms & 519.5 $\pm$  1.1 ms \\
Segmentation & 22.1 $\pm$  0.2 ms & 22.3 $\pm$ 0.2 ms \\
\hline
\end{tabular}
\caption{Runtime comparison between \ccB and \ccD.}
\label{tab_app:runtime_comparison}
\end{table}

\end{document}